\documentclass{article}

\usepackage{multirow}
\usepackage[table]{xcolor}
\usepackage{float}
\usepackage{diagbox}
\usepackage{subcaption}

\usepackage{microtype}
\usepackage{graphicx}
\usepackage{subcaption}
\usepackage{booktabs} 

\usepackage{hyperref}

\usepackage[preprint]{icml2026}

\usepackage{amsmath}
\usepackage{amssymb}
\usepackage{mathtools}
\usepackage{amsthm}

\usepackage[capitalize,noabbrev]{cleveref}

\theoremstyle{plain}

\theoremstyle{definition}

\theoremstyle{remark}

\usepackage[textsize=tiny]{todonotes}

\DeclareMathOperator*{\argmin}{arg\,min}

\makeatletter
\global\let\oriCT@@do@color\CT@@do@color 

\icmltitlerunning{Calibration and Transformation-Free Weight-Only LLMs Quantization via Dynamic Grouping}

\begin{document}

\twocolumn[
  \icmltitle{
Calibration and Transformation-Free Weight-Only LLMs Quantization via Dynamic Grouping \\
}

  \begin{icmlauthorlist}
    \icmlauthor{Xinzhe Zheng}{lu,hkust}
    \icmlauthor{Zhen-Qun YANG}{polyu}
    \icmlauthor{Zishan Liu}{lu}
    \icmlauthor{Haoran Xie}{lu}
    \icmlauthor{S. Joe Qin}{lu}
    \icmlauthor{Arlene Chen}{xiaoi}
    \icmlauthor{Fangzhen Lin}{hkust}
  \end{icmlauthorlist}

  \icmlaffiliation{lu}{Division of Artificial Intelligence, School of Data Science, Lingnan University, Hong Kong, China}
  \icmlaffiliation{hkust}{Department of Computer Science and Engineering, The Hong Kong University of Science and Technology, Hong Kong, China}
  \icmlaffiliation{polyu}{Department of Computing, The Hong Kong Polytechnic University, Hong Kong, China}
  \icmlaffiliation{xiaoi}{Xiaoi Robot Inc., Shanghai, China}

  \icmlcorrespondingauthor{Xinzhe Zheng}{xinzhezheng@ln.hk}
  \icmlcorrespondingauthor{Haoran Xie}{hrxie@ln.edu.hk}

  \icmlkeywords{Machine Learning, }

  \vskip 0.3in
]

\printAffiliationsAndNotice{}  

\begin{abstract}

Large Language Models (LLMs) deliver strong performance but are difficult to deploy under tight memory and compute constraints. Low-bit post-training quantization (PTQ) is a promising direction; however, it typically relies on calibration data, auxiliary transformations, and GPU tools.
To address these limitations, we propose MSB (Multi Scale Binary), a calibration-free and transformation-free PTQ method that generalizes binary quantization to multi-bit settings. MSB optimizes a dynamic grouping criterion that minimizes within group variance, yielding group-wise multiscale levels that can be applied consistently across granularities from per tensor to block-wise configurations with 64 elements groups per row, without calibration or intermediate transforms. We implement the optimization in a CPU based solver for the quantization step and evaluate using standard bfloat16 execution without low-bit packing.
On Llama 3.2 3B, MSB achieves 8.43 perplexity on WikiText-2 under 4-bit weight only block-wise quantization, compared to 7.81 in full precision and 12.23 with GPTQ its default setup. Overall, MSB provides a new optimization perspective for low-bit PTQ while simplifying the pipeline by removing calibration and transformations.

\end{abstract}

\section{Introduction}

Large Language Models (LLMs) such as LLaMA \cite{grattafiori2024LLaMA3herdmodels}, DeepSeek \cite{deepseekai2025deepseekr1incentivizingreasoningcapability}, and Gemma \cite{team2025gemma} achieve state-of-the-art (SOTA) performance across a wide range of Natural Language Processing (NLP) tasks, but their parameter scale creates substantial deployment costs. Even considering weight only storage, modern models can require hundreds of gigabytes of memory: for example, a 671B parameter model \cite{deepseekai2025deepseekr1incentivizingreasoningcapability} requires approximately 1.34~TB in FP16 and 671~GB in INT8 to store weights alone. This already exceeds the 640~GB total GPU memory of a single 8$\times$ H100 node \cite{10.1109/MM.2023.3256796}, motivating compression techniques that reduce the memory footprint of pretrained weights.

Quantization addresses this challenge by representing weights and optionally activations with fewer bits, reducing memory traffic and enabling low precision arithmetic \cite{dettmers2022llmint88bitmatrixmultiplication}. 
As a result, quantization is central to deployment in both cloud scale serving and edge/on-device environments \cite{tan2024mobilequant}. In particular, we focus on 4-bit weight only post-training quantization (PTQ), which is a widely studied low-bit regime (typically $\le4$ bits) offering substantial compression while often retaining competitive accuracy.

Existing low-bit quantization approaches for LLMs broadly fall into quantization-aware training (QAT) and PTQ.
QAT incorporates quantization effects during training so that weight parameters adapt to low precision computation \cite{wang2023bitnetscaling1bittransformers}, 
but it usually requires retraining or significant fine tuning under specialized pipelines, which can be expensive for models that have already been trained.
In contrast, PTQ produces a quantized model directly from pretrained weights \cite{frantar-gptq}, making it a practical choice when compute budgets, time constraints, or data access limitations preclude quantization aware retraining; we therefore focus on PTQ in this paper.

Within PTQ, we focus on 4-bit weight only quantization, a widely used operating point~\cite{frantar-gptq,liu2025spinquantllmquantizationlearned}. 
We do not target ultra low bit widths, which often constitute a qualitatively different regime. 
In this 4-bit setting, practical pipelines often pair the quantizer with calibration and/or function preserving reparameterizations, such as rescaling or rotations, to mitigate outliers and reduce quantization error 
\cite{lin2023awq,liu2025spinquant}.
While effective, these ingredients introduce extra dependencies and design choices.
In this work, we study a complementary setting: calibration-free and transformation-free PTQ that operates solely on pretrained weights.

Our design is motivated by binary PTQ, where uniform binarization is typically insufficient for LLMs. 
Effective methods therefore rely on structured partitioning: they reserve limited higher capacity representations (e.g., residual or mixed precision) for the most sensitive weights while binarizing the remainder \cite{huang2024BiLLMpushinglimitposttraining}. 
This suggests a broader principle that under tight precision budgets, performance depends not only on the quantizer, but also on how weights are partitioned and how representational capacity is allocated across groups of heterogeneous sensitivity.

\begin{figure}[h]
    \centering
    \includegraphics[width=1.0\linewidth]{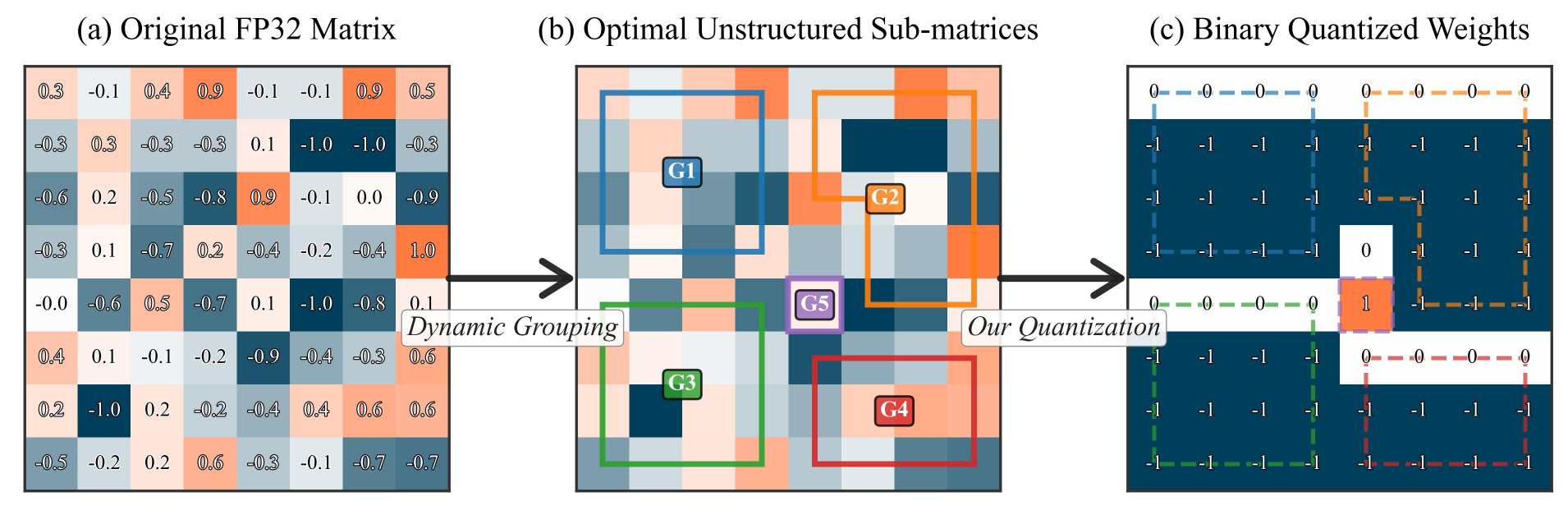}
    \caption{Overview of our dynamic grouping framework for calibration-free, transformation-free low-bit PTQ.}
    \label{framework}
\end{figure}

Building on this principle, we ask whether it can generalize to practical $n$-bit quantization beyond 1-bit. 
Instead of explicitly selecting salient weights, we allocate representational capacity through optimized grouping, by introducing a multi scale binary (MSB) approximation: different groups of weights share different scale parameters while retaining a binary sign structure.
We formulate an MSB objective and optimize it with variance-aware magnitude grouping using greedy solvers, yielding a calibration-free and transformation-free framework that applies to both whole-matrix and block-wise settings.

In this work, we (i) formulate a calibration-free, transformation-free objective for MSB weight approximation and derive an equivalent magnitude grouping form based on within group statistics, yielding a variance based grouping criterion; (ii) develop objective driven, distribution sensitive solvers, including a dynamic programming reference and efficient approximations, Greedy Grouping and Windowed Greedy Merging (WGM), implemented for CPU-based offline PTQ; and (iii) show that the same objective and solver template supports both per-tensor (6-bit) and block-wise (4-bit) settings, enabling practical 4-bit weight only PTQ with competitive accuracy relative to widely used baselines.

\section{Related Work}

\subsection{PTQ for LLMs}

\textbf{Rounding based weight-only baselines.}
A common starting point for PTQ is round-to-nearest (RTN), which applies uniform quantization using per tensor or block wise scaling (and optionally a zero point or clipping) 
and rounds each weight to the nearest quantization level.
Practical toolchains, e.g. bitsandbytes (BnB)~\cite{dettmers2022llmint88bitmatrixmultiplication, dettmers2023qlora}, further adopt blockwise scaling and packed low-bit formats, including widely used 4-bit representations such as FP4/NF4, to reduce memory footprint and improve throughput.

\textbf{Calibration-free optimization.}
Beyond direct rounding, calibration-free methods optimize quantized weights using only pretrained parameters.
For example, HQQ \cite{badri2023hqq} formulates weight-only quantization as a robust optimization problem and solves it efficiently (e.g., via half-quadratic splitting) without calibration data, aiming to reduce distortion compared with naive rounding while keeping a simple inference time representation.

\textbf{Calibration-based error minimization and second-order variants.}
Methods such as GPTQ \cite{frantar-gptq} improve weight-only PTQ by explicitly minimizing output error under calibration.
A key mechanism is to use a small calibration set to estimate layer-wise sensitivity via approximate second-order information and then perform sequential/blockwise quantization with error compensation, substantially reducing accuracy loss at 3 and 4 bits compared with pure rounding.

\textbf{Calibration- and transformation-based PTQ.}
Another widely used line of work combines low-bit quantization with lightweight function-preserving reparameterizations to mitigate outliers and improve quantizability.
Representative examples include equivalent rescaling/shifting based on activation statistics (e.g., AWQ \cite{lin2023awq}), differentiable tuning of clipping and equivalent transformations (e.g., OmniQuant \cite{shao2023omniquant}), and rotation-based reparameterizations that reduce outliers prior to quantization (e.g., SpinQuant \cite{liu2025spinquant}).
These approaches can be highly effective, but they introduce additional data dependencies for calibration and extra design choices for transformation parameterizations and tuning.

\subsection{Binary PTQ for LLMs}
Traditional binary quantization such as XNOR-Net \cite{10.1007/978-3-319-46493-0_32} represents weights with 1-bit values and is often characterized by an objective that approximates real-valued weights with a scaled binary form.
Modern binary PTQ for LLMs like BiLLM \cite{huang2024BiLLMpushinglimitposttraining} rarely binarizes all weights uniformly; instead, effective schemes apply heterogeneous treatment,  allocate limited higher capacity representations (e.g., residual terms) to sensitive weights while binarizing the remaining majority, preventing severe degradation.
This regime highlights a core principle: under extreme precision constraints, performance depends not only on the quantizer, but also on how weights are partitioned and how limited representational capacity is allocated across groups of heterogeneous sensitivity.

\textbf{Connection to our work.}
Motivated by the objective-centric view and heterogeneous treatment in binary PTQ, 
we generalize this perspective to MSB PTQ.
Our method studies a complementary setting that operates directly on pretrained weights, avoids calibration data and function-preserving transformations, and optimizes an $n$-bit objective via distribution-sensitive grouping.

\section{Methodology}

This section formalizes our dynamic grouping objective and presents four solvers spanning accuracy runtime trade-offs. We describe the boundary parameter $\lambda$ and analyze the time/memory costs of each variant.

\subsection{Preliminary}

The classical objective for binary quantization of XNOR-Net \cite{10.1007/978-3-319-46493-0_32} approximates a full precision weight tensor by a scaled binary codebook.
Given a weight matrix $\mathbf{W}\in\mathbb{R}^{m\times l}$, XNOR-style binarization seeks a binary matrix $\mathbf{B}\in\{-1,+1\}^{m\times l}$ and a positive scale $\alpha>0$ such that
\begin{equation*}
\label{eq:xnor_obj}
\alpha^{\ast}, \boldsymbol{B}^{\ast} = \argmin_{\alpha, \boldsymbol{B}} \left\| \boldsymbol{W} - \alpha \boldsymbol{B} \right\|_2^2. \tag{1}
\end{equation*}

This problem admits a closed form solution:
$\mathbf{B}^\star=\mathrm{sign}(\mathbf{W})$ and
$\alpha^\star=\|\mathbf{W}\|_1/(ml)$ \cite{10.1007/978-3-319-46493-0_32}.

This formulation captures the canonical scaled binary form $\{\pm\alpha\}$ and provides an objective-centric viewpoint that we later generalize to multi-bit settings.

\subsection{MSB Optimization Objective} 
Based on the ideas discussed above, we propose a new MSB optimization objective that minimizes the total quantization loss \ref{eq:xnor_obj} across all unstructured sub-matrices by identifying their optimal grouping based on a predetermined quantization loss. 

The following variables define our optimization objective:

\begin{itemize}
    \item Let $\boldsymbol{A} \in \mathbb{R}^{m \times n}$ be the given matrix and $\mathbb{A}$ be the all possible set of arbitrary grouping of $\boldsymbol{A}$.
    \item $g$ represents the number of non-overlapping unstructured sub-matrices, satisfying $1 \leq g \leq |\boldsymbol{A}|$, where $|\boldsymbol{A}|$ denotes the total number of non-zero elements in matrix $\boldsymbol{A}$.
    \item $\alpha \in \mathbb{R}$ and $\boldsymbol{B} \in \{-1, 1\}^{m \times n}$ are used for the binary approximation of $\boldsymbol{A}$. Each $\alpha_i$ and $\boldsymbol{B}_i$ approximates one $\boldsymbol{A}_i$, so $\boldsymbol{B}_i$ refers to the same set of matrix elements as partition $\boldsymbol{A}_i$.
\end{itemize}

Thus, we partition $\boldsymbol{B}$ identically to $\boldsymbol{A}$, such that $\boldsymbol{A}_1 = \boldsymbol{B}_1, \dots.,\boldsymbol{A}_g = \boldsymbol{B}_g$ and $\mathbb{A} = \mathbb{B}$ in turns of matrix index.
We group matrix $\boldsymbol{A}$ into $g$ non-overlapping sub-matrices $\{\boldsymbol{A}_i\}^g_{i=1} = \{\boldsymbol{A}_1, \boldsymbol{A}_2, \dots, \boldsymbol{A}_g\} \in \mathbb{A}$. For simplicity, we focus on non-zero elements and place any zero elements into a special group, if they exist. This special group is omitted from consideration as it has a zero loss. Additionally, $\mathbb{A}$ encompasses all possible non-overlapping partitions using all elements of $\boldsymbol{A}$.
Based on this, our proposed optimization objective is:
\begin{align*}
    g^{\ast},\{\boldsymbol{A}_i^{\ast}\}^{g^{\ast}}_{i=1} = 
    \argmin_{\substack{
        \{\boldsymbol{A}_i^{\ast}\}^{g^{\ast}}_{i=1} \in \mathbb{A}\\
        \{\boldsymbol{B}_i^{\ast}\}^{g^{\ast}}_{i=1} \in \mathbb{B}\\
        1 \leq g \leq |\boldsymbol{A}|
    }} \sum^g_{i=1}||\boldsymbol{A}_i - \alpha_i\boldsymbol{B}_i||_2^2 
    + \frac{1}{|\boldsymbol{B}_i|}. 
\end{align*}
The optimization objective consists of two main components. 
The first summation represents the quantization loss. The second summation serves as a regularization term that penalizes the inverse of the group size. We penalize the inverse of the group size because smaller groups lead to more partitions, resulting in lower efficiency due to a smaller group for each scale term \( \alpha \).
It is important to note that the partitions need not be block-structured; each partition is simply a set of elements from $\boldsymbol{A}$. When combining all $g$ sub-matrices, we recover the original matrix $\boldsymbol{A}$.
To solve the objective, by substituting the optimal values, we can rewrite the squared 2-norm.
\begin{align*}
    ||\boldsymbol{A} - \alpha^{\ast}\boldsymbol{B}^{\ast}||_2^2 = ||\boldsymbol{A}||_2^2 - \frac{||\boldsymbol{A}||_1^2}{|\boldsymbol{A}|}. 
\end{align*}
Consider the variance of absolute value of $\boldsymbol{A}_i$, let $\Tilde{\boldsymbol{A}_i}$ be $\boldsymbol{A}_i$ with element-wise absolute value, 
\begin{gather*}
    \text{Var}(\Tilde{\boldsymbol{A}_i}) 
    = \frac{||\boldsymbol{A}_i||_2^2}{|\boldsymbol{A}_i|} - \left( \frac{||\boldsymbol{A}_i||_1}{|\boldsymbol{A}_i|} \right)^2 \\
    |\boldsymbol{A}_i|\text{Var}(\Tilde{\boldsymbol{A}_i}) = ||\boldsymbol{A}_i||_2^2 - \frac{||\boldsymbol{A}_i||_1^2}{|\boldsymbol{A}_i|} = ||\boldsymbol{A}_i - \alpha_i^{\ast}\boldsymbol{B}_i^{\ast}||_2^2. 
\end{gather*}
Finally, Since $|\boldsymbol{A}_i| = |\boldsymbol{B}_i|$, rewrite the optimization objective above as
\begin{align*}
    g^{\ast}, \{\boldsymbol{A}_i^{\ast}\}^{g^{\ast}}_{i=1}
    = \argmin_{\substack{
        \{\boldsymbol{A}_i^{\ast}\}^{g^{\ast}}_{i=1} \in \mathbb{A}\\
        1 \leq g \leq |\boldsymbol{A}|
    }} \sum^g_{i=1}|\boldsymbol{A}_i|\text{Var}(\Tilde{\boldsymbol{A}_i}) + \frac{1}{|\boldsymbol{A}_i|} \label{objective}
    \tag{2}. 
\end{align*}

And for a Grouping $\mathcal{G}=\{\mathbf{A}_i\}_{i=1}^{g}\in\mathbb{A}$, define its objective value as
\begin{align*}
    \mathrm{cost}(\mathcal{G}) \;=\; \sum_{i=1}^{g}\Big( |\mathbf{A}_i|\,\mathrm{Var}(\tilde{\mathbf{A}}_i) \;+\; \frac{1}{|\mathbf{A}_i|} \Big).
\end{align*}

\subsection{Four Tailored Algorithms}
To address the optimization objective, we propose four algorithms, each tailored to different accuracy runtime regimes. Algorithm \ref{alg:DG}, Dynamic Grouping (DG), employs classic dynamic programming to guarantee optimal solutions. Meanwhile, Algorithm \ref{alg:GG}, Greedy Grouping (GG), adopts a heuristic strategy to improve computational efficiency with reasonable solution quality. Algorithm \ref{alg:WGM}, Windowed Greedy Merging (WGM), is an efficient approximation method based on Greedy Grouping that achieves a strong balance between quantization performance and speed. Algorithm 4, Local Optimizing Windowed Greedy Merging (WGM-LO), accelerates WGM by using equal-range binning for initialization and then refines adjacent group boundaries via a lightweight stochastic local search. Detailed pseudo code for Algorithm \ref{alg:DG}, \ref{alg:GG} and \ref{alg:LOWGM} is given in Appendix \ref{sec:3AlgoPseudocode}.

\subsubsection{Algorithm \ref{alg:DG}, Dynamic Grouping} 
This algorithm formulates the optimization problem as a dynamic programming task, systematically exploring all possible groupings to find the global optimum. 
Firstly, by observing that the partition with $\mathrm{cost}(\mathcal{G^{\ast}})$, each group corresponds to a contiguous interval in the sorted sequence, we sort all entries of $\Tilde{\boldsymbol{A}}$ in ascending order while keeping track of their original indices. This is because 
there exists an optimal solution minimizing variance in which each group corresponds to a contiguous interval in the sorted order.
We then break the problem into sub-problems. Suppose we aim to divide $n$ elements into $k$ groups with minimum cost. The optimal configuration must consist of the $k-1$ groups from the first $i$ elements with minimum cost and the remaining $n-i$ as the last group
\begin{align*}
    dp[k][n] = \min_{1 \leq i < n}dp[k-1][i] + f([i:n]) \label{recurrence}
    \tag{3} .
\end{align*}
Where $f$ is the optimization objective \ref{objective} for one group, $dp$ is the dynamic programming table, with the total row of max group number and total column of max number of elements (both including 0). And $dp[k][n]$ refers to the minimum cost of partitioning $n$ elements into $k$ groups, $f([i:n])$ refers to the cost for the remaining elements. We prove this is correct by contradiction. Assume there is a partition of $k$ groups for $n$ elements with lower cost than $dp[k][n]$
\begin{align*}
    f([0:n]) = dp[k-1][i^{\ast}] + f([i^{\ast}:n]) < dp[k][n].
\end{align*}
But $dp[k][n]$ is defined as the minimum over all possible $i$, including $i^{\ast}$. This contradicts the assumption, so $dp[k][n]$ from (\ref{recurrence}) is indeed the minimum cost.

We proved the sub-problem is correct, now construct the algorithm based on Equation (\ref{recurrence}). We set the initial $dp_0$ as a matrix full of infinite value except the first column and row.
Then We precompute prefix sums of the sorted values and their squares to evaluate the interval cost in Eq.~(\ref{objective}) in $O(1)$ time.
After filling the DP table, the optimal objective value is obtained by minimizing over the last column and the optimal group numbers is the row number of the optimal value.
We store the argmin split indices in an additional table to reconstruct the optimal partition via backtracking, and compute the per-group scale parameters (e.g., absolute means) for each recovered interval to obtain the corresponding $\alpha$ values.

\subsubsection{Algorithm \ref{alg:GG}, Greedy Grouping} 
Modern LLMs often have linear layers with about $10^6$ weights per matrix (e.g., $1024\times1024$) or more, making exact dynamic programming impractical; we therefore adopt a greedy merging heuristic, which is fast enough to be feasible at this scale.
We therefore approximate Algorithm~\ref{alg:DG} with a greedy merging heuristic: we sort nonzero entries, initialize singleton groups, and maintain a min-heap of adjacent merge costs (size $mn-1$). We repeatedly pop the minimum-cost merge, update its neighboring costs, and stop once the target number of groups is reached.

\subsubsection{Algorithm \ref{alg:WGM}, Windowed Greedy Merging} 
To improve Algorithm \ref{alg:DG}, and achieve a better efficiency-quality trade-off, we modify the initial merge cost array, and propose our Algorithm \ref{alg:WGM}. Instead of starting at $mn$ groups each of size 1, we start at $mn/k$ groups, where k is the window size for each initial group. We provide a detailed description of the algorithm \ref{alg:WGM} below.
\renewcommand{\thealgorithm}{3}
\begin{algorithm}
\caption{Windowed Greedy Merging (WGM)}\label{alg:WGM}
\begin{algorithmic}[1]
\STATE {\bfseries WGM: }{$\boldsymbol{A}\in\mathbb{R}^{m \times n}, g\text{ (max group}), k\text{ (window size})$}
\STATE sort absolute value of non-zero entry into an array
\STATE initial the merge cost array $[(i \text{ (group start}), i+k \text{( group end}), cost, \mu)]$ for $i$ in range(0, $mn$ - 1, k)
\STATE initialize the ignore array
\STATE heapify the merge cost array
\WHILE{len(merge cost array) $>$ g}
\STATE currentMerge $\gets$ heapop
\IF{currentMerge in ignore array}
\STATE remove currentMerge from ignore array
\STATE continue
\ENDIF
\STATE push two new neighbouring merges as updates
\STATE update ignore array by two old merges, invalidated by the two new merges
\ENDWHILE
\STATE \textbf{return} merge cost array
\end{algorithmic}
\end{algorithm}
\subsubsection{Algorithm 4, Local Optimizing Windowed Greedy Merging} 
Algorithm~\ref{alg:WGM} initializes groups with equal-size windows, producing balanced group sizes but many initial groups that can slow merging. 
Algorithm~\ref{alg:LOWGM} instead uses equal-range binning over $[w_{\min},w_{\max}]$, which groups numerically similar values and typically yields low within-bin variance while reducing the initial group count. This strong initialization allows WGM-LO to start from a much smaller number of initial groups while remaining well aligned with the variance-based objective.
Although this can create highly unbalanced bins, it substantially accelerates the greedy merging stage. 
To alleviate boundary artifacts induced by unbalanced bins, after the greedy merging phase, we apply a stochastic local optimization procedure that perturbs adjacent group boundaries and accepts a move only if it decreases the objective.
We terminate optimization when no improvement is found for a fixed number of consecutive attempts or when the improvement falls below a small threshold.
In practice, the merging phase becomes faster due to the small initial group count, and the main computational cost comes from the local optimization stage for per-tensor setting. However in block-wise setting the local optimization dominates and cost significant slow down.

In summary, our algorithmic framework provides both theoretical optimality (via Algorithm \ref{alg:DG}) and practical efficiency (via Algorithm \ref{alg:GG}, \ref{alg:WGM} and \ref{alg:LOWGM}), offering flexible solutions for a wide range of problem sizes and application scenarios. \textbf{Dynamic Grouping} identifies the optimal grouping under our objective, and serves as an oracle reference in our ablation study.
\textbf{Greedy Grouping} is the most accurate heuristic but becomes inefficient on very large instances due to its fine-grained merge schedule. 
\textbf{Windowed Greedy Merging (WGM)} mitigates this by coarsening early decisions, which is most beneficial in per-tensor quantization where the instance is large, while offering limited benefit on small block-wise tiles.
\textbf{Local-Optimized WGM (WGM-LO)} further accelerate WGM for large per-tensor instances via a coarse initialization followed by local boundary optimization.
Overall, these three heuristics provide practical solvers for modern hardware, offering a clear speed--quality trade-off while remaining faithful to the underlying optimization objective.

\subsection{Interpretation and Boundary of \texorpdfstring{$\lambda$}{lambda}}
We introduce a regularization weight $\lambda$ to trade off the variance term against a size-dependent penalty in the dynamic grouping objective. Since the penalty scales as $\lambda/|\mathbf{A}_i|$, smaller $\lambda$ favors finer partitions (more, smaller groups), whereas larger $\lambda$ encourages coarser groupings (fewer, larger groups). To keep the two terms comparable across group sizes, we normalize the variance contribution by group mass, leading to
\begin{align*}
    \mathrm{cost}(\mathcal{G}) \;=\; \sum^g_{i=1} \left( \textstyle \frac{|\boldsymbol{A}_i|}{|\boldsymbol{A}|} Var(\Tilde{\boldsymbol{A}_i}) + \textstyle \frac{\lambda}{|\boldsymbol{A}_i|} \right).
\end{align*}
Algorithmically, this only changes the per-group cost by applying the above normalization and scaling the penalty by $\lambda$.
We define $\lambda_{\min}$ and $\lambda_{\max}$ as the values that induce, respectively, the finest admissible partition and the single-group solution, and reparameterize $\lambda=\Lambda(\tilde{\lambda})$ with $\tilde{\lambda}\in[0,1]$ via a monotone map for interpretability.
See Appendix \ref{sec:Determining the Boundary of lambda} for additional details and derivation.

\subsection{Time Complexity}
Given the input matrix $\boldsymbol{A} \in \mathbb{R}^{m \times n}$, the sorting of all non-zero entries takes $O(mn\log(mn))$, prefix sum and prefix squared 2 norm takes $O(mn)$, complete the dynamic programming table takes $O((mn)^2\cdot \text{max group number} )$ since computing cost using prefix sum and prefix squared 2 norm takes $O(1)$. So \textbf{Dynamic Grouping}'s time complexity is $O((mn)^3)$, which is infeasible to run to completion. Then, \textbf{Greedy Grouping} is designed to approximate the optimal solution faster. We sort all non-zero entries and use a min heap storing the merging cost between each group, i.e. heap has size $mn$. Then, we update the heap until only the desired number of groups remains. This greedy merging only takes $O(mn\log(mn))$ since the sorting of non-zero entries takes $O(mn\log(mn))$, popping and updating the heap takes $O(mn\log(mn))$.
For the WGM, it effectively reduces the merging time complexity to $O(\frac{mn}{k}\log(\frac{mn}{k}))$ at the cost of accuracy, since we may overlook a better merge in the initial $mn/k$ groups of size $k$. Lastly for the WGM-LO, the merging complexity become only $O(klogk)$ but with a higher cost local optimizing parse of $O(Tg)$, where $k$ is the initial number of groups, T is the max iteration and the optimization and g is the final number of groups.

\section{Experiments}

In this section, We evaluate MSB under two granularities. Our primary setting is 4-bit block-wise quantization, 
in which each weight matrix is partitioned into 64‑row blocks that are quantized independently for efficiency
We also report 6-bit per-tensor results, where each matrix is quantized as a whole using a single global grouping. Algorithm~\ref{alg:WGM} (\textsc{WGM}) is used in both regimes, while Algorithm~\ref{alg:LOWGM} (\textsc{WGM-LO}) is evaluated only in the per-tensor setting.
Together, the results suggest that our framework is robust to the choice of quantization granularity, performing well in both tensor-level and block-level regimes.
Additional experimental details and results for all four algorithms, as well as extended tables for both per-tensor and block-wise settings, are provided in Appendix~\ref{sec:Supplementary Experiments, Settings, and Full Results}.

\subsection{Settings}

We evaluate under simulated MSB PTQ. For a target bit-width $b$, each weight in a group is represented as
\begin{align*}
    \hat{w} = \alpha_{z} \cdot s, \qquad s\in\{-1,+1\},\;\; z\in\{1,\dots,2^{b-1}\},
\end{align*}
i.e., a binary sign multiplied by one of $2^{b-1}$ group specific positive scales $\{\alpha_z\}$. This yields a symmetric $2^b$-level codebook $\{\pm \alpha_z\}$ while retaining a binary sign structure.
All quantized values are decoded and stored in \texttt{bfloat16}, 
so our results isolate the effect of the proposed grouping optimization from backend specific implementations.
As for storage, for 4-bit block-wise quantization, the theoretical effective storage is 6.00 bits/weight without double quantization (DQ) \cite{dettmers2023qlora} or 4.78 bits/weight with DQ. For per-tensor quantization, metadata overhead is negligible relative to the weight tensor size. We report main results without DQ, and Appendix~\ref{sec:dq} reports the model-dependent QA and PPL impact of DQ.
We adopt simulation because the goal of this work is to validate the proposed optimization objective; implementing a dedicated low-bit kernel is orthogonal and not prioritized given the relatively high time complexity of the quantization procedure.
Exact zeros, when present, are kept as \texttt{bfloat16} zeros; in practice such entries are extremely sparse and do not affect codebook construction. 

\subsubsection{Models and datasets}
\label{sec:Models and datasets}

We facilitate our method on the LLama, Falcon \cite{Falcon3} and Gemma families. We evaluate quantization performance using seven zero-shot common sense QA tasks used in related researches \cite{shang2023pb, huang2024BiLLMpushinglimitposttraining}, 
enabling direct comparability with prior research, i.e. ARC-C and ARC-E \cite{clark2018thinksolvedquestionanswering}, BooIQ \cite{clark-etal-2019-boolq}, HellaSwag \cite{zellers-etal-2019-hellaswag}, OPQA \cite{mihaylov-etal-2018-suit}, PIQA \cite{Bisk_Zellers_Le_bras_Gao_Choi_2020}, WinoGrande \cite{10.1145/3474381}. Next, we conduct experiments on perplexity (PPL) using three datasets that have also been utilized in previous research \cite{shang2023pb, huang2024BiLLMpushinglimitposttraining}, i.e. the Wikitext 2 (WK2) \cite{merity2016pointersentinelmixturemodels}, the PTB \cite{marcus-etal-1993-building} and the C4 \cite{JMLR:v21:20-074}.

\subsubsection{Baseline}
We choose baselines to reflect the main families of weight-only low-bit PTQ used for LLMs, including calibration-based methods (GPTQ) and calibration-free/data-free approaches (RTN, BnB, HQQ)
Specifically, we consider GPTQ as a strong calibration-based 4-bit PTQ reference, and RTN (round-to-nearest) as a simple, efficient quantization baseline. 
We further include BnB and HQQ as representative calibration-free baselines that do not require a separate calibration dataset and instead rely on weight statistics and/or a dedicated quantization objective; this makes them the closest comparisons to our approach, which also avoids calibration and auxiliary transformations.
For baselines (GPTQ, RTN, BnB, HQQ), we use official implementations and/or released quantized checkpoints, following their standard configurations. 

\begin{table}[H] 
  \footnotesize 
  \caption{Comparison of 4-bit block-wise and 6-bit per-tensor weight-only LLM quantization: average performance across seven QA task and average perplexity across three datasets. WGM: Windowed Greedy Merging algorithm with window size $w$ = 64 and $w$ = 1 for per-tensor and block-wise setting respectively. WGM-LO: Local Optimizing Windowed Greedy Merging algorithm with $T$=12, $k$ = 256. Standard error of all tasks are within 0.02. ``/'' denotes settings where the method is not applicable; we therefore do not run the corresponding experiment. Optimal results are in bold; our result rows are shaded.
  }
  \begin{tabular}{lc|cc|cc} 
    \hline
    && \multicolumn{2}{c|}{4 bit blockwise} & \multicolumn{2}{c}{6 bit per tensor}\\
    \hline
    \textbf{Model} & \textbf{Method} & \textbf{QA}$\uparrow$ & \textbf{PPL}$\downarrow$ & \textbf{QA}$\uparrow$ & \textbf{PPL}$\downarrow$\\
    \hline
    \multirow{7}{3em}{Llama 3.2 1B} & FP & 0.568 & 13.78 & 0.568 & 13.78\\
    \hline
    & GPTQ & 0.518 & 56.88 & / & /\\
    & RTN & 0.541 & 16.56 & 0.393 & 169.47\\
    & BnB & \textbf{0.553} & \textbf{15.34} & / & /\\
    & HQQ & 0.544 & 15.75 & 0.404 & 106.02\\
    \rowcolor{gray!20} 
    & WGM & 0.540 & 15.45 & 0.561 & \textbf{14.18}\\
    \rowcolor{gray!20} 
    & WGM-LO & / & / & \textbf{0.562} & 14.27\\
    \hline
    \multirow{7}{3em}{Llama 3.2 3B} & FP & 0.648 & 10.89 & 0.648 & 10.89\\
    \hline
    & GPTQ & 0.633 & 17.09 & / & /\\
    & RTN & 0.621 & 11.97 & 0.484 & 34.26\\
    & BnB & \textbf{0.641} & \textbf{11.56} & / & /\\
    & HQQ & 0.638 & 11.67 & 0.373 & 250.72\\
    \rowcolor{gray!20} 
    & WGM & 0.631 & 11.81 & 0.627 & 12.25\\
    \rowcolor{gray!20} 
    & WGM-LO & / & / & \textbf{0.644} & \textbf{11.15}\\
    \hline
    \multirow{7}{3em}{Falcon 3 1B} & FP & 0.598 & 15.49 & 0.598 & 15.49\\
    \hline
    & GPTQ & \textbf{0.596} & \textbf{15.73} & / & /\\
    & RTN & 0.583 & 16.07 & 0.532 & 20.74\\
    & BnB & 0.593 & 15.97 & / & /\\
    & HQQ & 0.588 & 15.99 & 0.554 & 18.72\\
    \rowcolor{gray!20} 
    & WGM & 0.591 & 15.78 & 0.592 & \textbf{16.26}\\
    \rowcolor{gray!20} 
    & WGM-LO & / & / & \textbf{0.595} & 16.91\\
    \hline
    \multirow{7}{3em}{Falcon 3 3B} & FP & 0.626 & 13.50 & 0.626 & 13.50\\
    \hline
    & GPTQ & \textbf{0.617} & \textbf{13.84} & / & /\\
    & RTN & 0.612 & 14.66 & 0.577 & 19.42\\
    & BnB & 0.608 & 14.20 & / & /\\
    & HQQ & 0.611 & 14.28 & 0.511 & 20.00\\
    \rowcolor{gray!20} 
    & WGM & 0.611 & 13.94 & \textbf{0.627} & 13.98\\
    \rowcolor{gray!20} 
    & WGM-LO & / & / & 0.622 & \textbf{13.72}\\
    \hline
    \multirow{7}{3em}{Gemma 3 1b} & FP & 0.585 & 46.10 & 0.585 & 46.10\\
    \hline
    & GPTQ & 0.570 & 59.30 & / & /\\
    & RTN & 0.559 & \textbf{55.60} & 0.528 & 137.74\\
    & BnB & \textbf{0.577} & 71.05 & / & /\\
    & HQQ & 0.575 & 59.23 & 0.506 & 161.76\\
    \rowcolor{gray!20} 
    & WGM & 0.576 & 62.31 & \textbf{0.583} & 52.92\\
    \rowcolor{gray!20} 
    & WGM-LO & / & / & 0.579 & \textbf{48.94}\\
    \hline
    \multirow{7}{3em}{Gemma 3 4b} & FP & 0.692 & 88.86 & 0.692 & 88.86\\
    \hline
    & GPTQ & 0.676 & 105.42 & / & /\\
    & RTN & 0.674 & 112.04 & 0.567 & 253.65\\
    & BnB & 0.682 & 103.32 & / & /\\
    & HQQ & 0.681 & \textbf{81.93} & 0.579 & 334.60\\
    \rowcolor{gray!20} 
    & WGM & \textbf{0.688} & 92.90 & \textbf{0.694} & \textbf{94.76}\\
    \rowcolor{gray!20} 
    & WGM-LO & / & / & 0.692 & 101.29\\
    \hline
  \end{tabular}
  \label{tab:summary}

\end{table}

\subsection{Results}

\subsubsection{Proxy Performance and speed comparison} 

As shown in Table~\ref{tab:higgs-whl}, WGM consistently yields the smallest weight reconstruction MSE when quantizing the first linear weight matrix of LLaMA 3.2 1B, for both per-tensor (4--6 bits) and block-wise (2--4 bits) settings. 
This improved reconstruction compared with baseline is obtained at the expense of higher wall-clock quantization time, with WGM being the slowest method under our CPU implementation. 
At the full-model scale in Table~\ref{tab:result-Q-time}, WGM remains feasible using an 8-core CPU implementation across two processor tiers, whereas baseline timings rely on their standard single-GPU implementations and are therefore not directly comparable in wall-clock time.

\begin{table}[h]
  \centering
  \setlength{\tabcolsep}{1mm}
  \footnotesize 
  \caption{Estimation of LLMs quantization performance through the first linear weight quantization MSE of meta-llama/Llama-3.2-1B following HIGGS. Per tensor quantization}
  \begin{tabular}{c|ccc|ccc} 
    \cmidrule(lr){2-4}\cmidrule(lr){5-7}
    \multicolumn{1}{c}{}&\multicolumn{3}{c}{per tensor} & \multicolumn{3}{c}{block-wise} \\
    \hline
    \textbf{Methods} & \textbf{Bits} & \textbf{Time} & \textbf{MSE} & \textbf{Bits} & \textbf{Time} & \textbf{MSE}\\
    \hline
    RTN & 6 & 0.339 s & 170.425  & 4 & 1.204 s & 69.209\\
    RTN & 5 & 0.252 s & 681.569  & 3 & 1.111 s & 310.216\\
    RTN & 4 & 0.245 s & 2375.157 & 2 & 1.079 s & 1586.022\\
    \hline
    HQQ & 6 & 1.148 s & 147.214  & 4 & 0.966 s & 48.157\\
    HQQ & 5 & 1.079 s & 590.187  & 3 & 0.909 s & 216.235\\
    HQQ & 4 & 1.039 s & 2112.234 & 2 & 0.902 s & 1175.011\\
    \hline \rowcolor{gray!20} 
    WGM & 6 & 15.857 s & 8.325   & 4 & 81.022 s & 33.086\\
    \rowcolor{gray!20} 
    WGM & 5 & 15.533 s & 31.594  & 3 & 74.900 s & 182.880\\
    \rowcolor{gray!20} 
    WGM & 4 & 15.535 s & 127.088 & 2 & 78.493 s & 768.449\\
    \hline
  \end{tabular}
  \label{tab:higgs-whl}
\end{table}

We report weight reconstruction error (Frobenius MSE) as a diagnostic proxy, motivated by recent theory linking layer-wise $\ell_2$ perturbations to perplexity increase in a well-behaved regime~\cite{malinovskii2025higgs}.
However, lower weight MSE alone does not guarantee better downstream quality: prior studies show that aggressively minimizing MSE can still degrade perplexity due to overfitting~\cite{Guo2024GPTQTQL}.
Accordingly, we use QA and PPL as primary evaluation metrics, and treat MSE as a \emph{secondary} signal for numerical fidelity and fast method selection.
In our experiments, MSE tends to correlate with QA/PPL across the compared settings for MSB, making it a useful preliminary proxy when full-model evaluation is costly.

\begin{table}[h]
  \centering
  \setlength{\tabcolsep}{1mm}
  \footnotesize
  \caption{
  Quantization time (seconds). WGM is timed on 8 CPU cores (Xeon Platinum 8480+; Xeon Silver 4410Y in parentheses). Baselines are timed using their standard single-GPU implementations. Differences across the two CPU entries reflect later code optimizations; the optimized version was not rerun on the Platinum system due to access constraints.
  }
  \begin{tabular}{@{} p{8.0em} |ccc>{\columncolor[gray]{0.9}}c@{}} 
    \hline
    \textbf{Model} & \textbf{GPTQ}  & \textbf{BnB} & \textbf{HQQ} & \textbf{WGM}\\
    \hline
    Llama 3.2 1B & 372.9 & 25.6 & 17.7 & 1018.4(1054.5)\\
    \hline
    Llama 3.2 3B & 780.4 & 52.8 & 34.8 & 3555.6(3053.9)\\
    \hline
    Falcon3 1B Instruct & 428.4 & 24.1 & 18.9 & 1377.4(1215.2)\\
    \hline
    Falcon3 3B Instruct & 772.5 & 35.2 & 35.2 & 2974.5(2695.5)\\
    \hline
    Gemma 3 1B it & 399.0 & 19.6 & 13.3 & 777.5(703.8)\\
    \hline
    Gemma 3 4B it & 1000.8 & 65.1 & 47.4 & 3633.9(3461.0)\\
    \hline
  \end{tabular}
  \label{tab:result-Q-time}
\end{table}

\subsubsection{Full QA and PPL Comparison}

Our experimental results in Table \ref{tab:summary} demonstrate that our proposed method is competitive with other widely adopted PTQ methods in block-wise scenario and outperform them in per tensor setting, demonstrating its effectiveness across different quantization granularity.

Across six LLMs, our method remains competitive under both per-tensor and block-wise settings using the same objective and algorithmic pipeline, without introducing setting-specific heuristics. This suggests the objective captures a granularity-agnostic quantization principle rather than overfitting to a particular parameterization.

The QA score aggregates diverse reasoning and commonsense benchmarks (e.g., ARC-e, HellaSwag, WinoGrande, PIQA, among others), so consistent performance indicates robustness across heterogeneous task sensitivities rather than gains driven by a single benchmark. Despite GPTQ is broken for Llama 3.2 1B pretrained model, it is not caused by experiment error, the details of failure is in appendix G.

Notably, these results are obtained under a constrained PTQ pipeline, no calibration data and no auxiliary transformations, isolating the effect of the objective itself. Achieving competitive QA/PPL under these constraints highlights the practicality of the approach when calibration data are unavailable or transformations are undesirable.

Beyond averages, our method exhibits strong cross-model consistency under the block-wise setting: it achieves a top-1 rank in $\text{1}/6$, a top-2 rank in $\text{1}/6$ models for QA and $\text{4}/6$ models for PPL.
Moreover, its worst-case degradation relative to the best baseline is tightly bounded, with at most $\Delta = 0.013$ in QA accuracy and $\Delta = 0.25$ in PPL (excluding Gemma, which exhibits unstable PPL even at full precision).
While being the best model all on task under the per tensor quantization setting.

Finally, we report \textsc{WGM-LO} only for the per-tensor setting with full QA and PPL, since its coarse-to-fine schedule is motivated by large per-tensor instances. We omit \textsc{WGM-LO} for block-wise quantization because WGM already operates with the smallest window and the block instance size is substantially smaller, so the additional coarse initialization step is not aligned with the intended use case. Empirically, \textsc{WGM-LO} wins on 3/6 models in both QA and PPL and is otherwise comparable, suggesting that post-merge local boundary refinement can modestly improve per-tensor quantization quality.

\subsubsection{Approximation algorithm performance.}

To assess how close our heuristic solutions are to the optimum of the proposed objective, we additionally report oracle results for a single matrix in the same setting as Table \ref{tab:summary} as proxy.
\begin{table}[h]
  \centering
  \setlength{\tabcolsep}{1mm}
  \footnotesize 
  \caption{
  Comparison of Estimation of LLMs quantization performance through the first linear weight quantization MSE of meta-llama/Llama-3.2-1B in block-wise setting.
  }
  \begin{tabular}{c|cccc} 
    \hline
    \textbf{Methods} & \textbf{Time 4b} & \textbf{MSE 4b} & \textbf{Time 3b} & \textbf{MSE 3b}\\
    \hline
    DP & 8 hrs 17 mins & 29.96 & 3 hrs 44 mins & 163.17\\
    WGM & 360.47 s & 33.09 & 417.55 s & 182.88 \\
    \hline
  \end{tabular}
  \label{tab:higgs-o}
\end{table}
As shown in Table~\ref{tab:higgs-o}, the exact solver achieves  strictly lower MSE than WGM under identical per-tile bit-width settings (3–4 bits), with improvements of $\Delta 3.13$ and $\Delta 19.71$ for the 3-bit and 4-bit cases, respectively. However, DP requires orders-of-magnitude more computation time (hours versus seconds). 
We therefore use it solely as an oracle reference; it is impractical for full‑model LLM quantization. See Appendix~\ref{app:ext_tab} for details.

\subsubsection{Hyperparameters.}

As indicated in Table \ref{tab:Lambda}, we observe that downstream PPL is largely stable across a broad range of $\lambda$ values; we therefore set $\lambda=0.75$ for all experiments and treat $\lambda$ as a low-sensitivity hyperparameter. 
The rationale for choosing $0.75$ instead of $0$ is detailed in Appendix~\ref{sec:Determining the Boundary of lambda}; full sweeps appear in Appendix~\ref{app:ext_tab}.

\begin{table}
  \centering
  \setlength{\tabcolsep}{1mm}
  \caption{
  Performance comparison for selecting $\lambda$. Utilizing Llama3.2 1B to analyze Perplexity across varying $\lambda$ values with $w$=256 and $g$=256 for efficient experimentation. 
  }
  \begin{tabular}{@{}c|cccccc@{}} 
    \hline
    \textbf{$\lambda$} & 0.0 & 0.2 & 0.4 & 0.6 & 0.8 & 1.0 \\
    \hline
    \textbf{Time} & 340.10 & 385.21 & 377.17 & 370.80 & 364.53 & 364.32\\
    \textbf{Avg.} & 15.06 & 15.09 & 15.09 & 15.11 & 15.10 & 15.10\\
    \hline
  \end{tabular}
  \label{tab:Lambda}
\end{table}

\begin{table}[h]
  \centering
  \setlength{\tabcolsep}{1mm}
  \footnotesize 
  \caption{
  4-bit blockwise quantization MSE and time of weight of first linear of Llama 3.2 1B under different block $t$ and window $w$ size. Detailed table is in Appendix \ref{app:ext_tab}
  }
  \begin{tabular}{@{}c|cccccc@{}} 
    \hline
    $t$/$w$ & \textbf{2048/32} &\textbf{1024/16} & \textbf{512/8} & \textbf{256/4} & \textbf{128/2} & \textbf{64/1}\\
    \hline
    MSE & 72.50 & 71.23 & 67.97 & 63.20 & 53.88 & 33.09\\
    Time & 295.54 & 295.96 & 301.16 & 317.16 & 312.64 & 360.47\\
    \hline
  \end{tabular}
  \label{tab:wVSt-mse}
\end{table}

\begin{table}[h]
  \centering
  \caption{Using Llama3.2 1B to analyze Perplexity performance variation for different bit length, i.e. “max group” $g$, with $w$=256 and different $w$ with $g$=256}
  \begin{subtable}[t]{0.48\linewidth}
    \centering
    \caption{max group}
    \label{tab:groupAndTable_a}
    \begin{tabular}{@{}cc|c@{}} 
    \hline
    \textbf{Bit} & \textbf{Time} &\textbf{Avg.} \\
    \hline
    4 & 429.50 & 5887\\
    5 & 374.81 & 55.17\\
    6 & 371.82 & 15.69\\
    7 & 428.05 & 15.19\\
    8 & 385.05 & 15.13\\
    9 & 393.64 & 15.11\\
    10 & 383.77 & 15.11\\
    \hline
    \end{tabular}
  \end{subtable}\hfill
  \begin{subtable}[t]{0.48\linewidth}
    \centering
    \caption{Window size}
    \label{tab:groupAndTable_b}
    \begin{tabular}{@{}cc|c@{}} 
    \hline
    \textbf{$w$} & \textbf{Time} & \textbf{Avg.} \\
    \hline
    8 & 7606.56 & 13.99\\
    16 & 3753.57 & 14.02\\
    32 & 1900.39 & 14.04\\
    64 & 971.81 & 14.00\\
    128 & 571.83 & 14.26\\
    256 & 393.64 & 15.11\\
    512 & 296.62 & 19.52\\
    \hline
    \end{tabular}
  \end{subtable}
  \label{tab:groupAndTable}
\end{table}

In per tensor setting, the analysis of window size $w$ in Table \ref{tab:groupAndTable} indicates that perplexity degrades noticeably when $w$ exceeds 64, leading us to choose $w=64$ for an optimal balance between perplexity and quantization time. Similarly, Table \ref{tab:groupAndTable} shows that performance saturates around a maximum group size $g=32$, showing WGM is the best at 6-bit per-tensor quantization.

In block-wise setting, table \ref{tab:wVSt-mse} shows MSE decrease as both the window size and block size are reduced, we therefor use 64-row block size and 1 for window size in our block-wise experiments. 
The associated increase in runtime reflects a standard efficiency optimality trade‑off from more fine‑grained merging, but remains feasible in practice, so we adopt the most optimal setting.

\section{Conclusion, Limitations and Future Work}

In this work, we introduced \textsc{MSB PTQ}, a calibration‑free and transformation‑free, binary-inspired formulation for multi-bit post-training quantization, together with a new optimization objective and four tailored solvers. By generalizing binary quantization to multi-bit regime via an unstructured partitioning objective that minimizes within-group variance, our approach offers a simple, implementation-friendly route to low-bit LLM quantization without relying on calibration data or auxiliary transformations.
We demonstrate the effectiveness of our approach on three datasets and various evaluation criteria. The Algorithm \ref{alg:WGM}, WGM performs competitively to representative, commonly used low-bit PTQ quantization in QA tasks and perplexity, despite we have zero calibration and transformation, even no zero point shift. We achieving a top-1 rank in $\text{1}/6$, a top-2 rank in $\text{1}/6$ models for QA and $\text{4}/6$ models for PPL. Moreover, our methods can run on common CPUs, while while other requires GPUs by default. These findings suggest a practical path toward CPU-only PTQ for LLMs and clarify the conceptual link between binary and low-bit quantization.

\textbf{Limitations and future work.}
Our evaluation is primarily simulation-based: we emulate low-bit weights using higher precision arithmetic, and thus do not report end-to-end speedups from specialized kernel implementations. As a result, our runtime analysis focuses on the optimization/quantization procedure itself rather than full inference throughput. The compute cost of the solver also limits the breadth of hyperparameter sweeps and model scales.
Moreover, by construction, \textsc{MSB PTQ} omits calibration and transformations to isolate the contribution of the objective and solvers; while this keeps the approach simple and broadly deployable, it forgoes well-studied PTQ components that can improve performance. These components are orthogonal to our objective and can be integrated without changing the core formulation.
Future works include: (i) developing faster, higher-accuracy approximations to make the optimization scale; (ii) implementing optimized low-bit kernels to enable end-to-end throughput evaluation and practical inference at scale; and (iii) integrating optional calibration and transformation modules on top of MSB PTQ after further solver acceleration.

\section*{Software and Data}

Code: https://github.com/johnnyzheng0636/MSBPTQ

\section*{Acknowledgements}

The core methodology of this work was developed while the author was affiliated with HKUST and later revised during PhD studies at Lingnan University.

\section*{Impact Statement}

This paper presents work whose goal is to advance the field of Machine
Learning. There are many potential societal consequences of our work, none
which we feel must be specifically highlighted here.

\nocite{langley00}

\bibliography{ref}

@misc{grattafiori2024llama3herdmodels,
      title={The Llama 3 Herd of Models},
      author={Dubey, Abhimanyu and Jauhri, Abhinav and Pandey, Abhinav and Kadian, Abhishek and Al-Dahle, Ahmad and Letman, Aiesha and Mathur, Akhil and Schelten, Alan and Yang, Amy and Fan, Angela and others},
      year={2024},
      eprint={2407.21783},
      archivePrefix={arXiv},
      primaryClass={cs.AI},
      url={https://arxiv.org/abs/2407.21783}, 
}

@misc{deepseekai2025deepseekr1incentivizingreasoningcapability,
      title={DeepSeek-R1: Incentivizing Reasoning Capability in LLMs via Reinforcement Learning}, 
      author={Guo, Daya and Yang, Dejian and Zhang, Haowei and Song, Junxiao and Zhang, Ruoyu and Xu, Runxin and Zhu, Qihao and Ma, Shirong and Wang, Peiyi and Bi, Xiao and others},
      year={2025},
      eprint={2501.12948},
      archivePrefix={arXiv},
      primaryClass={cs.CL},
      url={https://arxiv.org/abs/2501.12948}
}

@article{team2025gemma,
  title={Gemma 3 technical report},
  author={Gemma-Team, Gemma and Kamath, Aishwarya and Ferret, Johan and Pathak, Shreya and Vieillard, Nino and Merhej, Ramona and Perrin, Sarah and Matejovicova, Tatiana and Ram{\'e}, Alexandre and Rivi{\`e}re, Morgane and others},
  journal={arXiv preprint arXiv:2503.19786},
  year={2025}
}

@misc{dettmers2022llmint88bitmatrixmultiplication,
      title={LLM.int8(): 8-bit Matrix Multiplication for Transformers at Scale}, 
      author={Tim Dettmers and Mike Lewis and Younes Belkada and Luke Zettlemoyer},
      year={2022},
      eprint={2208.07339},
      archivePrefix={arXiv},
      primaryClass={cs.LG},
      url={https://arxiv.org/abs/2208.07339}, 
}

@misc{wang2023bitnetscaling1bittransformers,
      title={BitNet: Scaling 1-bit Transformers for Large Language Models}, 
      author={Hongyu Wang and Shuming Ma and Li Dong and Shaohan Huang and Huaijie Wang and Lingxiao Ma and Fan Yang and Ruiping Wang and Yi Wu and Furu Wei},
      year={2023},
      eprint={2310.11453},
      archivePrefix={arXiv},
      primaryClass={cs.CL},
      url={https://arxiv.org/abs/2310.11453}, 
}

@misc{frantar-gptq,
  title={{GPTQ}: Accurate Post-training Compression for Generative Pretrained Transformers}, 
  author={Elias Frantar and Saleh Ashkboos and Torsten Hoefler and Dan Alistarh},
  year={2022},
  eprint={2210.17323},
  archivePrefix={arXiv},
  primaryClass={cs.LG},
}

@misc{liu2025spinquantllmquantizationlearned,
      title={SpinQuant: LLM quantization with learned rotations}, 
      author={Zechun Liu and Changsheng Zhao and Igor Fedorov and Bilge Soran and Dhruv Choudhary and Raghuraman Krishnamoorthi and Vikas Chandra and Yuandong Tian and Tijmen Blankevoort},
      year={2025},
      eprint={2405.16406},
      archivePrefix={arXiv},
      primaryClass={cs.LG},
      url={https://arxiv.org/abs/2405.16406}, 
}

@inproceedings{huang2024BiLLMpushinglimitposttraining,
  title={BiLLM: pushing the limit of post-training quantization for LLMs},
  author={Huang, Wei and Liu, Yangdong and Qin, Haotong and Li, Ying and Zhang, Shiming and Liu, Xianglong and Magno, Michele and Qi, Xiaojuan},
  booktitle={Proceedings of the 41st International Conference on Machine Learning},
  pages={20023--20042},
  year={2024}
}

@inproceedings{10.1007/978-3-319-46493-0_32,
author="Rastegari, Mohammad
and Ordonez, Vicente
and Redmon, Joseph
and Farhadi, Ali",
year="2016",
title="XNOR-Net: ImageNet Classification Using Binary Convolutional Neural Networks",
booktitle="Computer Vision -- ECCV 2016",
pages="525--542",
address="Cham",
publisher="Springer International Publishing",
}

@misc{shang2023pb,
  title={Pb-llm: Partially binarized large language models},
  author={Shang, Yuzhang and Yuan, Zhihang and Wu, Qiang and Dong, Zhen},
  year={2023},
  eprint={2310.00034},
  archivePrefix={arXiv},
  primaryClass={cs.LG},
}

@inproceedings{clark-etal-2019-boolq,
    title = "{B}ool{Q}: Exploring the Surprising Difficulty of Natural Yes/No Questions",
    author = "Clark, Christopher  and
      Lee, Kenton  and
      Chang, Ming-Wei  and
      Kwiatkowski, Tom  and
      Collins, Michael  and
      Toutanova, Kristina",
    editor = "Burstein, Jill  and
      Doran, Christy  and
      Solorio, Thamar",
    booktitle = "Proceedings of the 2019 Conference of the North {A}merican Chapter of the Association for Computational Linguistics: Human Language Technologies, Volume 1 (Long and Short Papers)",
    month = jun,
    year = "2019",
    address = "Minneapolis, Minnesota",
    publisher = "Association for Computational Linguistics",
    url = "https://aclanthology.org/N19-1300/",
    doi = "10.18653/v1/N19-1300",
    pages = "2924--2936"
}

@article{Bisk_Zellers_Le_bras_Gao_Choi_2020, title={PIQA: Reasoning about Physical Commonsense in Natural Language}, volume={34}, url={https://ojs.aaai.org/index.php/AAAI/article/view/6239}, DOI={10.1609/aaai.v34i05.6239}, abstractNote={&lt;p&gt;To apply eyeshadow without a brush, should I use a &lt;em&gt;cotton swab or a toothpick&lt;/em&gt;? Questions requiring this kind of &lt;strong&gt;physical commonsense&lt;/strong&gt; pose a challenge to today’s natural language understanding systems. While recent pretrained models (such as BERT) have made progress on question answering over more &lt;em&gt;abstract&lt;/em&gt; domains – such as news articles and encyclopedia entries, where text is plentiful – in more &lt;em&gt;physical&lt;/em&gt; domains, text is inherently limited due to reporting bias. Can AI systems learn to reliably answer physical commonsense questions without experiencing the physical world?&lt;/p&gt;&lt;p&gt;In this paper, we introduce the task of physical commonsense reasoning and a corresponding benchmark dataset &lt;strong&gt;Physical Interaction: Question Answering&lt;/strong&gt; or &lt;strong&gt;PIQA&lt;/strong&gt;. Though humans find the dataset easy (95% accuracy), large pretrained models struggle (∼75%). We provide analysis about the dimensions of knowledge that existing models lack, which offers significant opportunities for future research.&lt;/p&gt;}, number={05}, journal={Proceedings of the AAAI Conference on Artificial Intelligence}, author={Bisk, Yonatan and Zellers, Rowan and Le bras, Ronan and Gao, Jianfeng and Choi, Yejin}, year={2020}, month={Apr.}, pages={7432-7439} }

@inproceedings{zellers-etal-2019-hellaswag,
    title = "{H}ella{S}wag: Can a Machine Really Finish Your Sentence?",
    author = "Zellers, Rowan  and
      Holtzman, Ari  and
      Bisk, Yonatan  and
      Farhadi, Ali  and
      Choi, Yejin",
    editor = "Korhonen, Anna  and
      Traum, David  and
      M{\`a}rquez, Llu{\'i}s",
    booktitle = "Proceedings of the 57th Annual Meeting of the Association for Computational Linguistics",
    month = jul,
    year = "2019",
    address = "Florence, Italy",
    publisher = "Association for Computational Linguistics",
    url = "https://aclanthology.org/P19-1472/",
    doi = "10.18653/v1/P19-1472",
    pages = "4791--4800"
}

@article{10.1145/3474381,
author = {Sakaguchi, Keisuke and Bras, Ronan Le and Bhagavatula, Chandra and Choi, Yejin},
title = {WinoGrande: an adversarial winograd schema challenge at scale},
year = {2021},
issue_date = {September 2021},
publisher = {Association for Computing Machinery},
address = {New York, NY, USA},
volume = {64},
number = {9},
issn = {0001-0782},
url = {https://doi.org/10.1145/3474381},
doi = {10.1145/3474381},
journal = {Commun. ACM},
month = aug,
pages = {99–106},
numpages = {8}
}

@misc{clark2018thinksolvedquestionanswering,
      title={Think you have Solved Question Answering? Try ARC, the AI2 Reasoning Challenge}, 
      author={Peter Clark and Isaac Cowhey and Oren Etzioni and Tushar Khot and Ashish Sabharwal and Carissa Schoenick and Oyvind Tafjord},
      year={2018},
      eprint={1803.05457},
      archivePrefix={arXiv},
      primaryClass={cs.AI},
      url={https://arxiv.org/abs/1803.05457}, 
}

@inproceedings{mihaylov-etal-2018-suit,
    title = "Can a Suit of Armor Conduct Electricity? A New Dataset for Open Book Question Answering",
    author = "Mihaylov, Todor  and
      Clark, Peter  and
      Khot, Tushar  and
      Sabharwal, Ashish",
    editor = "Riloff, Ellen  and
      Chiang, David  and
      Hockenmaier, Julia  and
      Tsujii, Jun{'}ichi",
    booktitle = "Proceedings of the 2018 Conference on Empirical Methods in Natural Language Processing",
    month = oct # "-" # nov,
    year = "2018",
    address = "Brussels, Belgium",
    publisher = "Association for Computational Linguistics",
    url = "https://aclanthology.org/D18-1260/",
    doi = "10.18653/v1/D18-1260",
    pages = "2381--2391"
}

@misc{merity2016pointersentinelmixturemodels,
      title={Pointer Sentinel Mixture Models}, 
      author={Stephen Merity and Caiming Xiong and James Bradbury and Richard Socher},
      year={2016},
      eprint={1609.07843},
      archivePrefix={arXiv},
      primaryClass={cs.CL},
      url={https://arxiv.org/abs/1609.07843}, 
}

@article{JMLR:v21:20-074,
  author  = {Colin Raffel and Noam Shazeer and Adam Roberts and Katherine Lee and Sharan Narang and Michael Matena and Yanqi Zhou and Wei Li and Peter J. Liu},
  title   = {Exploring the Limits of Transfer Learning with a Unified Text-to-Text Transformer},
  journal = {Journal of Machine Learning Research},
  year    = {2020},
  volume  = {21},
  number  = {140},
  pages   = {1--67},
  url     = {http://jmlr.org/papers/v21/20-074.html}
}

@article{marcus-etal-1993-building,
    title = "Building a Large Annotated Corpus of {E}nglish: The {P}enn {T}reebank",
    author = "Marcus, Mitchell P.  and
      Santorini, Beatrice  and
      Marcinkiewicz, Mary Ann",
    editor = "Hirschberg, Julia",
    journal = "Computational Linguistics",
    volume = "19",
    number = "2",
    year = "1993",
    address = "Cambridge, MA",
    publisher = "MIT Press",
    url = "https://aclanthology.org/J93-2004/",
    pages = "313--330"
}

@misc{Falcon3,
    title = {The Falcon 3 Family of Open Models},
    url = {https://huggingface.co/blog/falcon3},
    author = {Falcon-LLM-Team},
    month = {December},
    year = {2024}
}

@article{Guo2024GPTQTQL,
  title={GPTQT: Quantize Large Language Models Twice to Push the Efficiency},
  author={Yipin Guo and Yilin Lang and Qinyuan Ren},
  journal={2024 IEEE International Conference on Cybernetics and Intelligent Systems (CIS) and IEEE International Conference on Robotics, Automation and Mechatronics (RAM)},
  year={2024},
  pages={368-373},
  url={https://api.semanticscholar.org/CorpusID:270924047}
}

@article{10.1109/MM.2023.3256796,
author = {Choquette, Jack},
title = {NVIDIA Hopper H100 GPU: Scaling Performance},
year = {2023},
issue_date = {May-June 2023},
publisher = {IEEE Computer Society Press},
address = {Washington, DC, USA},
volume = {43},
number = {3},
issn = {0272-1732},
url = {https://doi.org/10.1109/MM.2023.3256796},
doi = {10.1109/MM.2023.3256796},
journal = {IEEE Micro},
month = may,
pages = {9–17},
numpages = {9}
}

@inproceedings{tan2024mobilequant,
  title={MobileQuant: Mobile-friendly Quantization for On-device Language Models},
  author={Tan, Fuwen and Lee, Royson and Dudziak, Lukasz and Hu, Shell Xu and Bhattacharya, Sourav and Hospedales, Timothy M and Tzimiropoulos, Georgios and Mart{\'\i}nez, Brais},
  booktitle={EMNLP (Findings)},
  year={2024}
}

@inproceedings{lin2023awq,
	author = {Lin, Ji and Tang, Jiaming and Tang, Haotian and Yang, Shang and Chen, Wei-Ming and Wang, Wei-Chen and Xiao, Guangxuan and Dang, Xingyu and Gan, Chuang and Han, Song},
	booktitle = {Conference on {Machine} {Learning} and {Systems}},
	year = {2023},
	pages = {},
	organization = {},
	title = {AWQ: Activation-aware {Weight} {Quantization} for {On}-{Device} {LLM} {Compression} and {Acceleration}},
	volume = {},
}

@inproceedings{shao2023omniquant,
	author = {Shao, Wenqi and Chen, Mengzhao and Zhang, Zhaoyang and Xu, Peng and Zhao, Lirui and Li, Zhiqiang and Zhang, Kaipeng and Gao, Peng and Qiao, Y. and Luo, Ping},
	booktitle = {International {Conference} on {Learning} {Representations}},
	year = {2023},
	pages = {},
	organization = {},
	title = {OmniQuant: Omnidirectionally {Calibrated} {Quantization} for {Large} {Language} {Models}},
	volume = {abs/2308.13137},
}

@inproceedings{
liu2025spinquant,
title={SpinQuant: {LLM} Quantization with Learned Rotations},
author={Zechun Liu and Changsheng Zhao and Igor Fedorov and Bilge Soran and Dhruv Choudhary and Raghuraman Krishnamoorthi and Vikas Chandra and Yuandong Tian and Tijmen Blankevoort},
booktitle={The Thirteenth International Conference on Learning Representations},
year={2025},
url={https://openreview.net/forum?id=ogO6DGE6FZ}
}

@misc{badri2023hqq,
title  = {Half-Quadratic Quantization of Large Machine Learning Models},
url    = {https://dropbox.github.io/hqq_blog/},
author = {Hicham Badri and Appu Shaji},
month  = {November},
year   = {2023}
}

@inproceedings{malinovskii2025higgs,
	author = {Malinovskii, Vladimir and Panferov, Andrei and Ilin, Ivan and Guo, Han and Richt{\' a}rik, Peter and Alistarh, Dan},
	booktitle = {North {American} {Chapter} of the {Association} for {Computational} {Linguistics}},
	year = {2025},
	pages = {10857--10886},
	organization = {},
	title = {HIGGS: Pushing the {Limits} of {Large} {Language} {Model} {Quantization} via the {Linearity} {Theorem}},
	volume = {},
}

@article{dettmers2023qlora,
  title={Qlora: Efficient finetuning of quantized llms},
  author={Dettmers, Tim and Pagnoni, Artidoro and Holtzman, Ari and Zettlemoyer, Luke},
  journal={arXiv preprint arXiv:2305.14314},
  year={2023}
}
\bibliographystyle{icml2026}

\newpage
\appendix
\onecolumn

\section{Detailed Derivation of Proposed Methods}
\label{sec:Detailed Derivation of Proposed Methods}

Given the objective
\begin{align*}
    g^{\ast},\{\boldsymbol{A}_i^{\ast}\}^{g^{\ast}}_{i=1} = 
    \argmin_{\substack{
        \{\boldsymbol{A}_i^{\ast}\}^{g^{\ast}}_{i=1} \in \mathbb{A}\\
        \{\boldsymbol{B}_i^{\ast}\}^{g^{\ast}}_{i=1} \in \mathbb{B}\\
        1 \leq g \leq |\boldsymbol{A}|
    }} \sum^g_{i=1}||\boldsymbol{A}_i - \alpha_i\boldsymbol{B}_i||_2^2 
    + \frac{1}{|\boldsymbol{B}_i|}
\end{align*}
Since $|\boldsymbol{A}_i| = |\boldsymbol{B}_i|$, the optimization objective can be rewritten to
\begin{align*}
    g^{\ast},\{\boldsymbol{A}_i^{\ast}\}^{g^{\ast}}_{i=1} = 
    \argmin_{\substack{
        \{\boldsymbol{A}_i^{\ast}\}^{g^{\ast}}_{i=1} \in \mathbb{A}\\
        \{\boldsymbol{B}_i^{\ast}\}^{g^{\ast}}_{i=1} \in \mathbb{B}\\
        1 \leq g \leq |\boldsymbol{A}|
    }} \sum^g_{i=1}||\boldsymbol{A}_i - \alpha_i\boldsymbol{B}_i||_2^2 
    + \frac{1}{|\boldsymbol{A}_i|}
\end{align*}
As shown in \ref{eq:xnor_obj},
\begin{align*}
    &\argmin_{\alpha^{\ast}, \boldsymbol{B}^{\ast}} ||\boldsymbol{A} - \alpha\boldsymbol{B}||_2^2, \quad \alpha^{\ast} = \frac{||\boldsymbol{A}||_1}{|\boldsymbol{A}|}, \; \boldsymbol{B}^{\ast} = \text{sign}(\boldsymbol{A})
\end{align*}
By substituting the optimal values, we can rewrite the squared 2-norm.
\begin{align*}
    ||\boldsymbol{A} - \alpha^{\ast}\boldsymbol{B}^{\ast}||_2^2 &= \sum^i\sum^j(A_{i,j} - \alpha^{\ast} \cdot \text{sign}(A_{i,j}))^2\\
    &= \sum^i\sum^j(|A_{i,j}| - \alpha^{\ast})^2\\
    &= \sum^i\sum^j |A_{i,j}|^2 - 2\alpha^{\ast}\sum^i\sum^j |A_{i,j}| + \sum^i\sum^j {\alpha^{\ast}}^2\\
    &= ||\boldsymbol{A}||_2^2 - 2\alpha^{\ast} ||\boldsymbol{A}||_1 + |\boldsymbol{A}|{\alpha^{\ast}}^2\\
    &= ||\boldsymbol{A}||_2^2 - 2 \frac{||\boldsymbol{A}||_1}{|\boldsymbol{A}|} ||\boldsymbol{A}||_1 + |\boldsymbol{A}| \left(\frac{||\boldsymbol{A}||_1}{|\boldsymbol{A}|}\right)^2\\
    &= ||\boldsymbol{A}||_2^2 - 2 \frac{||\boldsymbol{A}||_1^2}{|\boldsymbol{A}|} + \frac{||\boldsymbol{A}||_1^2}{|\boldsymbol{A}|}\\
    &= ||\boldsymbol{A}||_2^2 - \frac{||\boldsymbol{A}||_1^2}{|\boldsymbol{A}|}
\end{align*}
Consider the variance of absolute value of $\boldsymbol{A}_i$, let $\Tilde{\boldsymbol{A}_i}$ be $\boldsymbol{A}_i$ with element-wise absolute value, such as $|\Tilde{\boldsymbol{A}_i}| = |\boldsymbol{A}_i|$ and $\Tilde{A}_{i_{j,k}} = |A_{i_{j,k}}|$.
\begin{align*}
    Var(\Tilde{\boldsymbol{A}_i}) &= E[\Tilde{\boldsymbol{A}_i}^2] - E[\Tilde{\boldsymbol{A}_i}]^2\\
    &= \frac{1}{|\boldsymbol{A}_i|}\sum^j\sum^k |A_{i_{j,k}}|^2 - \left(\frac{1}{|\boldsymbol{A}_i|}\sum^j\sum^k |A_{i_{j,k}}|\right)^2\\
    &= \frac{||\boldsymbol{A}_i||_2^2}{|\boldsymbol{A}_i|} - \left( \frac{||\boldsymbol{A}_i||_1}{|\boldsymbol{A}_i|} \right)^2\\
    \\
    |\boldsymbol{A}_i|Var(\Tilde{\boldsymbol{A}_i}) &= ||\boldsymbol{A}_i||_2^2 - \frac{||\boldsymbol{A}_i||_1^2}{|\boldsymbol{A}_i|}
\end{align*}
Together, we can conclude
\begin{align*}
    ||\boldsymbol{A}_i - \alpha_i^{\ast}\boldsymbol{B}_i^{\ast}||_2^2 = |\boldsymbol{A}_i|Var(\Tilde{\boldsymbol{A}_i})
\end{align*}
Finally, we rewrite the initial optimization objective as
\begin{gather*}
    g^{\ast}, \{\boldsymbol{A}_i^{\ast}\}^{g^{\ast}}_{i=1}
    = \argmin_{\substack{
        \{\boldsymbol{A}_i^{\ast}\}^{g^{\ast}}_{i=1} \in \mathbb{A}\\
        1 \leq g \leq |\boldsymbol{A}|
    }} \sum^g_{i=1}|\boldsymbol{A}_i|\text{Var}(\Tilde{\boldsymbol{A}_i}) + \frac{1}{|\boldsymbol{A}_i|} 
\end{gather*}

\section{Remaining Three Algorithms}
\label{sec:3AlgoPseudocode}

In this section, we present the pseudocode for the remaining three of our designed algorithms:

\begin{itemize}
\item Algorithms \ref{alg:DG}: Dynamic Grouping.
\item Algorithms \ref{alg:GG}: Greedy Grouping. 
\item Algorithms \ref{alg:LOWGM}: Local optimizing Windowed Greedy Grouping. 
\end{itemize}

\renewcommand{\thealgorithm}{1}
\begin{algorithm}
\caption{Dynamic Grouping}
\label{alg:DG}
\small
\begin{algorithmic}[1]
\STATE {\bfseries Sort: }{$\boldsymbol{A}$}
\STATE sort the absolute value of input matrix $\boldsymbol{A}$
\STATE \textbf{Return} Sorted array of value and index in $\boldsymbol{A}$
\newline

\STATE {\bfseries Prefix sum: }{Array}
\STATE prefix\_sum, squared\_prefix\_sum $\gets$ empty array
\FOR{$i$ in Array}
\STATE prefix\_sum $\gets$ [prefix\_sum, $i$]
\STATE squared\_prefix\_sum $\gets$ [squared\_prefix\_sum, $i^2$]
\ENDFOR
\STATE \textbf{Return} prefix\_sum, squared\_prefix\_sum
\newline

\STATE {\bfseries Cost: }{$j$, $k$, prefix\_sum, squared\_prefix\_sum}
\STATE group\_size $\gets k-j$ 
\STATE abs\_sum $\gets$ prefix\_sum[$k$] - prefix\_sum[$j$]
\STATE abs\_sum\_sqr$\gets$squared\_prefix\_sum[$k$]-squared\_prefix\_sum[$j$]
\STATE $\mu \gets \frac{\text{abs\_sum}}{\text{group\_size}}$
\STATE $\sigma^2 \gets \frac{\text{abs\_sum\_sqr}}{\text{group\_size}} - \mu^2$
\STATE cost $\gets$ group\_size$\sigma^2$ + $\frac{1}{\text{group\_size}}$
\STATE \textbf{Return} cost, $\mu$
\newline

\STATE {\bfseries Dynamic Grouping: }{$\boldsymbol{A}\in\mathbb{R}^{m \times n}$}
\STATE $dp \gets dp_0, \boldsymbol{j^{\ast}}, \boldsymbol{\mu} \gets \emptyset \in \mathbb{R}^{m \times n} $
\STATE $\boldsymbol{a, i} \gets \textbf{Sort}(\boldsymbol{A})$
\STATE $p_1, p_2 \gets \textbf{Prefix sum}(\boldsymbol{a})$
\FOR{$i$ in 1 to $mn$}
    \FOR{$j$ in 1 to $mn$}
        \STATE $c_{min} = \inf, j^{\ast} = -1, \mu^{\ast} = -1$
        \FOR{$k$ in 0 to $j-1$}
            \IF{$k$ in $\boldsymbol{j}^{\ast}[:i][k-1]$}
                \STATE continue
            \ENDIF
            \STATE $c, \mu \gets \textbf{Cost}(k, j, p_1, p_2)$
            \STATE $c \gets c + dp[i-1][k]$
            \IF{$c < c_{min}$}
                \STATE $c_{min} \gets c, j^{\ast} = j, \mu^{\ast} = \mu$
            \ENDIF
        \ENDFOR
        \STATE $dp[i][j] \gets c_{min}, \boldsymbol{j^{\ast}}[i][j] \gets j^{\ast}, \boldsymbol{\mu}[i][j] \gets \mu^{\ast}$
    \ENDFOR
\ENDFOR
\STATE \textbf{Return} $dp, \boldsymbol{j^{\ast}}, \boldsymbol{\mu}$
\end{algorithmic}
\end{algorithm}

\renewcommand{\thealgorithm}{2}
\begin{algorithm}
\caption{Greedy Grouping}
\label{alg:GG}
\begin{algorithmic}[1]
\STATE {\bfseries Greedy merging: }{$\boldsymbol{A}\in\mathbb{R}^{m \times n}, g\text{ (max group})$}
\STATE sort absolute value of non-zero entry into an array
\STATE initial the merge cost array $[(i \text{ (group start}), i+1 \text{( group end}), cost, \mu)]$ for $i$ in range($mn$ - 1)
\STATE initially the ignore array
\STATE heapify the merge cost array
\WHILE{len(merge cost array) $>$ g}
\STATE currentMerge $\gets$ heap pop
\IF{currentMerge in ignore array}
\STATE remove currentMerge from ignore array
\STATE continue
\ENDIF
\STATE push two new merges as updates
\STATE update ignore array by two merged, invalidated by the two new merges
\ENDWHILE
\STATE \textbf{return} merge cost array
\end{algorithmic}
\end{algorithm}

\renewcommand{\thealgorithm}{4}
\begin{algorithm}
\caption{Local Optimizing Windowed Greedy Merging}
\label{alg:LOWGM}
\begin{algorithmic}[1]
\STATE {\bfseries LOWGM: }{$\boldsymbol{A}\in\mathbb{R}^{m \times n}, g\text{ (max group}), k\text{ (window number}), b\text{ (local optimization range}), T \text{ (max iterations)}, \epsilon \text{ (converge threshold)}$}
\STATE sort absolute value of non-zero entry into an array
\STATE initial the merge cost array with equal range binning, bin width $\Delta=(a_{max}-a_{min})/k$ and map each magnitude $|a|$ to bin index $=\min(k-1,\lfloor(|a|-a_{min})/\Delta\rfloor)$
\STATE initially the ignore array
\STATE heapify the merge cost array
\WHILE{len(merge cost array) $>$ g}
\STATE currentMerge $\gets$ heapop
\IF{currentMerge in ignore array}
\STATE remove currentMerge from ignore array
\STATE continue
\ENDIF
\STATE push two new neighbouring merges as updates
\STATE update ignore array by two old merges, invalidated by the two new merges
\ENDWHILE
\STATE $i = 0, \varepsilon = \inf$
\WHILE{$\varepsilon > \epsilon$ or $i \geq T$}
\FOR{each group}
\STATE Try the move group boundary randomly within $b$ one time. If the total loss is lower, update the group boundary and $\varepsilon$.
\STATE $i++$
\ENDFOR
\ENDWHILE
\STATE \textbf{return} merge cost array
\end{algorithmic}
\end{algorithm}

\section{Determining the Boundary of \texorpdfstring{$\lambda$}{lambda}}
\label{sec:Determining the Boundary of lambda}
We further analysis the boundary of $\lambda$, the objective in this section is 

\begin{align*}
    \min_{g \in \mathbb{N}}  \sum^g_{i=1} \left( \textstyle \frac{|\boldsymbol{A}_i|}{|\boldsymbol{A}|} Var(\Tilde{\boldsymbol{A}_i}) + \textstyle \frac{\lambda}{|\boldsymbol{A}_i|} \right) \label{eq:lambdaObj} \tag{4}
\end{align*}

Firstly, we determine the lower bound of $\lambda$. Considering the case of group size 1 and number of groups $g$ is the number of non-zero elements $n$. The cost is 

\begin{align*}
    \sum^{g}0 + \lambda \frac{1}{1}=n\lambda \label{lowerBound1}\tag{5} 
\end{align*}

The squared 2 norm is 0 since in the case of group size one $\boldsymbol{A}_i = \alpha_i\boldsymbol{B}_i$.

Now consider the case of the number of groups = $g-1$, so we have $g-2$ groups of size 1 and 1 group of size 2. The cost is

\begin{align*}
    &\sum^{g-2} \left(0 + \lambda \frac{1}{1}\right) + \frac{1}{n}(|a_1| - \alpha^{\ast})^2 + \frac{1}{n}(|a_2| - \alpha^{\ast})^2 + \frac{\lambda}{2}\\
    =&\sum^{n-2} \left(0 + \lambda \frac{1}{1}\right) + \frac{1}{n}(|a_1| - \frac{|a_1| + |a_2|}{2})^2 \\ & + \frac{1}{n}(|a_2| - \frac{|a_1| + |a_2|}{2})^2 + \frac{\lambda}{2}\\
    =&(n-2)\lambda + \frac{1}{n}(\frac{|a_1| - |a_2|}{2})^2 + \frac{1}{n}(\frac{|a_2| - |a_1|}{2})^2 + \frac{\lambda}{2}\\
    =&(n-2)\lambda + \frac{1}{n}(\frac{|a_1| - |a_2|}{2})^2 + \frac{1}{n}(\frac{|a_1| - |a_2|}{2})^2 + \frac{\lambda}{2}\\
    =&(n-2)\lambda + \frac{(|a_1| - |a_2|)^2}{2n} + \frac{\lambda}{2} && \tag{6} \label{lowerBound2}\\
\end{align*}

Since we are minimizing the objective of equation \ref{eq:lambdaObj}, when (\ref{lowerBound1}) $<$ (\ref{lowerBound2}) we have n groups, else at most n-1 groups. Because we are using quantization for higher efficiency, a smaller number of groups, each with a larger size is desired. Therefore, we must avoid the n group case, i.e.(\ref{lowerBound1}) $>$ (\ref{lowerBound2}).

\begin{align*}
    n\lambda &> (n-2)\lambda + \frac{(|a_1| - |a_2|)^2}{2n} + \frac{\lambda}{2}\\
    2\lambda &> \frac{(|a_1| - |a_2|)^2}{2n} + \frac{\lambda}{2}\\
    \frac{3\lambda}{2} &> \frac{(|a_1| - |a_2|)^2}{2n}\\
    \lambda &> \frac{(|a_1| - |a_2|)^2}{3n}\\
    \lambda &> \min_{1 \leq i < n}\frac{(|a_i| - |a_{i+1}|)^2}{3n}
\end{align*}

Since we also want to avoid the n-1 group case, and any group number that leads to a small group size, in general. So a good $\lambda$ will not be in the close neighbourhood of the actual lower bound, thus we can estimate the lower bound roughly as

\begin{align*}
    \lambda > \frac{(|a_1| - |a_2|)^2}{3n} > \min_{1 \leq i < n}\frac{(|a_i| - |a_{i+1}|)^2}{3n}
\end{align*}

We can just use the first two elements in the sorted array for fast computation and estimation of the lower bound of $\lambda$, so we have a lower bound

\begin{align*}
    \lambda > \frac{(|a_1| - |a_2|)^2}{3n} \tag{7} \label{lowerbound}
\end{align*}

Next, we deduce the upper bound for $\lambda$ with a similar method. Considering the case of one group, the cost is 

\begin{align*}
    &\frac{1}{|\boldsymbol{A}|}|\boldsymbol{A}|Var(\Tilde{\boldsymbol{A}}) + \frac{\lambda}{|\boldsymbol{A}|}\\
    =&\frac{1}{n}nVar(\Tilde{\boldsymbol{A}}) + \frac{\lambda}{n}\\
    =&Var(\Tilde{\boldsymbol{A}}) + \frac{\lambda}{n} && \tag{8} \label{upperBound1}
\end{align*}

Considering two group of matrix $\boldsymbol{A}_1, \boldsymbol{A}_2$ split at $k^{\text{th}}$ sorted non-zero element of $\boldsymbol{A}$, the cost is 

\begin{align*}
    &\frac{1}{|\boldsymbol{A}|}|\boldsymbol{A}_1|Var(\Tilde{\boldsymbol{A}_1}) + \frac{\lambda}{|\boldsymbol{A}_1|} + \frac{1}{|\boldsymbol{A}|}|\boldsymbol{A}_2|Var(\Tilde{\boldsymbol{A}_2}) + \frac{\lambda}{|\boldsymbol{A}_2|}\\
    =&\frac{1}{n}\left(kVar(\Tilde{\boldsymbol{A}_1}) + (n-k)Var(\Tilde{\boldsymbol{A}_2})\right) + \frac{\lambda}{k} + \frac{\lambda}{n-k} && 
    \tag{9} \label{upperBound2}
\end{align*}

If we want to have at least 2 group, (\ref{upperBound2}) $<$ (\ref{upperBound1}),

\begin{align*}
    & \frac{1}{n} \left( k \, \text{Var}(\Tilde{\boldsymbol{A}_1}) + (n-k) \, \text{Var}(\Tilde{\boldsymbol{A}_2}) \right) + \frac{\lambda}{k} + \frac{\lambda}{n-k} \nonumber < \text{Var}(\Tilde{\boldsymbol{A}}) + \frac{\lambda}{n} \\
    & k \, \text{Var}(\Tilde{\boldsymbol{A}_1}) + (n-k) \, \text{Var}(\Tilde{\boldsymbol{A}_2}) + \frac{n\lambda}{k} + \frac{n\lambda}{n-k} \nonumber< n \, \text{Var}(\Tilde{\boldsymbol{A}}) + \lambda \\
    & \frac{n\lambda}{k} + \frac{n\lambda}{n-k} - \lambda \nonumber< n \, \text{Var}(\Tilde{\boldsymbol{A}}) - \left( k \, \text{Var}(\Tilde{\boldsymbol{A}_1}) + (n-k) \, \text{Var}(\Tilde{\boldsymbol{A}_2}) \right) \\
    & \lambda \left( \frac{n}{k} + \frac{n}{n-k} - 1 \right) \nonumber < n \, \text{Var}(\Tilde{\boldsymbol{A}}) - \left( k \, \text{Var}(\Tilde{\boldsymbol{A}_1}) + (n-k) \, \text{Var}(\Tilde{\boldsymbol{A}_2}) \right) \\
    & \lambda \nonumber < \frac{n \, \text{Var}(\Tilde{\boldsymbol{A}}) - \left( k \, \text{Var}(\Tilde{\boldsymbol{A}_1}) + (n-k) \, \text{Var}(\Tilde{\boldsymbol{A}_2}) \right)}{\frac{n}{k} + \frac{n}{n-k} - 1}
\end{align*}

Notice we can simplify the above expression by rewriting $nVar(\Tilde{\boldsymbol{A}})$

\begin{align*}
    nVar(\Tilde{\boldsymbol{A}}) = \sum_{x \in \Tilde{\boldsymbol{A}}} (x - \mu)^2 = \sum_{x_1 \in \Tilde{\boldsymbol{A}_1}} (x_1 - \mu)^2 + \sum_{x_2 \in \Tilde{\boldsymbol{A}_2}} (x_2 - \mu)^2 
\end{align*}

Where $\mu$ is the average of elements of $\Tilde{\boldsymbol{A}}$. Notice

\begin{align*}
    & \sum_{x_1 \in \Tilde{\boldsymbol{A}_1}} (x_1 - \mu)^2 = \sum_{x_1 \in \Tilde{\boldsymbol{A}_1}} ((x_1 - \mu_1) + (\mu_1 - \mu))^2\\
    = &\sum_{x_1 \in \Tilde{\boldsymbol{A}_1}} (x_1 - \mu_1)^2 + 2(x_1 - \mu_1)(\mu_1 - \mu) + (\mu_1 - \mu)^2\\
    = &\left(\sum_{x_1 \in \Tilde{\boldsymbol{A}_1}} \left((x_1 - \mu_1)^2 + (\mu_1 - \mu)^2\right)\right)+\\
    &\left(2(\mu_1 - \mu)\sum_{x_1 \in \Tilde{\boldsymbol{A}_1}} (x_1 - \mu_1)\right)\\
    = &\left(\sum_{x_1 \in \Tilde{\boldsymbol{A}_1}} \left((x_1 - \mu_1)^2 + (\mu_1 - \mu)^2\right)\right) + \\
    &\left(2(\mu_1 - \mu)\left(\left(\sum_{x_1 \in \Tilde{\boldsymbol{A}_1}} x_1 \right)- |\boldsymbol{A}_1|\mu_1\right)\right)\\
    = &\left(\sum_{x_1 \in \Tilde{\boldsymbol{A}_1}} \left((x_1 - \mu_1)^2 + (\mu_1 - \mu)^2\right)\right) + \left(2(\mu_1 - \mu)(0)\right)\\
    = &\left(\sum_{x_1 \in \Tilde{\boldsymbol{A}_1}} \left((x_1 - \mu_1)^2 + (\mu_1 - \mu)^2\right)\right)\\
    = &\sum_{x_1 \in \Tilde{\boldsymbol{A}_1}} (x_1 - \mu_1)^2 + k(\mu_1 - \mu)^2
\end{align*}

Similarly

\begin{align*}
    \sum_{x_2 \in \Tilde{\boldsymbol{A}_2}} (x_2 - \mu)^2 &= \sum_{x_2 \in \Tilde{\boldsymbol{A}_2}} \left((x_2 - \mu_2)^2 + (\mu_2 - \mu)^2\right)\\
    &= \sum_{x_2 \in \Tilde{\boldsymbol{A}_2}} (x_2 - \mu_2)^2 + (n-k)(\mu_2 - \mu)^2 
\end{align*}

Substitute back

\begin{align*}
    &nVar(\Tilde{\boldsymbol{A}})\\
    =& \sum_{x_1 \in \Tilde{\boldsymbol{A}_1}} (x_1 - \mu_1)^2 + k(\mu_1 - \mu)^2 + \\
    &\sum_{x_2 \in \Tilde{\boldsymbol{A}_2}} (x_2 - \mu_2)^2 + (n-k)(\mu_2 - \mu)^2 \\
    =& kVar(\Tilde{\boldsymbol{A}_1}) + (n-k)Var(\Tilde{\boldsymbol{A}_2}) + k(\mu_1 - \mu)^2 + (n-k)(\mu_2 - \mu)^2
\end{align*}

Substituting this into the upper bound, we get

\begin{align*}
    \lambda &< \frac{n Var(\Tilde{\boldsymbol{A}}) - \left(k Var(\Tilde{\boldsymbol{A}_1}) + (n-k)Var(\Tilde{\boldsymbol{A}_2})\right)}{\frac{n}{k} + \frac{n}{n-k} - 1}\\
    &= \frac{k(\mu_1 - \mu)^2 + (n-k)(\mu_2 - \mu)^2}{\frac{n}{k} + \frac{n}{n-k} - 1}\\
    &<  \max_{1 \leq k < n} \frac{k(\mu_1 - \mu)^2 + (n-k)(\mu_2 - \mu)^2}{\frac{n}{k} + \frac{n}{n-k} - 1}
\end{align*}

We can further simplify the numerator.

\begin{align*}
    &k(\mu_1 - \mu)^2 + (n-k)(\mu_2 - \mu)^2\\ 
    =& k(\mu_1 - \frac{k\mu_1 + (n-k)\mu_2}{n})^2 + (n-k)(\mu_2 - \frac{k\mu_1 + (n-k)\mu_2}{n})^2\\
    =& k(\frac{n\mu_1 - k\mu_1 - (n-k)\mu_2}{n})^2 + (n-k)(\frac{n\mu_2 - k\mu_1 - (n-k)\mu_2}{n})^2\\
    =& k(\frac{(n-k)\mu_1 - (n-k)\mu_2}{n})^2 + (n-k)(\frac{ - k\mu_1 + k\mu_2}{n})^2\\
    =& k(\frac{(n-k)(\mu_1 - \mu_2)}{n})^2 + (n-k)(\frac{ - k(\mu_1 - \mu_2)}{n})^2\\
    =& \frac{k(n-k)^2(\mu_1 - \mu_2)^2 + (n-k)k^2(\mu_1 - \mu_2)^2}{n^2}\\
    =& \frac{kn - k^2 + k^2}{n^2}(n-k)(\mu_1 - \mu_2)^2\\
    =& \frac{k}{n}(n-k)(\mu_1 - \mu_2)^2
\end{align*}

So the upper bound is 

\begin{align*}
    \lambda < \max_{1 \leq k < n} \frac{\frac{k}{n}(n-k)(\mu_1 - \mu_2)^2}{\frac{n}{k} + \frac{n}{n-k} - 1}
\end{align*}

To balance efficiency and accuracy, we avoid approaching the precise upper bound, as it maximizes efficiency at the cost of minimal accuracy. Instead, a faster approximation is employed to estimate the upper bound. Using $k = \frac{n}{2}$
\begin{align*}
    \lambda &< \frac{\frac{\frac{n}{2}}{n}(n-\frac{n}{2})(\mu_1 - \mu_2)^2}{\frac{n}{\frac{n}{2}} + \frac{n}{n-\frac{n}{2}} - 1} < \max_{1 \leq k < n} \frac{\frac{k}{n}(n-k)(\mu_1 - \mu_2)^2}{\frac{n}{k} + \frac{n}{n-k} - 1}\\
    \lambda &< \frac{\frac{1}{2}(\frac{n}{2})(\mu_1 - \mu_2)^2}{2 + \frac{n}{\frac{n}{2}} - 1}\\
    &= \frac{\frac{n}{4}(\mu_1 - \mu_2)^2}{3}\\
    &= \frac{n(\mu_1 - \mu_2)^2}{12}
\end{align*}

Finally, we have 

\begin{align*}
    \lambda_{min} \approx \frac{(|a_1| - |a_2|)^2}{3n} <\lambda < \frac{n(\mu_1 - \mu_2)^2}{12} \approx \lambda_{max}\tag{10} \label{bound}
\end{align*}

And we can control $\lambda$ through $\Tilde{\lambda} \in [0,1]$ by 

\begin{align*}
    \lambda &= \Lambda(\Tilde{\lambda})\\
    &= \lambda_{min} + \Tilde{\lambda} (\lambda_{max} - \lambda_{min}) \\
    &= \frac{(|a_1| - |a_2|)^2}{3n} + \Tilde{\lambda}(\frac{n(\mu_1 - \mu_2)^2}{12} - \frac{(|a_1| - |a_2|)^2}{3n})
\end{align*}

For the optimal $\lambda$, the hypothesis is that a suitable value of $\lambda$ should lie within the latter half of the range, but not too close to the upper bound, which prioritizes efficiency over accuracy. A precise choice of $\lambda$ leads to a specific optimal max group, which can be identified via binary search over the result max group of Algorithm \ref{alg:DG}. Since this process is too computationally intensive, we hypothesize that $\Tilde{\lambda}^{\ast} \approx 0.75$ provides a reasonable approximation.
Empirical results on Algorithm \ref{alg:GG} and \ref{alg:WGM}, however, indicate that $\lambda$ has negligible impact on quantization performance  of PPL. The same holds for $\Tilde{\lambda}$, and the corresponding table is omitted, as it is virtually identical to Table \ref{tab:Lambda}. This observation is theoretically supported: $\lambda$ is originally introduced to determine the optimal number of groups in Algorithm \ref{alg:DG}, whereas in other algorithms, the number of groups is treated as a user-defined hyperparameter, rendering it inapplicable. For consistency of notation and to preserve the theoretical formulation of Algorithm~\ref{alg:DG}, we keep the $\lambda$ term throughout.

\section{Supplementary Experiments}
\label{sec:Supplementary Experiments, Settings, and Full Results}

\subsection{Additional experiments setup}

Since the main text shows that weight MSE is proportional with downstream quantization quality, we use MSE as a proxy in controlled ablations. we evaluate the proposed algorithms on synthetic matrices sampled from $\mathcal{N}(0,1)$ and run all experiments on a single CPU. These standardized test cases allow us to compare quantization loss, runtime, and hyperparameter sensitivity using MSE as a proxy.

\subsection{Comparison for quantization loss}

We evaluate the three core variants (Algorithms~\ref{alg:DG}, \ref{alg:GG}, and \ref{alg:WGM}) by proxy MSE on synthetic square matrices. Figures~\ref{fig:loss-s} and \ref{fig:loss-l} compare against \textsc{XNOR}, \textsc{Blocked-XNOR}, and an all-zero dummy baseline. As expected, the all-zero dummy yields the largest error, while \textsc{XNOR} and \textsc{Blocked-XNOR} produce moderate MSE. In contrast, our methods consistently reduce MSE to near zero on these controlled instances.

In Figure~\ref{fig:loss-s}, Algorithm~\ref{alg:WGM} eventually matches \textsc{XNOR}. This is an implementation artifact of the dynamic window schedule: when the window size grows to $n$, the matrix dimension, windowed merging degenerates to standard \textsc{XNOR}. For sufficiently large inputs, the code increases the window automatically, causing this convergence behavior.

In Figure~\ref{fig:loss-l}, we omit Algorithm~\ref{alg:DG} due to prohibitive runtime at larger sizes. We evaluate square matrices with side length $n\in\{2,4,\dots,2048\}$. As $n$ grows, the loss curves become smoother analogs of the small-scale behavior, suggesting that the qualitative trends generalize. Moreover, Algorithms~\ref{alg:GG} and \ref{alg:WGM} closely track the optimum provided by Algorithm~\ref{alg:DG}, indicating that their approximation gap is negligible in terms of MSE on these instances. The relative ordering of errors typically follows our expectation (DG $\le$ GG $\le$ WGM), but the absolute differences remain small.

We omit Algorithm~4 (\textsc{WGM-LO}) from this proxy ablation because the goal here is to benchmark heuristic grouping strategies against the dynamic-programming oracle (Algorithm~\ref{alg:DG}). Algorithm~\ref{alg:WGM} already achieves MSE nearly indistinguishable from the oracle on these controlled instances, establishing that its approximation gap is negligible in the proxy regime. In contrast, \textsc{WGM-LO} is a local refinement applied on top of Algorithm~\ref{alg:WGM} and it is feasible to run to completion on LLMs, we therefore evaluate \textsc{WGM-LO} directly against Algorithm~\ref{alg:WGM} using full PPL/QA rather than the MSE proxy.

\begin{figure}
\centering
\begin{minipage}{.5\textwidth}
  \centering
  \includegraphics[width=1.0\linewidth]{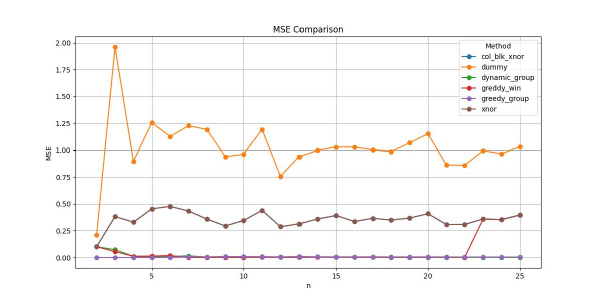}
  \captionsetup{width=0.95\linewidth}
  \caption{Algorithm comparison for small matrix, quantization loss against matrix sized $\mathbb{R}^{n \times n}$.}
  \label{fig:loss-s}
\end{minipage}%
\begin{minipage}{.5\textwidth}
  \centering
  \includegraphics[width=1.0\linewidth]{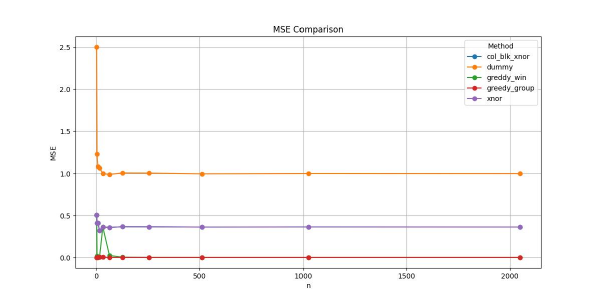}
  \captionsetup{width=0.95\linewidth}
  \caption{Algorithm comparison for large matrix, quantization loss against matrix sized $\mathbb{R}^{n \times n}$.}
  \label{fig:loss-l}
\end{minipage}
\end{figure}

\subsection{Comparison for quantization speed}

Following the setup in the previous section, we replace the MSE proxy with wall-clock quantization time and compare runtime on both small and large square matrices. In Figure~\ref{fig:speed-s}, \textsc{XNOR} and \textsc{Blocked-XNOR} are the fastest baselines, as they rely on lightweight mean/sign computations. Among our methods, Algorithms~\ref{alg:GG} and \ref{alg:WGM} are substantially faster than the oracle Algorithm~\ref{alg:DG}, consistent with the fact that Algorithm~\ref{alg:DG} trades increased computational complexity for optimality.

The abrupt runtime change observed for Algorithm~\ref{alg:WGM} is due to its dynamic window schedule: once the window size grows to the matrix dimension $n$, the method degenerates to standard \textsc{XNOR}, leading to a sudden speed-up.

As the matrix size increases, runtime grows rapidly and Algorithm~\ref{alg:DG} becomes impractical. Figure~\ref{fig:speed-l} reports the same comparison on larger matrices. Algorithm~\ref{alg:GG} can require on the order of $10^3$ seconds at the largest size, whereas Algorithm~\ref{alg:WGM} completes within $\sim 10$ seconds, i.e., over two orders of magnitude faster. \textsc{XNOR} and \textsc{Blocked-XNOR} remain the fastest, but their advantage narrows as the matrix size grows. Taken together with its near-oracle loss in the proxy study, Algorithm~\ref{alg:WGM} offers the best accuracy--efficiency trade-off and is therefore used as the default variant in subsequent experiments.

\begin{figure}
\centering
\begin{minipage}{.5\textwidth}
  \centering
  \includegraphics[width=1.0\linewidth]{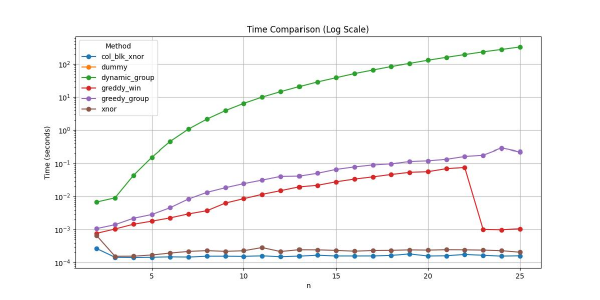}
  \captionsetup{width=0.95\linewidth}
  \caption{Algorithm comparison for small matrix, time used for quantization on CPU against matrix sized $\mathbb{R}^{n \times n}$.}
  \label{fig:speed-s}
\end{minipage}%
\begin{minipage}{.5\textwidth}
  \centering
  \includegraphics[width=1.0\linewidth]{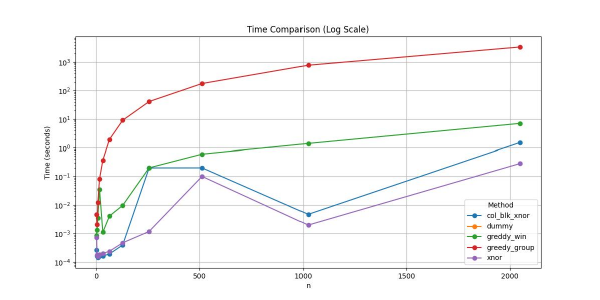}
  \captionsetup{width=0.95\linewidth}
  \caption{Algorithm comparison for large matrix, time used for quantization on CPU against matrix sized $\mathbb{R}^{n \times n}$.}
  \label{fig:speed-l}
\end{minipage}
\end{figure}

\subsection{Comparison for loss against \texorpdfstring{$\lambda$}{lambda} for greedy algorithm and windowed version}

Figure~\ref{fig:lambda} reports the MSE against $\lambda$ on a synthetic matrix in $\mathbb{R}^{512\times512}$, chosen as a compromise between computational cost and performance. We evaluate only Algorithm~\ref{alg:GG} and Algorithm~\ref{alg:WGM}, since Algorithm~\ref{alg:DG} is not computationally feasible at this scale and Algorithm~\ref{alg:LOWGM} is a modified Algorithm~\ref{alg:WGM}. Varying $\lambda$ has limited practical effect for Algorithm~\ref{alg:GG} and Algorithm~\ref{alg:WGM}, where the targeted number of groups is specified externally rather than optimized through $\lambda$. In particular, both methods attain their best MSE at $\lambda=0$, and Algorithm~\ref{alg:WGM} and~\ref{alg:LOWGM} shows negligible variation and no general trends across the entire range of $\lambda$ respectively, mirroring the weak dependence observed in downstream evaluations in the main text. These results suggest that, outside Algorithm~\ref{alg:DG}, $\lambda$ does not serve as an effective control knob for trading performance and efficiency.

\begin{figure}
\centering
  \centering
  \includegraphics[width=1.0\linewidth]{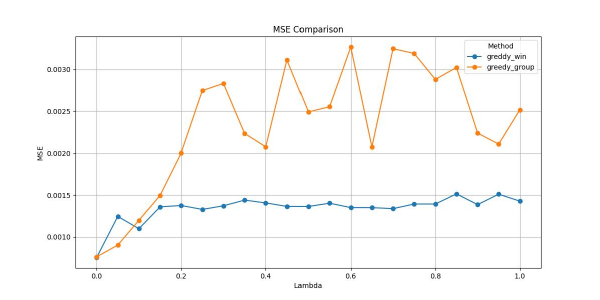}
  \captionsetup{width=0.95\linewidth}
  \caption{Loss against $\lambda$ for greedy algorithm and windowed version.}
  \label{fig:lambda}
\end{figure}

\subsection{Comparison for max groups against MSE and speed}
Since we need to choose the max group manually, we evaluate its impact on quantization error and speed using matrix $\in \mathbb{R}^{512 \times 512}$ to discover the best operating bit length in Figure \ref{fig:g-loss} and \ref{fig:g-speed}. As shown in the figures, the MSE for Algorithm \ref{alg:GG} and \ref{alg:WGM} improves and then stabilizes, plateaus around 32 groups, while quantization time decreases slowly as the max group increases.

\begin{figure}
\centering
\begin{minipage}{.5\textwidth}
  \centering
  \includegraphics[width=1.0\linewidth]{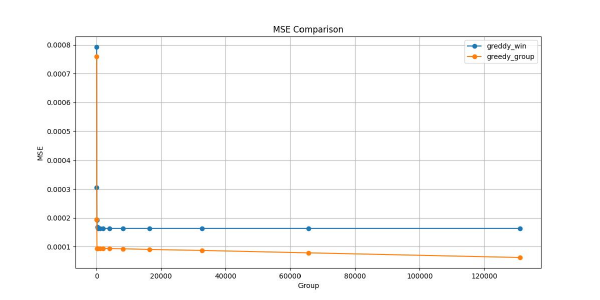}
  \captionsetup{width=0.95\linewidth}
  \caption{Max groups against MSE.}
  \label{fig:g-loss}
\end{minipage}%
\begin{minipage}{.5\textwidth}
  \centering
  \includegraphics[width=1.0\linewidth]{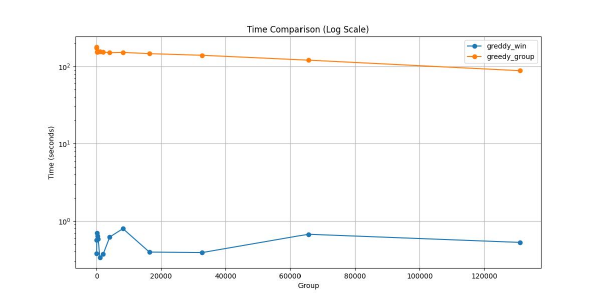}
  \captionsetup{width=0.95\linewidth}
  \caption{Max groups against quantization speed.}
    \label{fig:g-speed}
\end{minipage}
\end{figure}

\subsection{Comparison for window (initial group) size against quantization loss and speed.}
Lastly, the impact of different window sizes in Algorithm \ref{alg:WGM} on MSE and quantization speed was evaluated in the matrix simulation experiments, as shown in Figure~\ref{fig:w-loss} and \ref{fig:w-speed}. It was observed that increasing the window size led to higher MSE and reduced quantization time. When the window size was below 64, the MSE remained near its minimum, while the speed improvement flattened out between 64 and 1024. Therefore, a window size of 64 is the best balance point.

\begin{figure}
\centering
\begin{minipage}{.5\textwidth}
  \centering
  \includegraphics[width=1.0\linewidth]{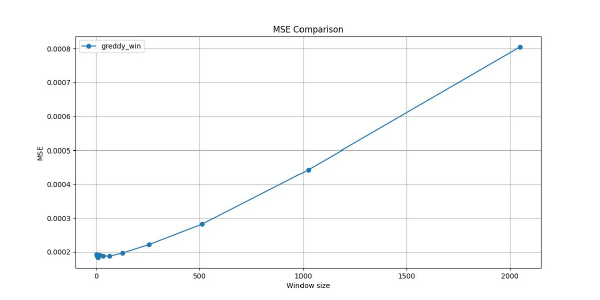}
  \captionsetup{width=0.95\linewidth}
  \caption{Window (initial group) size against loss.}
  \label{fig:w-loss}
\end{minipage}%
\begin{minipage}{.5\textwidth}
  \centering
  \includegraphics[width=1.0\linewidth]{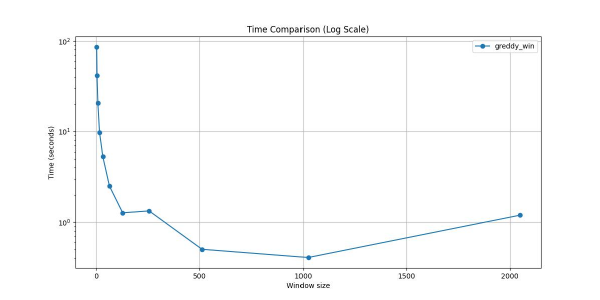}
  \captionsetup{width=0.95\linewidth}
  \caption{Window (initial group) size against quantization speed.}
  \label{fig:w-speed}
\end{minipage}
\end{figure}

\section{Expanded tables}
\label{app:ext_tab}
For completeness, we provide full versions of several condensed tables from the main text. Specifically, Table~\ref{tab:groupAndTable} is expanded into Tables~\ref{tab:max-group-size-full} and~\ref{tab:window-size-full}; Table~\ref{tab:wVSt-mse} is expanded into Tables~\ref{tab:wVSt-mse_full} and~\ref{tab:wVSt-time_full}; and Table~\ref{tab:Lambda} is expanded into Table~\ref{tab:Lambda_full}. The qualitative trends are consistent with the condensed results in the main text.

\begin{table}[h]
  \centering
  {\fontsize{9}{11}\selectfont 
  \setlength{\tabcolsep}{0.6mm} 
  \footnotesize
  \caption{
  Performance comparison for selecting “max group” $g$.  Utilizing Llama3.2 1B to analyze Perplexity across varying $g$ values with $w$=256 for efficient experimentation. 
  }
  \begin{tabular}{@{}ccc|cccc@{}} 
    \hline
    \textbf{g} & \textbf{Bit} & \textbf{Time} & \textbf{WK2} & \textbf{PTB} & \textbf{C4} & \textbf{Avg.} \\
    \hline
    8 & 4 & 429.50(177.39) & 7677 & 5165 & 4819 & 5887\\
    16 & 5 & 374.81(128.15) & 12.61 & 24.05 & 18.51 & 55.17\\
    \rowcolor{gray!20} 32 & 6 & 371.82 (127.25) & 10.94 & 20.21 & 15.92 & 15.69\\
    64 & 7 & 428.05(174.66) & 10.65 & 19.45 & 15.45 & 15.19\\
    128 & 8 & 385.05(135.06) & 10.62 & 19.37 & 15.39 & 15.13\\
    256 & 9 & 393.64(145.25) & 10.62 & 19.36 & 15.37 & 15.11\\
    512 & 10 & 383.77(145.77) & 10.62 & 19.35 & 15.36 & 15.11\\
    \hline
  \end{tabular}
  \label{tab:max-group-size-full}
   }
\end{table}

\begin{table}[h]
  \centering
  \setlength{\tabcolsep}{1mm}
  \footnotesize 
  \caption{
  Performance comparison for selecting window size $w$, using Llama3.2 1B to analyze perplexity across varying $w$ values, with $g$ = 256 for efficient experimentation. 
  }
  \begin{tabular}{@{}ccc|cccc@{}} 
    \hline
    \textbf{$w$} &\textbf{Bit} & \textbf{Time} & \textbf{WK2} & \textbf{PTB} & \textbf{C4} & \textbf{Avg.} \\
    \hline
    8 & 1.0015 & 7606.56 (139.23) & 9.90 & 17.83 & 14.22 & 13.99\\
    16 & 1.0015 & 3753.57 (147.63) & 9.90 & 17.91 & 14.24 & 14.02\\
    32 & 1.0015 & 1900.39 (134.34) & 9.90 & 17.98 & 14.24 & 14.04\\
    \rowcolor{gray!20} 64 & 1.0015 & 971.81 (133.88) & 9.92 & 17.83 & 14.24 & 14.00\\
    128 & 1.0015 & 571.83 (130.67) & 10.11 & 18.08 & 14.59 & 14.26\\
    256 & 1.0015 & 393.64 (145.25) & 10.62 & 19.36 & 15.37 & 15.11\\
    512 & 1.0015 & 296.62 (130.72) & 12.42 & 27.90 & 18.25 & 19.52\\
    \hline
  \end{tabular}
  \label{tab:window-size-full}
\end{table}

\begin{table}[h]
  \centering
  \setlength{\tabcolsep}{1mm}
  \footnotesize 
  \caption{
  Performance comparison for selecting $\lambda$. Utilizing Llama3.2 1B to analyze Perplexity across varying $\lambda$ values with $w$=256 and $g$=256 for efficient experimentation. 
  }
  \begin{tabular}{@{}cc|cccc@{}} 
    \hline
    \textbf{$\lambda$} & \textbf{Time} & \textbf{WK2} & \textbf{PTB} & \textbf{C4} & \textbf{Avg.} \\
    \hline
    0.0 & 340.10 (111.23) & 10.60 & 19.24 & 15.34 & 15.06\\
    0.1 & 391.23 (132.98) & 10.61 & 19.29 & 15.37 & 15.09\\
    0.2 & 385.21 (133.20) & 10.61 & 19.29 & 15.36 & 15.09\\
    0.3 & 394.20 (134.01) & 10.61 & 19.27 & 15.36 & 15.08\\
    0.4 & 377.17 (130.04) & 10.61 & 19.29 & 15.38 & 15.09\\
    0.5 & 386.47 (129.55) & 10.61 & 19.28 & 15.37 & 15.09\\
    0.6 & 370.80 (126.68) & 10.62 & 19.32 & 15.39 & 15.11\\
    0.7 & 378.87 (127.96) & 10.62 & 19.35 & 15.38 & 15.11\\
    0.8 & 364.53 (130.19) & 10.62 & 19.34 & 15.36 & 15.10\\
    0.9 & 374.72 (133.59) & 10.62 & 19.34 & 15.39 & 15.12\\
    1.0 & 364.32 (129.50) & 10.61 & 19.32 & 15.37 & 15.10\\
    \hline
  \end{tabular}
  \label{tab:Lambda_full}
\end{table}

\begin{table}[h]
  \centering
  \setlength{\tabcolsep}{1mm}
  \footnotesize 
  \caption{
  4-bit per tile quantization MSE of weight of first linear of Llama 3.2 1B under different block $t$ and window $w$ size
  }
  \begin{tabular}{@{}c|cccccc@{}} 
    \hline
    \backslashbox{$w$}{$t$}&\textbf{2048} &\textbf{1024} & \textbf{512} & \textbf{256} & \textbf{128} & \textbf{64}\\
    \hline
    64 & 82.43 & 107.70 & 177.52 & 394.00 & 2289.91 & /\\
    32 & 72.50 & 81.51 & 105.77 & 176.14 & 391.27 & 2272.30\\
    16 & 70.33 & 71.23 & 78.61 & 103.37 & 173.03 & 385.56\\
    8 & 69.43 & 68.41 & 67.97 & 74.85 & 98.11 & 167.19\\
    4 & 69.15 & 67.13 & 64.34 & 63.20 & 67.70 & 88.76\\
    2 & 68.98 & 66.29 & 62.64 & 57.88 & 53.88 & 53.84\\
    1 & 68.69 & 66.04 & 61.21 & 54.46 & 44.93 & 33.09\\
    \hline
  \end{tabular}
  \label{tab:wVSt-mse_full}
\end{table}

\begin{table}[h]
  \centering
  \setlength{\tabcolsep}{1mm}
  \footnotesize 
  \caption{
  4-bit per tile quantization time of weight of first linear of Llama 3.2 1B under different block $t$ and window $w$ size
  }
  \begin{tabular}{@{}c|cccccc@{}} 
    \hline
    \backslashbox{$w$}{$t$}&\textbf{2048} &\textbf{1024} & \textbf{512} & \textbf{256} & \textbf{128} & \textbf{64}\\
    \hline
    64 & 292.41 & 289.67 & 288.19 & 289.73 & 306.34 & /\\
    32 & 295.54 & 291.21 & 295.51 & 297.21 & 288.91 & 324.07\\
    16 & 296.79 & 295.96 & 297.53 & 301.66 & 308.03 & 325.27\\
    8 & 305.58 & 291.62 & 301.16 & 288.76 & 312.58 & 289.76\\
    4 & 304.28 & 320.45 & 293.23 & 317.16 & 322.78 & 290.00\\
    2 & 361.37 & 355.98 & 350.84 & 349.16 & 312.64 & 336.02\\
    1 & 394.94 & 410.03 & 385.05 & 404.13 & 406.32 & 360.47\\
    \hline
  \end{tabular}
  \label{tab:wVSt-time_full}
\end{table}

\clearpage

\section{Supplementary PPL and QA}

For completeness, we provide full versions of the condensed results in Table~\ref{tab:summary}. 
Tables~\ref{ab:full-QA_bw4} and~\ref{tab:full-PPL_bw4} report per-task QA scores and per-dataset perplexities for the block-wise 4-bit setting, respectively, whose averages correspond to the 4-bit columns in Table~\ref{tab:summary}.
Similarly, Tables~\ref{tab:full-QA_pt6} and~\ref{tab:full-PPL_pt6} provide the corresponding breakdown for the per-tensor 6-bit setting.
All qualitative trends are consistent with the condensed table in the main text.

We also provide full QA/PPL tables in the same settings for instruction-tuned models (block-wise 4-bit) and for per-tensor 5-bit/4-bit. The instruction-tuned block-wise 4-bit results follow the same trends as the base models. Per-tensor 5-bit slightly degrades \textsc{WGM}, while \textsc{HQQ} and \textsc{RTN} collapse; per-tensor 4-bit causes substantial failure across all methods in our evaluation pipeline. Collectively, these results indicate that the comparative trends and failure modes are largely stable across a broad range of settings.

\begin{table*}[h] 
  \centering
  \setlength{\tabcolsep}{1mm}
  \footnotesize 
  \caption{
  QA performance of pretrained LLMs under 4-bit block-wise quantization across seven benchmarks.
  }
  \begin{tabular}{@{}lcc| cccccccccc@{}} 
    \hline
    \textbf{Model} & \textbf{Method} & \textbf{Bits} & \textbf{ARC-C} $\uparrow$& \textbf{ARC-E} $\uparrow$& \textbf{BoolQ} $\uparrow$& \textbf{Hellaswag} $\uparrow$& \textbf{OPQA} $\uparrow$& \textbf{PIQA} $\uparrow$& \textbf{Winogrande} $\uparrow$& \textbf{Avg.} $\uparrow$\\
    \hline
    \multirow{6}{3em}{Llama 3.2 1B} & FP & 16 & 0.364 & 0.605 & 0.639 & 0.636 & 0.374 & 0.744 & 0.612 & 0.568\\
    \cline{2-11}
    & GPTQ  & 4 & 0.334 & 0.561 & 0.578 & 0.455 & \textbf{0.372} & 0.723 & 0.600 & 0.518\\
    & RTN  & 4 & 0.347 & 0.558 & 0.626 & 0.600 & 0.356 & 0.727 & 0.571 & 0.541\\
    & BnB  & 4 & \textbf{0.350} & \textbf{0.587} & \textbf{0.635} & \textbf{0.610} & 0.348 & 0.737 & 0.602 & \textbf{0.553}\\
    & HQQ  & 4 & 0.348 & 0.553 & 0.596 & 0.605 & 0.358 & \textbf{0.741} & \textbf{0.607} & 0.544\\
    \rowcolor{gray!20} 
    \global\let\CT@@do@color\relax &
    \global\let\CT@@do@color\oriCT@@do@color
    \textbf{WGM}  & 4 & 0.343 & 0.570 & 0.625 & 0.607 & 0.330 & 0.725 & 0.583 & 0.540\\
    
    \hline
    \multirow{6}{3em}{Llama 3.2 3B} & FP & 16 & 0.458 & 0.716 & 0.728 & 0.736 & 0.432 & 0.775 & 0.693 & 0.648\\
    \cline{2-11}
    & GPTQ  & 4 & 0.444 & 0.694 & 0.708 & 0.722 & 0.426 & 0.763 & 0.677 & 0.633\\
    & RTN  & 4 & 0.431 & 0.662 & 0.699 & 0.720 & 0.400 & 0.771 & 0.663 & 0.621\\
    & BnB  & 4 & 0.447 & \textbf{0.713} & 0.707 & \textbf{0.728} & \textbf{0.428} & \textbf{0.777} & 0.684 & \textbf{0.641}\\
    & HQQ  & 4 & \textbf{0.452} & 0.694 & \textbf{0.734} & 0.722 & 0.412 & 0.769 & 0.685 & 0.638\\
    \rowcolor{gray!20} 
    \global\let\CT@@do@color\relax &
    \global\let\CT@@do@color\oriCT@@do@color
    \textbf{WGM}  & 4 & 0.434 & 0.683 & 0.726 & 0.714 & 0.406 & 0.761 & \textbf{0.694} & 0.631\\

    \hline
    \multirow{6}{3em}{Falcon 3 1B} & FP & 16 & 0.421 & 0.662 & 0.719 & 0.618 & 0.406 & 0.745 & 0.617 & 0.598\\
    \cline{2-11}
    & GPTQ  & 4 & \textbf{0.422} & 0.662 & \textbf{0.720} & 0.607 & \textbf{0.408} & 0.744 & 0.605 & \textbf{0.596}\\
    & RTN  & 4 & 0.398 & 0.642 & 0.696 & 0.607 & 0.402 & 0.745 & 0.590 & 0.583\\
    & BnB  & 4 & 0.421 & \textbf{0.671} & 0.718 & 0.606 & 0.386 & 0.744 & 0.601 & 0.593\\
    & HQQ  & 4 & 0.410 & 0.645 & 0.697 & \textbf{0.612} & 0.396 & \textbf{0.747} & \textbf{0.607} & 0.588\\
    \rowcolor{gray!20} 
    \global\let\CT@@do@color\relax &
    \global\let\CT@@do@color\oriCT@@do@color
    \textbf{WGM}  & 4 & 0.415 & 0.662 & 0.700 & 0.611 & \textbf{0.408} & 0.746 & 0.598 & 0.591\\

    \hline
    \multirow{6}{3em}{Falcon 3 3B} & FP & 16 & 0.472 & 0.727 & 0.735 & 0.652 & 0.394 & 0.751 & 0.650 & 0.626\\
    \cline{2-11}
    & GPTQ  & 4 & 0.451 & 0.709 & 0.740 & \textbf{0.640} & 0.382 & \textbf{0.757} & \textbf{0.640} & \textbf{0.617}\\
    & RTN  & 4 & 0.444 & 0.703 & \textbf{0.764} & 0.631 & 0.374 & 0.736 & 0.632 & 0.612\\
    & BnB  & 4 & 0.448 & 0.709 & 0.710 & 0.635 & 0.372 & 0.748 & 0.632 & 0.608\\
    & HQQ  & 4 & 0.451 & 0.696 & 0.725 & 0.632 & \textbf{0.388} & 0.746 & 0.635 & 0.611\\
    \rowcolor{gray!20} 
    \global\let\CT@@do@color\relax &
    \global\let\CT@@do@color\oriCT@@do@color
    \textbf{WGM}  & 4 & \textbf{0.455} & \textbf{0.714} & 0.703 & 0.635 & 0.384 & 0.751 & 0.638 & 0.611\\

    \hline
    \multirow{6}{3em}{Gemma 3 1b} & FP & 16 & 0.381 & 0.722 & 0.663 & 0.622 & 0.374 & 0.749 & 0.584 & 0.585\\
    \cline{2-11}
    & GPTQ  & 4 & 0.356 & 0.703 & 0.640 & 0.610 & 0.358 & 0.743 & 0.584 & 0.570\\
    & RTN  & 4 & 0.354 & 0.700 & 0.603 & 0.597 & 0.350 & 0.730 & 0.580 & 0.559\\
    & BnB  & 4 & 0.367 & 0.694 & \textbf{0.656} & 0.605 & \textbf{0.374} & 0.742 & \textbf{0.602} & \textbf{0.577}\\
    & HQQ  & 4 & 0.369 & \textbf{0.705} & 0.646 & 0.604 & 0.364 & \textbf{0.745} & 0.590 & 0.575\\
    \rowcolor{gray!20} 
    \global\let\CT@@do@color\relax &
    \global\let\CT@@do@color\oriCT@@do@color
    \textbf{WGM}  & 4 & \textbf{0.370} & 0.703 & 0.647 & \textbf{0.617} & \textbf{0.374} & 0.742 & 0.581 & 0.576\\

    \hline
    \multirow{6}{3em}{Gemma 3 4b} & FP & 16 & 0.547 & 0.817 & 0.789 & 0.758 & 0.434 & 0.803 & 0.695 & 0.692\\
    \cline{2-11}
    & GPTQ  & 4 & 0.506 & 0.793 & 0.778 & 0.744 & 0.426 & 0.799 & 0.685 & 0.676\\
    & RTN  & 4 & 0.507 & 0.798 & 0.760 & 0.743 & 0.424 & \textbf{0.800} & 0.684 & 0.674\\
    & BnB  & 4 & 0.526 & 0.798 & 0.776 & 0.747 & \textbf{0.452} & 0.791 & 0.683 & 0.682\\
    & HQQ  & 4 & \textbf{0.529} & 0.808 & 0.770 & 0.738 & 0.444 & 0.794 & 0.682 & 0.681\\
    \rowcolor{gray!20} 
    \global\let\CT@@do@color\relax &
    \global\let\CT@@do@color\oriCT@@do@color
    \textbf{WGM}  & 4 & \textbf{0.529} & \textbf{0.817} & \textbf{0.786} & \textbf{0.754} & 0.434 & 0.799 & \textbf{0.698} & \textbf{0.688}\\

    \hline
  \end{tabular}
  
  \label{ab:full-QA_bw4}
\end{table*}

\begin{table}
  \centering
  \setlength{\tabcolsep}{0.4mm}
  \small 
  \caption{
  PPL performance of pretrained LLMs under 4-bit block-wise quantization across three datasets.
  }
  \begin{tabular}{@{}lcc| cccc@{}} 
  
    \hline
    \textbf{Model} & \textbf{Method} & \textbf{Bits} & \textbf{Wikitext 2} $\downarrow$& \textbf{PTB} $\downarrow$& \textbf{C4} $\downarrow$& \textbf{Avg.} $\downarrow$\\
    \hline
    \multirow{6}{3em}{Llama 3.2 1B} & FP & 16 & 9.75 & 17.59 & 14.01 & 13.78\\
    \cline{2-7}
    & GPTQ  & 4 & 40.87 & 64.06 & 65.72 & 56.88\\
    & RTN  & 4 & 11.58 & 20.99 & 17.10 & 16.56\\
    & BnB  & 4 & \textbf{10.78} & \textbf{19.43 }& \textbf{15.82 }& \textbf{15.34}\\
    & HQQ  & 4 & 11.07 & 19.90 & 16.29 & 15.75\\
    \rowcolor{gray!20} 
    \global\let\CT@@do@color\relax &
    \global\let\CT@@do@color\oriCT@@do@color
    \textbf{WGM}  & 4 & 10.89 & 19.47 & 15.98 & 15.45\\

    \hline
    \multirow{6}{3em}{Llama 3.2 3B} & FP & 16 & 7.81 & 13.53 & 11.33 & 10.89\\
    \cline{2-7}
    & GPTQ  & 4 & 12.23 & 23.60 & 15.43 & 17.09\\
    & RTN  & 4 & 8.57 & 14.66 & 12.68 & 11.97\\
    & BnB  & 4 & \textbf{8.28} & \textbf{14.24 }& \textbf{12.15 }& \textbf{11.56}\\
    & HQQ  & 4 & 8.32 & 14.35 & 12.32 & 11.67\\
    \rowcolor{gray!20} 
    \global\let\CT@@do@color\relax &
    \global\let\CT@@do@color\oriCT@@do@color
    \textbf{WGM}  & 4 & 8.43 & 14.54 & 12.49 & 11.81\\

    \hline
    \multirow{6}{3em}{Falcon 3 1B} & FP & 16 & 9.15 & 19.67 & 17.65 & 15.49\\
    \cline{2-7}
    & GPTQ  & 4 & \textbf{9.27} & \textbf{20.04 }& \textbf{17.89 }& \textbf{15.73}\\
    & RTN  & 4 & 9.46 & 20.46 & 18.30 & 16.07\\
    & BnB  & 4 & 9.45 & 20.36 & 18.11 & 15.97\\
    & HQQ  & 4 & 9.42 & 20.41 & 18.14 & 15.99\\
    \rowcolor{gray!20} 
    \global\let\CT@@do@color\relax &
    \global\let\CT@@do@color\oriCT@@do@color
    \textbf{WGM}  & 4 & 9.31 & 20.08 & 17.96 & 15.78\\

    \hline
    \multirow{6}{3em}{Falcon 3 3B} & FP & 16 & 8.02 & 16.49 & 15.99 & 13.50\\
    \cline{2-7}
    & GPTQ  & 4 & \textbf{8.21} & \textbf{16.96 }& \textbf{16.35 }& \textbf{13.84}\\
    & RTN  & 4 & 8.63 & 18.22 & 17.15 & 14.66\\
    & BnB  & 4 & 8.39 & 17.47 & 16.73 & 14.20\\
    & HQQ  & 4 & 8.45 & 17.59 & 16.81 & 14.28\\
    \rowcolor{gray!20} 
    \global\let\CT@@do@color\relax &
    \global\let\CT@@do@color\oriCT@@do@color
    \textbf{WGM}  & 4 & 8.26 & 17.10 & 16.47 & 13.94\\
    
    \hline
\textbf{}    \multirow{6}{3em}{Gemma 3 1b} & FP & 16 & 14.17& 104.12 & 20.00 & 46.10\\
    \cline{2-7}
    & GPTQ  & 4 & 15.46 & 140.91 & \textbf{21.52 }& 59.30\\
    & RTN  & 4 & 16.37 & \textbf{127.22 }& 23.21 & \textbf{55.60}\\
    & BnB  & 4 & 17.22 & 172.16 & 23.77 & 71.05\\
    & HQQ  & 4 & 16.85 & 137.64 & 23.20 & 59.23\\
    \rowcolor{gray!20} 
    \global\let\CT@@do@color\relax &
    \global\let\CT@@do@color\oriCT@@do@color
    \textbf{WGM}  & 4 & \textbf{15.39} & 149.58 & 21.95 & 62.31\\

    \hline
    \multirow{6}{3em}{Gemma 3 4b} & FP & 16 & 10.77 & 239.02 & 16.78 & 88.86\\
    \cline{2-7}
    & GPTQ  & 4 & 11.66 & 286.68 & 17.94 & 105.42\\
    & RTN  & 4 & 12.27 & 305.39 & 18.45 & 112.04\\
    & BnB  & 4 & 11.48 & 281.24 & \textbf{17.23 }& 103.32\\
    & HQQ  & 4 & 11.28 & \textbf{217.25 }& 17.25 & \textbf{81.93}\\
    \rowcolor{gray!20} 
    \global\let\CT@@do@color\relax &
    \global\let\CT@@do@color\oriCT@@do@color
    \textbf{WGM}  & 4 & \textbf{11.19} & 249.96 & 17.55 & 92.90\\
    \hline
  \end{tabular}
  \label{tab:full-PPL_bw4}
\end{table}

\begin{table*}[t] 
  \centering
  \setlength{\tabcolsep}{1mm}
  \footnotesize 
  \caption{
  QA performance of instruction finetuned LLMs under 4-bit block-wise quantization across seven benchmarks.
  }
  \begin{tabular}{@{}lcc| cccccccccc@{}} 
    \hline
    \textbf{Model} & \textbf{Method} & \textbf{Bits} & \textbf{ARC-C} $\uparrow$& \textbf{ARC-E} $\uparrow$& \textbf{BoolQ} $\uparrow$& \textbf{Hellaswag} $\uparrow$& \textbf{OPQA} $\uparrow$& \textbf{PIQA} $\uparrow$& \textbf{Winogrande} $\uparrow$& \textbf{Avg.} $\uparrow$\\
    \hline
    \multirow{6}{3em}{Llama 3.2 1B} & FP & 16 & 0.379 & 0.634 & 0.692 & 0.608 & 0.344 & 0.745 & 0.603 & 0.572\\
    \cline{2-11}
    & GPTQ  & 4 & \textbf{0.363} & 0.573 & 0.674 & 0.577 & 0.344 & 0.706 & 0.576 & 0.545\\
    & RTN  & 4 & 0.343 & 0.609 & 0.673 & 0.565 & 0.332 & 0.718 & 0.581 & 0.546\\
    & BnB  & 4 & \textbf{0.363} & \textbf{0.616} & \textbf{0.684} & 0.585 & 0.340 & \textbf{0.730} & 0.577 & \textbf{0.556}\\
    & HQQ  & 4 & 0.357 & 0.609 & 0.664 & 0.585 & \textbf{0.346} & 0.728 & \textbf{0.591} & 0.554\\
    \rowcolor{gray!20} 
    \global\let\CT@@do@color\relax &
    \global\let\CT@@do@color\oriCT@@do@color
    \textbf{WGM}  & 4 & \textbf{0.363} & 0.607 & 0.668 & \textbf{0.587} & 0.336 & 0.720 & 0.577 & 0.551\\

    \hline
    \multirow{6}{3em}{Llama 3.2 3B} & FP & 16 & 0.460 & 0.678 & 0.784 & 0.704 & 0.360 & 0.756 & 0.675 & 0.631\\
    \cline{2-11}
    & GPTQ  & 4 & 0.433 & 0.658 & 0.753 & 0.694 & 0.360 & 0.745 & 0.657 & 0.614\\
    & RTN  & 4 & 0.431 & 0.639 & \textbf{0.779} & 0.696 & \textbf{0.372} & \textbf{0.753} & \textbf{0.680} & \textbf{0.621}\\
    & BnB  & 4 & \textbf{0.447} & 0.641 & 0.770 & 0.699 & 0.368 & 0.746 & 0.668 & 0.620\\
    & HQQ  & 4 & 0.433 & 0.657 & 0.762 & \textbf{0.701} & 0.370 & 0.749 & 0.662 & 0.619\\
    \rowcolor{gray!20} 
    \global\let\CT@@do@color\relax &
    \global\let\CT@@do@color\oriCT@@do@color
    \textbf{WGM}  & 4 & 0.437 & \textbf{0.681} & 0.734 & 0.686 & \textbf{0.372} & 0.732 & 0.655 & 0.614\\

    \hline
    \multirow{6}{3em}{Falcon3 1B} & FP & 16 & 0.453 & 0.680 & 0.732 & 0.631 & 0.404 & 0.750 & 0.601 & 0.607\\
    \cline{2-11}
    & GPTQ  & 4 & 0.442 & 0.678 & \textbf{0.731} & 0.626 & 0.396 & 0.747 & 0.594 & 0.602\\
    & RTN  & 4 & 0.436 & 0.675 & 0.712 & 0.626 & 0.404 & 0.737 & 0.599 & 0.598\\
    & BnB  & 4 & 0.455 & 0.670 & 0.728 & 0.625 & 0.408 & 0.742 & \textbf{0.609} & 0.605\\
    & HQQ  & 4 & 0.452 & 0.643 & 0.710 & 0.624 & \textbf{0.410} & 0.740 & 0.586 & 0.595\\
    \rowcolor{gray!20} 
    \global\let\CT@@do@color\relax &
    \global\let\CT@@do@color\oriCT@@do@color
    \textbf{WGM}  & 4 & \textbf{0.459} & \textbf{0.683} & 0.728 & \textbf{0.627} & 0.402 & \textbf{0.751} & 0.606 & \textbf{0.608}\\

    \hline
    \multirow{6}{3em}{Falcon3 3B} & FP & 16 & 0.544 & 0.749 & 0.788 & 0.701 & 0.422 & 0.759 & 0.649 & 0.659\\
    \cline{2-11}
    & GPTQ  & 4 & 0.506 & 0.716 & 0.775 & \textbf{0.680} & 0.406 & \textbf{0.754} & \textbf{0.648} & 0.641\\
    & RTN  & 4 & 0.495 & 0.734 & 0.783 & 0.671 & 0.410 & 0.738 & 0.625 & 0.637\\
    & BnB  & 4 & 0.506 & 0.732 & \textbf{0.789} & 0.679 & 0.410 & 0.752 & 0.626 & 0.642\\
    & HQQ  & 4 & \textbf{0.521} & \textbf{0.751} & 0.787 & 0.676 & 0.414 & 0.748 & 0.637 & \textbf{0.648}\\
    \rowcolor{gray!20} 
    \global\let\CT@@do@color\relax &
    \global\let\CT@@do@color\oriCT@@do@color
    \textbf{WGM}  & 4 & 0.480 & 0.735 & 0.720 & 0.638 & \textbf{0.416} & 0.749 & 0.595 & 0.619\\

    \hline
    \multirow{6}{3em}{Gemma3 1b} & FP & 16 & 0.385 & 0.637 & 0.759 & 0.578 & 0.384 & 0.724 & 0.589 & 0.579\\
    \cline{2-11}
    & GPTQ  & 4 & 0.363 & 0.606 & \textbf{0.738} & 0.536 & \textbf{0.374} & 0.709 & \textbf{0.599} & 0.561\\
    & RTN  & 4 & 0.338 & 0.564 & 0.712 & 0.536 & 0.360 & 0.714 & 0.575 & 0.543\\
    & BnB  & 4 & 0.356 & 0.609 & 0.703 & \textbf{0.562} & 0.366 & \textbf{0.725} & 0.579 & 0.557\\
    & HQQ  & 4 & 0.353 & 0.583 & 0.717 & 0.551 & 0.370 & 0.720 & 0.587 & 0.555\\
    \rowcolor{gray!20} 
    \global\let\CT@@do@color\relax &
    \global\let\CT@@do@color\oriCT@@do@color
    \textbf{WGM}  & 4 & \textbf{0.394} & \textbf{0.611} & 0.735 & \textbf{0.562} & 0.368 & 0.724 & 0.584 & \textbf{0.568}\\

    \hline
    \multirow{6}{3em}{Gemma3 4b} & FP & 16 & 0.571 & 0.779 & 0.839 & 0.744 & 0.468 & 0.774 & 0.695 & 0.696\\
    \cline{2-11}
    & GPTQ  & 4 & 0.539 & 0.765 & 0.827 & 0.727 & 0.450 & 0.771 & 0.666 & 0.678\\
    & RTN  & 4 & 0.532 & \textbf{0.779} & 0.826 & 0.733 & 0.448 & 0.773 & \textbf{0.680} & 0.681\\
    & BnB  & 4 & 0.529 & 0.772 & \textbf{0.840} & 0.731 & 0.464 & \textbf{0.779} & \textbf{0.680} & 0.685\\
    & HQQ  & 4 & 0.542 & 0.764 & 0.829 & 0.732 & 0.460 & 0.768 & 0.669 & 0.681\\
    \rowcolor{gray!20} 
    \global\let\CT@@do@color\relax &
    \global\let\CT@@do@color\oriCT@@do@color
    \textbf{WGM}  & 4 & \textbf{0.561} & 0.754 & 0.831 & \textbf{0.738} & \textbf{0.468} & 0.772 & 0.679 & \textbf{0.686}\\

    \hline
  \end{tabular}
  
  \label{tab:result-QA-task}
\end{table*}

\makeatletter
\global\let\oriCT@@do@color\CT@@do@color

\begin{table}
  \centering
  \setlength{\tabcolsep}{0.4mm}
  \small 
  \caption{
  PPL performance of instruction finetuned LLMs under 4-bit block-wise quantization across three datasets.
  }
  \begin{tabular}{@{}lcc| cccc@{}} 
  
    \hline
    \textbf{Model} & \textbf{Method} & \textbf{Bits} & \textbf{Wikitext 2} $\downarrow$& \textbf{PTB} $\downarrow$& \textbf{C4} $\downarrow$& \textbf{Avg.} $\downarrow$\\
    \hline
    \multirow{6}{3em}{Llama 3.2 1B} & FP & 16 & 13.16 & 25.69 & 21.31 & 20.05\\
    \cline{2-7}
    & GPTQ  & 4 & 16.86 & 30.03 & 26.35 & 24.41\\
    & RTN  & 4 & 16.22 & 31.91 & 26.59 & 24.91\\
    & BnB  & 4 & \textbf{14.67} & \textbf{28.80 }& \textbf{23.37 }& \textbf{22.28}\\
    & HQQ  & 4 & 15.05 & 29.19 & 24.18 & 22.81\\
    \rowcolor{gray!20} 
    \global\let\CT@@do@color\relax &
    \global\let\CT@@do@color\oriCT@@do@color
    \textbf{WGM}  & 4 & 15.25 & 29.39 & 23.90 & 22.85\\

    \hline
    \multirow{6}{3em}{Llama 3.2 3B} & FP & 16 & 11.04 & 20.42 & 16.49 & 15.98\\
    \cline{2-7}
    & GPTQ  & 4 & 13.07 & 25.48 & 17.93 & 18.83\\
    & RTN  & 4 & 11.84 & 21.50 & 17.70 & 17.01\\
     & BnB  & 4 & \textbf{11.67} & \textbf{20.55 }& \textbf{17.16 }& \textbf{16.46}\\
    & HQQ  & 4 & 11.70 & 21.85 & 17.42 & 16.99\\
    \rowcolor{gray!20} 
    \global\let\CT@@do@color\relax &
    \global\let\CT@@do@color\oriCT@@do@color
    \textbf{WGM}  & 4 & 12.02 & 24.55 & 17.47 & 18.01\\

    \hline
    \multirow{6}{3em}{Falcon 3 1B} & FP & 16 & 10.51 & 22.29 & 19.47 & 17.42\\
    \cline{2-7}
    & GPTQ  & 4 & 27.11 & 63.73 & 62.39 & 51.08\\
    & RTN  & 4 & 10.72 & 23.07 & 20.08 & 17.96\\
    & BnB  & 4 & 10.73 & 22.67 & 19.86 & 17.75\\
    & HQQ  & 4 & 10.75 & 22.74 & 19.87 & 17.79\\
    \rowcolor{gray!20} 
    \global\let\CT@@do@color\relax &
    \global\let\CT@@do@color\oriCT@@do@color
    \textbf{WGM}  & 4 & \textbf{10.42} & \textbf{22.51 }& \textbf{19.62 }& \textbf{17.51}\\

    \hline
    \multirow{6}{3em}{Falcon 3 3B} & FP & 16 & 10.11 & 20.19 & 18.27 & 16.19\\
    \cline{2-7}
    & GPTQ  & 4 & 15.57 & 42.65 & 40.17 & 32.80\\
    & RTN  & 4 & 10.51 & 21.21 & 19.18 & 16.97\\
    & BnB  & 4 & 10.57 & 21.20 & 19.16 & 16.98\\
    & HQQ  & 4 & \textbf{9.96} & \textbf{20.90 }& \textbf{18.86 }& \textbf{16.57}\\
    \rowcolor{gray!20} 
    \global\let\CT@@do@color\relax &
    \global\let\CT@@do@color\oriCT@@do@color
    \textbf{WGM}  & 4 & 11.01 & 21.66 & 20.31 & 17.66\\

    \hline
    \multirow{6}{3em}{Gemma 3 1b} & FP & 16 & 27.67 & 212.14 & 33.21 & 91.01\\
    \cline{2-7}
    & GPTQ  & 4 & \textbf{31.00} & \textbf{251.43 }& 38.00 & \textbf{106.81}\\
    & RTN  & 4 & 37.07 & 362.66 & 43.29 & 147.67\\
    & BnB  & 4 & 33.90 & 311.52 & 39.89 & 128.44\\
    & HQQ  & 4 & 34.77 & 310.36 & 40.58 & 128.57\\
    \rowcolor{gray!20} 
    \global\let\CT@@do@color\relax &
    \global\let\CT@@do@color\oriCT@@do@color
    \textbf{WGM}  & 4 & 31.71 & 266.08 & \textbf{37.33 }& 111.71\\

    \hline
    \multirow{6}{3em}{Gemma 3 4b} & FP & 16 & 17.36 & 644.68 & 23.38 & 228.47\\
    \cline{2-7}
    & GPTQ  & 4 & 19.03 & 818.43 & 25.48 & 287.65\\
    & RTN  & 4 & 18.48 & \textbf{558.52 }& 25.64 & \textbf{200.88}\\
    & BnB  & 4 & 19.82 & 811.48 & 25.92 & 285.74\\
    & HQQ  & 4 & \textbf{18.25} & 642.86 & \textbf{24.53 }& 228.55\\
    \rowcolor{gray!20} 
    \global\let\CT@@do@color\relax &
    \global\let\CT@@do@color\oriCT@@do@color
    \textbf{WGM}  & 4 & 18.34 & 736.24 & \textbf{24.53} & 259.71\\
    \hline
  \end{tabular}
  \label{tab:PPL-comparison}
\end{table}

\makeatletter
\global\let\oriCT@@do@color\CT@@do@color 

\begin{table*}[t] 
  \centering
  \setlength{\tabcolsep}{1mm}
  \footnotesize 
  \caption{
  QA performance of pretrained LLMs under 6-bit per-tensor quantization across seven benchmarks.
  }
  \begin{tabular}{@{}lcc| cccccccccc@{}} 
    \hline
    \textbf{Model} & \textbf{Method} & \textbf{Bits} & \textbf{ARC-C} $\uparrow$& \textbf{ARC-E} $\uparrow$& \textbf{BoolQ} $\uparrow$& \textbf{Hellaswag} $\uparrow$& \textbf{OPQA} $\uparrow$& \textbf{PIQA} $\uparrow$& \textbf{Winogrande} $\uparrow$& \textbf{Avg.} $\uparrow$\\
    \hline
    \multirow{7}{3em}{Llama 3.2 1B} & FP & 16 & 0.364 & 0.605 & 0.639 & 0.636 & 0.374 & 0.744 & 0.612 & 0.568\\
    \cline{2-11}
    & RTN  & 6 & 0.228 & 0.331 & 0.551 & 0.327 & 0.224 & 0.563 & 0.527 & 0.393\\
    & HQQ  & 6 & 0.240 & 0.379 & 0.439 & 0.398 & 0.258 & 0.591 & 0.520 & 0.404\\
    \rowcolor{gray!20} 
    \global\let\CT@@do@color\relax &
    \global\let\CT@@do@color\oriCT@@do@color
    \textbf{WGM}  & 6 & \textbf{0.369} & \textbf{0.596} & 0.636 & \textbf{0.636} & \textbf{0.366} & 0.732 & 0.590 & 0.561\\
    \rowcolor{gray!20} 
    \global\let\CT@@do@color\relax &
    \global\let\CT@@do@color\oriCT@@do@color
    \textbf{WGM-HI} & 6 & 0.359 & 0.592 & \textbf{0.640} & 0.632 & 0.362 & \textbf{0.747} & \textbf{0.600} & \textbf{0.562}\\
    
    \hline
    \multirow{7}{3em}{Llama 3.2 3B} & FP & 16 & 0.458 & 0.716 & 0.728 & 0.736 & 0.432 & 0.775 & 0.693 & 0.648\\
    \cline{2-11}
    & RTN  & 6 & 0.298 & 0.445 & 0.585 & 0.544 & 0.312 & 0.633 & 0.571 & 0.484\\
    & HQQ  & 6 & 0.224 & 0.282 & 0.545 & 0.279 & 0.248 & 0.524 & 0.510 & 0.373\\
    \rowcolor{gray!20} 
    \global\let\CT@@do@color\relax &
    \global\let\CT@@do@color\oriCT@@do@color
    \textbf{WGM}  & 6 & 0.444 & 0.695 & 0.667 & 0.720 & 0.408 & 0.770 & 0.685 & 0.627\\
    \rowcolor{gray!20} 
    \global\let\CT@@do@color\relax &
    \global\let\CT@@do@color\oriCT@@do@color
    \textbf{WGM-HI} & 6 & \textbf{0.463} & \textbf{0.701} & \textbf{0.698} & \textbf{0.733} & \textbf{0.434} & \textbf{0.779} & \textbf{0.699} & \textbf{0.644}\\

    \hline
    \multirow{7}{3em}{Falcon 3 1B} & FP & 16 & 0.421 & 0.662 & 0.719 & 0.618 & 0.406 & 0.745 & 0.617 & 0.598\\
    \cline{2-11}
    & RTN  & 6 & 0.363 & 0.593 & 0.570 & 0.558 & 0.348 & 0.708 & 0.585 & 0.532\\
    & HQQ  & 6 & 0.375 & 0.629 & 0.582 & 0.583 & 0.384 & 0.736 & 0.591 & 0.554\\
    \rowcolor{gray!20} 
    \global\let\CT@@do@color\relax &
    \global\let\CT@@do@color\oriCT@@do@color
    \textbf{WGM}  & 6 & 0.421 & 0.663 & \textbf{0.700} & \textbf{0.614} & 0.402 & 0.745 & 0.596 & 0.592\\
    \rowcolor{gray!20} 
    \global\let\CT@@do@color\relax &
    \global\let\CT@@do@color\oriCT@@do@color
    \textbf{WGM-HI} & 6 & \textbf{0.426} & \textbf{0.670} & 0.697 & 0.613 & \textbf{0.406} & \textbf{0.748} & \textbf{0.604} & \textbf{0.595}\\

    \hline
    \multirow{7}{3em}{Falcon 3 3B} & FP & 16 & 0.472 & 0.727 & 0.735 & 0.652 & 0.394 & 0.751 & 0.650 & 0.626\\
    \cline{2-11}
    & RTN  & 6 & 0.401 & 0.649 & 0.713 & 0.583 & 0.374 & 0.727 & 0.595 & 0.577\\
    & HQQ  & 6 & 0.393 & 0.633 & 0.668 & 0.540 & 0.342 & 0.703 & 0.576 & 0.551\\
    \rowcolor{gray!20} 
    \global\let\CT@@do@color\relax &
    \global\let\CT@@do@color\oriCT@@do@color
    \textbf{WGM}  & 6 & 0.469 & \textbf{0.715} & \textbf{0.761} & \textbf{0.654} & \textbf{0.398} & \textbf{0.755} & 0.637 & \textbf{0.627}\\
    \rowcolor{gray!20} 
    \global\let\CT@@do@color\relax &
    \global\let\CT@@do@color\oriCT@@do@color
    \textbf{WGM-HI} & 6 & \textbf{0.473} & 0.714 & 0.735 & 0.653 & 0.382 & 0.752 & \textbf{0.648} & 0.622\\

    \hline
    \multirow{7}{3em}{Gemma 3 1b} & FP & 16 & 0.381 & 0.722 & 0.663 & 0.622 & 0.374 & 0.749 & 0.584 & 0.585\\
    \cline{2-11}
    & RTN  & 6 & 0.333 & 0.629 & 0.565 & 0.536 & 0.342 & 0.713 & 0.575 & 0.528\\
    & HQQ  & 6 & 0.321 & 0.623 & 0.448 & 0.549 & 0.338 & 0.713 & 0.551 & 0.506\\
    \rowcolor{gray!20} 
    \global\let\CT@@do@color\relax &
    \global\let\CT@@do@color\oriCT@@do@color
    \textbf{WGM}  & 6 & \textbf{0.386} & \textbf{0.722} & \textbf{0.664} & \textbf{0.623} & 0.358 & 0.747 & \textbf{0.583} & \textbf{0.583}\\
    \rowcolor{gray!20} 
    \global\let\CT@@do@color\relax &
    \global\let\CT@@do@color\oriCT@@do@color
    \textbf{WGM-HI} & 6 & 0.373 & 0.714 & 0.663 & 0.621 & \textbf{0.368} & \textbf{0.748} & 0.568 & 0.579\\

    \hline
    \multirow{7}{3em}{Gemma 3 4b} & FP & 16 & 0.547 & 0.817 & 0.789 & 0.758 & 0.434 & 0.803 & 0.695 & 0.692\\
    \cline{2-11}
    & RTN  & 6 & 0.375 & 0.641 & 0.562 & 0.626 & 0.382 & 0.746 & 0.639 & 0.567\\
    & HQQ  & 6 & 0.403 & 0.693 & 0.536 & 0.647 & 0.382 & 0.754 & 0.635 & 0.579\\
    \rowcolor{gray!20} 
    \global\let\CT@@do@color\relax &
    \global\let\CT@@do@color\oriCT@@do@color
    \textbf{WGM}  & 6 & 0.545 & 0.817 & \textbf{0.790} & \textbf{0.764} & 0.436 & \textbf{0.808} & \textbf{0.696} & \textbf{0.694}\\
    \rowcolor{gray!20} 
    \global\let\CT@@do@color\relax &
    \global\let\CT@@do@color\oriCT@@do@color
    \textbf{WGM-HI} & 6 & \textbf{0.554} & \textbf{0.821} & 0.776 & 0.760 & \textbf{0.444} & 0.804 & 0.687 & 0.692\\

    \hline
  \end{tabular}
  
  \label{tab:full-QA_pt6}
\end{table*}

\begin{table}
  \centering
  \setlength{\tabcolsep}{0.4mm}
  \small 
  \caption{
  PPL performance of pretrained LLMs under 6-bit per-tensor quantization across three datasets.
  }
  \begin{tabular}{@{}lcc| cccc@{}} 
  
    \hline
    \textbf{Model} & \textbf{Method} & \textbf{Bits} & \textbf{Wikitext 2} $\downarrow$& \textbf{PTB} $\downarrow$& \textbf{C4} $\downarrow$& \textbf{Avg.} $\downarrow$\\
    \hline
    \multirow{7}{3em}{Llama 3.2 1B} & FP & 16 & 9.75 & 17.59 & 14.01 & 13.78\\
    \cline{2-7}
    & RTN  & 6 & 132.89 & 190.70 & 184.83 & 169.47\\
    & HQQ  & 6 & 82.94 & 132.95 & 102.17 & 106.02\\
    \rowcolor{gray!20} 
    \global\let\CT@@do@color\relax &
    \global\let\CT@@do@color\oriCT@@do@color
    \textbf{WGM}  & 6 & \textbf{10.04} & \textbf{18.04} & \textbf{14.45} & \textbf{14.18}\\
    \rowcolor{gray!20} 
    \global\let\CT@@do@color\relax &
    \global\let\CT@@do@color\oriCT@@do@color
    \textbf{WGM-HI}  & 6 & 10.09 & 18.16 & 14.57 & 14.27\\

    \hline
    \multirow{7}{3em}{Llama 3.2 3B} & FP & 16 & 7.81 & 13.53 & 11.33 & 10.89\\
    \cline{2-7}
    & RTN  & 6 & 22.26 & 48.53 & 32.01 & 34.26\\
    & HQQ  & 6 & 148.64 & 261.89 & 341.63 & 250.72\\
    \rowcolor{gray!20} 
    \global\let\CT@@do@color\relax &
    \global\let\CT@@do@color\oriCT@@do@color
    \textbf{WGM}  & 6 & 8.53 & 15.24 & 12.97 & 12.25\\
    \rowcolor{gray!20} 
    \global\let\CT@@do@color\relax &
    \global\let\CT@@do@color\oriCT@@do@color
    \textbf{WGM-HI}  & 6 & \textbf{7.99} & \textbf{13.82} & \textbf{11.64} & \textbf{11.15}\\

    \hline
    \multirow{7}{3em}{Falcon 3 1B} & FP & 16 & 9.15 & 19.67 & 17.65 & 15.49\\
    \cline{2-7}
    & RTN  & 6 & 12.28 & 26.58 & 23.37 & 20.74\\
    & HQQ  & 6 & 10.92 & 24.13 & 21.12 & 18.72\\
    \rowcolor{gray!20} 
    \global\let\CT@@do@color\relax &
    \global\let\CT@@do@color\oriCT@@do@color
    \textbf{WGM}  & 6 & \textbf{9.55} & \textbf{20.87} & \textbf{18.35} & \textbf{16.26}\\
    \rowcolor{gray!20} 
    \global\let\CT@@do@color\relax &
    \global\let\CT@@do@color\oriCT@@do@color
    \textbf{WGM-HI}  & 6 & 9.82 & 21.93 & 18.97 & 16.91\\

    \hline
    \multirow{7}{3em}{Falcon 3 3B} & FP & 16 & 8.02 & 16.49 & 15.99 & 13.50\\
    \cline{2-7}
    & RTN  & 6 & 11.38 & 24.64 & 22.24 & 19.42\\
    & HQQ  & 6 & 11.83 & 24.98 & 23.20 & 20.00\\
    \rowcolor{gray!20} 
    \global\let\CT@@do@color\relax &
    \global\let\CT@@do@color\oriCT@@do@color
    \textbf{WGM}  & 6 & 8.27 & 17.25 & 16.44 & 13.98\\
    \rowcolor{gray!20} 
    \global\let\CT@@do@color\relax &
    \global\let\CT@@do@color\oriCT@@do@color
    \textbf{WGM-HI}  & 6 & \textbf{8.13} & \textbf{16.84} & \textbf{16.18} & \textbf{13.72}\\

    \hline
    \multirow{7}{3em}{Gemma 3 1b} & FP & 16 & 14.17& 104.12 & 20.00 & 46.10\\
    \cline{2-7}
    & RTN  & 6 & 29.04 & 345.07 & 39.10 & 137.74\\
    & HQQ  & 6 & 31.11 & 414.47 & 39.70 & 161.76\\
    \rowcolor{gray!20} 
    \global\let\CT@@do@color\relax &
    \global\let\CT@@do@color\oriCT@@do@color
    \textbf{WGM}  & 6 & \textbf{14.37} & 124.18 & \textbf{20.21} & 52.92\\
    \rowcolor{gray!20} 
    \global\let\CT@@do@color\relax &
    \global\let\CT@@do@color\oriCT@@do@color
    \textbf{WGM-HI}  & 6 & 14.49 & \textbf{111.63 }& 20.69 & \textbf{48.94}\\

    \hline
    \multirow{7}{3em}{Gemma 3 4b} & FP & 16 & 10.77 & 239.02 & 16.78 & 88.86\\
    \cline{2-7}
    & RTN  & 6 & 24.46 & 703.53 & 32.97 & 253.65\\
    & HQQ  & 6 & 22.70 & 947.37 & 33.72 & 334.60\\
    \rowcolor{gray!20} 
    \global\let\CT@@do@color\relax &
    \global\let\CT@@do@color\oriCT@@do@color
    \textbf{WGM}  & 6 & \textbf{10.77} & \textbf{256.62 }& \textbf{16.90} & \textbf{94.76}\\
    \rowcolor{gray!20} 
    \global\let\CT@@do@color\relax &
    \global\let\CT@@do@color\oriCT@@do@color
    \textbf{WGM-HI}  & 6 & 10.96 & 275.90 & 17.03 & 101.29\\
    \hline
  \end{tabular}
  \label{tab:full-PPL_pt6}
\end{table}

\makeatletter
\global\let\oriCT@@do@color\CT@@do@color 

\begin{table*}[t] 
  \centering
  \setlength{\tabcolsep}{1mm}
  \footnotesize 
  \caption{
  QA performance of pretrained LLMs under 5-bit per-tensor quantization across seven benchmarks.
  }
  \begin{tabular}{@{}lcc| cccccccccc@{}} 
    \hline
    \textbf{Model} & \textbf{Method} & \textbf{Bits} & \textbf{ARC-C} $\uparrow$& \textbf{ARC-E} $\uparrow$& \textbf{BoolQ} $\uparrow$& \textbf{Hellaswag} $\uparrow$& \textbf{OPQA} $\uparrow$& \textbf{PIQA} $\uparrow$& \textbf{Winogrande} $\uparrow$& \textbf{Avg.} $\uparrow$\\
    \hline
    \multirow{7}{3em}{Llama 3.2 1B} & FP & 16 & 0.364 & 0.605 & 0.639 & 0.636 & 0.374 & 0.744 & 0.612 & 0.568\\
    \cline{2-11}
    & RTN  & 5 & 0.258 & 0.256 & 0.385 & 0.260 & 0.272 & 0.516 & 0.498 & 0.349\\
    & HQQ  & 5 & 0.272 & 0.264 & 0.381 & 0.257 & 0.284 & 0.493 & 0.511 & 0.352\\
    \rowcolor{gray!20} 
    \global\let\CT@@do@color\relax &
    \global\let\CT@@do@color\oriCT@@do@color
    \textbf{WGM}  & 5 & \textbf{0.353} & \textbf{0.587} & 0.569 & \textbf{0.617} & \textbf{0.350} & \textbf{0.724} & 0.578 & \textbf{0.540}\\
    \rowcolor{gray!20} 
    \global\let\CT@@do@color\relax &
    \global\let\CT@@do@color\oriCT@@do@color
    \textbf{WGM-LO}  & 5 & 0.346 & 0.524 & \textbf{0.606} & 0.596 & 0.346 & 0.715 & \textbf{0.579} & 0.530\\
    
    \hline
    \multirow{7}{3em}{Llama 3.2 3B} & FP & 16 & 0.458 & 0.716 & 0.728 & 0.736 & 0.432 & 0.775 & 0.693 & 0.648\\
    \cline{2-11}
    & RTN  & 5 & \textbf{0.265} & 0.267 & 0.378 & 0.264 & \textbf{0.282} & 0.481 & 0.508 & 0.349\\
    & HQQ  & 5 & 0.264 & 0.262 & 0.423 & 0.263 & \textbf{0.282} & 0.508 & 0.492 & 0.356\\
    \rowcolor{gray!20} 
    \global\let\CT@@do@color\relax &
    \global\let\CT@@do@color\oriCT@@do@color
    \textbf{WGM}  & 5 & 0.212 & 0.317 & 0.394 & 0.289 & 0.244 & 0.551 & 0.511 & 0.360\\
    \rowcolor{gray!20} 
    \global\let\CT@@do@color\relax &
    \global\let\CT@@do@color\oriCT@@do@color
    \textbf{WGM-LO}  & 5 & 0.219 & \textbf{0.389} & \textbf{0.466} & \textbf{0.310} & 0.256 & \textbf{0.610} & \textbf{0.525} & \textbf{0.396}\\

    \hline
    \multirow{7}{3em}{Falcon 3 1B} & FP & 16 & 0.421 & 0.662 & 0.719 & 0.618 & 0.406 & 0.745 & 0.617 & 0.598\\
    \cline{2-11}
    & RTN  & 5 & 0.293 & 0.465 & 0.480 & 0.415 & 0.296 & 0.636 & 0.534 & 0.446\\
    & HQQ  & 5 & 0.304 & 0.557 & 0.635 & 0.462 & 0.290 & 0.677 & 0.541 & 0.495\\
    \rowcolor{gray!20} 
    \global\let\CT@@do@color\relax &
    \global\let\CT@@do@color\oriCT@@do@color
    \textbf{WGM}  & 5 & \textbf{0.399} & \textbf{0.646} & \textbf{0.699} & \textbf{0.600} & \textbf{0.400} & \textbf{0.742} & \textbf{0.594} & \textbf{0.583}\\
    \rowcolor{gray!20} 
    \global\let\CT@@do@color\relax &
    \global\let\CT@@do@color\oriCT@@do@color
    \textbf{WGM-LO}  & 5 & 0.397 & 0.643 & 0.683 & 0.562 & 0.376 & 0.726 & 0.579 & 0.567\\

    \hline
    \multirow{7}{3em}{Falcon 3 3B} & FP & 16 & 0.472 & 0.727 & 0.735 & 0.652 & 0.394 & 0.751 & 0.650 & 0.626\\
    \cline{2-11}
    & RTN  & 5 & 0.272 & 0.402 & 0.469 & 0.381 & 0.284 & 0.588 & 0.518 & 0.416\\
    & HQQ  & 5 & 0.246 & 0.317 & 0.541 & 0.292 & 0.272 & 0.547 & 0.491 & 0.386\\
    \rowcolor{gray!20} 
    \global\let\CT@@do@color\relax &
    \global\let\CT@@do@color\oriCT@@do@color
    \textbf{WGM}  & 5 & \textbf{0.442} & 0.683 & \textbf{0.747} & 0.618 & 0.382 & 0.741 & 0.623 & 0.605\\
    \rowcolor{gray!20} 
    \global\let\CT@@do@color\relax &
    \global\let\CT@@do@color\oriCT@@do@color
    \textbf{WGM-LO}  & 5 & 0.437 & \textbf{0.689} & 0.737 & \textbf{0.619} & \textbf{0.396} & \textbf{0.748} & \textbf{0.634} & \textbf{0.608}\\

    \hline
    \multirow{7}{3em}{Gemma 3 1b} & FP & 16 & 0.381 & 0.722 & 0.663 & 0.622 & 0.374 & 0.749 & 0.584 & 0.585\\
    \cline{2-11}
    & RTN  & 5 & 0.222 & 0.292 & 0.389 & 0.271 & 0.270 & 0.533 & 0.482 & 0.351\\
    & HQQ  & 5 & 0.231 & 0.330 & 0.490 & 0.294 & 0.270 & 0.545 & 0.500 & 0.380\\
    \rowcolor{gray!20} 
    \global\let\CT@@do@color\relax &
    \global\let\CT@@do@color\oriCT@@do@color
    \textbf{WGM}  & 5 & 0.382 & 0.710 & 0.652 & 0.628 & \textbf{0.364} & 0.744 & \textbf{0.581} & 0.580\\
    \rowcolor{gray!20} 
    \global\let\CT@@do@color\relax &
    \global\let\CT@@do@color\oriCT@@do@color
    \textbf{WGM-LO}  & 5 & \textbf{0.383} & \textbf{0.711} & \textbf{0.661} & \textbf{0.632} & 0.360 & \textbf{0.748} & 0.579 & \textbf{0.582}\\

    \hline
    \multirow{7}{3em}{Gemma 3 4b} & FP & 16 & 0.547 & 0.817 & 0.789 & 0.758 & 0.434 & 0.803 & 0.695 & 0.692\\
    \cline{2-11}
    & RTN  & 5 & 0.255 & 0.283 & 0.437 & 0.277 & 0.260 & 0.529 & 0.509 & 0.364\\
    & HQQ  & 5 & 0.244 & 0.303 & 0.474 & 0.288 & 0.254 & 0.521 & 0.518 & 0.372\\
    \rowcolor{gray!20} 
    \global\let\CT@@do@color\relax &
    \global\let\CT@@do@color\oriCT@@do@color
    \textbf{WGM}  & 5 & \textbf{0.529} & \textbf{0.807} & \textbf{0.747} & 0.738 & \textbf{0.436} & \textbf{0.798} & \textbf{0.676} & \textbf{0.676}\\
    \rowcolor{gray!20} 
    \global\let\CT@@do@color\relax &
    \global\let\CT@@do@color\oriCT@@do@color
    \textbf{WGM-LO}  & 5 & 0.501 & 0.790 & \textbf{0.747} & \textbf{0.747} & 0.418 & 0.797 & \textbf{0.677} & 0.668\\

    \hline
  \end{tabular}
  
  \label{tab:result-QA-task}
\end{table*}

\begin{table}
  \centering
  \setlength{\tabcolsep}{0.4mm}
  \small 
  \caption{
  PPL performance of pretrained LLMs under 5-bit per-tensor quantization across three datasets.
  }
  \begin{tabular}{@{}lcc| cccc@{}} 
  
    \hline
    \textbf{Model} & \textbf{Method} & \textbf{Bits} & \textbf{Wikitext 2} $\downarrow$& \textbf{PTB} $\downarrow$& \textbf{C4} $\downarrow$& \textbf{Avg.} $\downarrow$\\
    \hline
    \multirow{7}{3em}{Llama 3.2 1B} & FP & 16 & 9.75 & 17.59 & 14.01 & 13.78\\
    \cline{2-7}
    & RTN  & 5 & 160604.80 & 80296.23 & 112862.91 & 117921.32\\
    & HQQ  & 5 & 152015.38 & 89273.23 & 93242.31 & 111510.31\\
    \rowcolor{gray!20} 
    \global\let\CT@@do@color\relax &
    \global\let\CT@@do@color\oriCT@@do@color
    \textbf{WGM}  & 5 & 11.32 & 20.28 & 16.47 & 16.02\\
    \rowcolor{gray!20} 
    \global\let\CT@@do@color\relax &
    \global\let\CT@@do@color\oriCT@@do@color
    \textbf{WGM-LO}  & 5 & 13.21 & 26.06 & 19.43 & 19.57\\

    \hline
    \multirow{7}{3em}{Llama 3.2 3B} & FP & 16 & 7.81 & 13.53 & 11.33 & 10.89\\
    \cline{2-7}
    & RTN  & 5 & 196842.44 & 115724.54 & 87181.27 & 133249.41\\
    & HQQ  & 5 & 23792.54 & 31359.14 & 22844.35 & 25998.68\\
    \rowcolor{gray!20} 
    \global\let\CT@@do@color\relax &
    \global\let\CT@@do@color\oriCT@@do@color
    \textbf{WGM}  & 5 & 2325.18 & 2728.98 & 2814.28 & 2622.81\\
    \rowcolor{gray!20} 
    \global\let\CT@@do@color\relax &
    \global\let\CT@@do@color\oriCT@@do@color
    \textbf{WGM-LO}  & 5 & 446.43 & 548.06 & 764.34 & 586.28\\

    \hline
    \multirow{7}{3em}{Falcon 3 1B} & FP & 16 & 9.15 & 19.67 & 17.65 & 15.49\\
    \cline{2-7}
    & RTN  & 5 & 55.00 & 131.81 & 96.32 & 94.38\\
    & HQQ  & 5 & 24.87 & 52.11 & 46.64 & 41.21\\
    \rowcolor{gray!20} 
    \global\let\CT@@do@color\relax &
    \global\let\CT@@do@color\oriCT@@do@color
    \textbf{WGM}  & 5 & 10.37 & 23.75 & 20.08 & 18.07\\
    \rowcolor{gray!20} 
    \global\let\CT@@do@color\relax &
    \global\let\CT@@do@color\oriCT@@do@color
    \textbf{WGM-LO}  & 5 & 17.32 & 46.43 & 31.88 & 31.88\\

    \hline
    \multirow{7}{3em}{Falcon 3 3B} & FP & 16 & 8.02 & 16.49 & 15.99 & 13.50\\
    \cline{2-7}
    & RTN  & 5 & 82.58 & 189.23 & 148.03 & 139.95\\
    & HQQ  & 5 & 387.51 & 705.79 & 457.26 & 516.85\\
    \rowcolor{gray!20} 
    \global\let\CT@@do@color\relax &
    \global\let\CT@@do@color\oriCT@@do@color
    \textbf{WGM}  & 5 & 9.53 & 20.92 & 18.92 & 16.46\\
    \rowcolor{gray!20} 
    \global\let\CT@@do@color\relax &
    \global\let\CT@@do@color\oriCT@@do@color
    \textbf{WGM-LO}  & 5 & 9.49 & 20.60 & 18.81 & 16.30\\

    \hline
    \multirow{6}{3em}{Gemma 3 1b} & FP & 16 & 14.17& 104.12 & 20.00 & 46.10\\
    \cline{2-7}
    & RTN  & 5 & 2226.50 & 20169.60 & 1670.33 & 8022.14\\
    & HQQ  & 5 & 1868.31 & 12820.58 & 1592.09 & 5426.99\\
    \rowcolor{gray!20} 
    \global\let\CT@@do@color\relax &
    \global\let\CT@@do@color\oriCT@@do@color
    \textbf{WGM}  & 5 & 17.20 & 236.83 & 23.94 & 92.66\\
    \rowcolor{gray!20} 
    \global\let\CT@@do@color\relax &
    \global\let\CT@@do@color\oriCT@@do@color
    \textbf{WGM-LO}  & 5 & 15.71 & 162.77 & 22.39 & 66.96\\\\

    \hline
    \multirow{7}{3em}{Gemma 3 4b} & FP & 16 & 10.77 & 239.02 & 16.78 & 88.86\\
    \cline{2-7}
    & RTN  & 5 & 42114252.00 & 99411056.00 & 4541906.50 & 48689071.50\\
    & HQQ  & 5 & 20682.83 & 958119.62 & 9247.24 & 329349.90\\
    \rowcolor{gray!20} 
    \global\let\CT@@do@color\relax &
    \global\let\CT@@do@color\oriCT@@do@color
    \textbf{WGM}  & 5 & 13.21 & 553.78 & 20.95 & 195.98\\
    \rowcolor{gray!20} 
    \global\let\CT@@do@color\relax &
    \global\let\CT@@do@color\oriCT@@do@color
    \textbf{WGM-LO}  & 5 & 12.65 & 531.05 & 20.10 & 187.93\\
    \hline
  \end{tabular}
  \label{tab:PPL-comparison}
\end{table}

\makeatletter
\global\let\oriCT@@do@color\CT@@do@color 

\begin{table*}[t] 
  \centering
  \setlength{\tabcolsep}{1mm}
  \footnotesize 
  \caption{
  QA performance of pretrained LLMs under 4-bit per-tensor quantization across seven benchmarks.
  }
  \begin{tabular}{@{}lcc| cccccccccc@{}} 
    \hline
    \textbf{Model} & \textbf{Method} & \textbf{Bits} & \textbf{ARC-C} $\uparrow$& \textbf{ARC-E} $\uparrow$& \textbf{BoolQ} $\uparrow$& \textbf{Hellaswag} $\uparrow$& \textbf{OPQA} $\uparrow$& \textbf{PIQA} $\uparrow$& \textbf{Winogrande} $\uparrow$& \textbf{Avg.} $\uparrow$\\
    \hline
    \multirow{7}{3em}{Llama 3.2 1B} & FP & 16 & 0.364 & 0.605 & 0.639 & 0.636 & 0.374 & 0.744 & 0.612 & 0.568\\
    \cline{2-11}
    & RTN  & 4 & 0.257 & 0.254 & \textbf{0.600} & 0.263 & \textbf{0.298} & 0.517 & 0.500 & \textbf{0.384}\\
    & HQQ  & 4 & \textbf{0.263} & 0.256 & 0.540 & \textbf{0.265} & 0.286 & 0.511 & 0.490 & 0.373\\
    \rowcolor{gray!20} 
    \global\let\CT@@do@color\relax &
    \global\let\CT@@do@color\oriCT@@do@color
    \textbf{WGM}  & 4 & 0.240 & \textbf{0.281} & 0.381 & 0.262 & 0.276 & 0.511 & 0.497 & 0.350\\
    \rowcolor{gray!20} 
    \global\let\CT@@do@color\relax &
    \global\let\CT@@do@color\oriCT@@do@color
    \textbf{WGM-LO}  & 4 & 0.237 & 0.279 & 0.403 & 0.264 & 0.246 & \textbf{0.520} & \textbf{0.502} & 0.350\\
    
    \hline
    \multirow{7}{3em}{Llama 3.2 3B} & FP & 16 & 0.458 & 0.716 & 0.728 & 0.736 & 0.432 & 0.775 & 0.693 & 0.648\\
    \cline{2-11}
    & RTN  & 4 & \textbf{0.266} & 0.251 & \textbf{0.598} & 0.263 & \textbf{0.290} & 0.502 & \textbf{0.526} & \textbf{0.385}\\
    & HQQ  & 4 & 0.253 & 0.244 & 0.564 & 0.265 & 0.278 & \textbf{0.529} & 0.492 & 0.375\\
    \rowcolor{gray!20} 
    \global\let\CT@@do@color\relax &
    \global\let\CT@@do@color\oriCT@@do@color
    \textbf{WGM}  & 4 & 0.257 & \textbf{0.269} & 0.381 & 0.268 & 0.280 & 0.519 & 0.515 & 0.356\\
    \rowcolor{gray!20} 
    \global\let\CT@@do@color\relax &
    \global\let\CT@@do@color\oriCT@@do@color
    \textbf{WGM-LO}  & 4 & 0.251 & 0.268 & 0.378 & \textbf{0.269} & 0.274 & 0.508 & 0.503 & 0.350\\

    \hline
    \multirow{7}{3em}{Falcon 3 1B} & FP & 16 & 0.421 & 0.662 & 0.719 & 0.618 & 0.406 & 0.745 & 0.617 & 0.598\\
    \cline{2-11}
    & RTN  & 4 & 0.271 & 0.286 & 0.519 & 0.266 & 0.288 & 0.508 & 0.500 & 0.377\\
    & HQQ  & 4 & 0.253 & 0.263 & 0.459 & 0.262 & 0.282 & 0.512 & 0.491 & 0.360\\
    \rowcolor{gray!20} 
    \global\let\CT@@do@color\relax &
    \global\let\CT@@do@color\oriCT@@do@color
    \textbf{WGM}  & 4 & 0.279 & \textbf{0.467} & 0.500 & \textbf{0.406} & 0.298 & \textbf{0.648} & \textbf{0.526} & \textbf{0.446}\\
    \rowcolor{gray!20} 
    \global\let\CT@@do@color\relax &
    \global\let\CT@@do@color\oriCT@@do@color
    \textbf{WGM-LO}  & 4 & \textbf{0.281} & 0.429 & \textbf{0.549} & 0.379 & \textbf{0.306} & 0.628 & 0.504 & 0.439\\

    \hline
    \multirow{7}{3em}{Falcon 3 3B} & FP & 16 & 0.472 & 0.727 & 0.735 & 0.652 & 0.394 & 0.751 & 0.650 & 0.626\\
    \cline{2-11}
    & RTN  & 4 & 0.239 & 0.265 & 0.386 & 0.264 & 0.314 & 0.494 & 0.515 & 0.354\\
    & HQQ  & 4 & 0.244 & 0.255 & 0.452 & 0.259 & 0.292 & 0.523 & 0.532 & 0.365\\
    \rowcolor{gray!20} 
    \global\let\CT@@do@color\relax &
    \global\let\CT@@do@color\oriCT@@do@color
    \textbf{WGM}  & 4 & \textbf{0.315} & \textbf{0.534} & \textbf{0.618} & \textbf{0.483} & \textbf{0.328} & \textbf{0.683} & \textbf{0.536} & \textbf{0.499}\\
    \rowcolor{gray!20} 
    \global\let\CT@@do@color\relax &
    \global\let\CT@@do@color\oriCT@@do@color
    \textbf{WGM-LO}  & 4 & 0.271 & 0.435 & 0.508 & 0.384 & 0.298 & 0.642 & 0.510 & 0.435\\

    \hline
    \multirow{7}{3em}{Gemma 3 1b} & FP & 16 & 0.381 & 0.722 & 0.663 & 0.622 & 0.374 & 0.749 & 0.584 & 0.585\\
    \cline{2-11}
    & RTN  & 4 & 0.255 & 0.256 & 0.457 & 0.259 & 0.298 & 0.503 & 0.504 & 0.362\\
    & HQQ  & 4 & 0.252 & 0.270 & 0.442 & 0.266 & 0.292 & 0.526 & 0.485 & 0.362\\
    \rowcolor{gray!20} 
    \global\let\CT@@do@color\relax &
    \global\let\CT@@do@color\oriCT@@do@color
    \textbf{WGM}  & 4 & \textbf{0.366} & 0.633 & 0.529 & 0.609 & 0.370 & 0.720 & 0.563 & 0.541\\
    \rowcolor{gray!20} 
    \global\let\CT@@do@color\relax &
    \global\let\CT@@do@color\oriCT@@do@color
    \textbf{WGM-LO}  & 4 & 0.361 & \textbf{0.659} & \textbf{0.603} & \textbf{0.617} & \textbf{0.380} & \textbf{0.725} & \textbf{0.579} & \textbf{0.561}\\

    \hline
    \multirow{7}{3em}{Gemma 3 4b} & FP & 16 & 0.547 & 0.817 & 0.789 & 0.758 & 0.434 & 0.803 & 0.695 & 0.692\\
    \cline{2-11}
    & RTN  & 4 & 0.265 & 0.253 & 0.499 & 0.262 & 0.284 & 0.494 & 0.506 & 0.366\\
    & HQQ  & 4 & 0.261 & 0.263 & 0.549 & 0.267 & 0.290 & 0.506 & 0.497 & 0.376\\
    \rowcolor{gray!20} 
    \global\let\CT@@do@color\relax &
    \global\let\CT@@do@color\oriCT@@do@color
    \textbf{WGM}  & 4 & \textbf{0.413} & 0.668 & 0.575 & \textbf{0.631} & 0.390 & \textbf{0.749} & \textbf{0.620} & \textbf{0.578}\\
    \rowcolor{gray!20} 
    \global\let\CT@@do@color\relax &
    \global\let\CT@@do@color\oriCT@@do@color
    \textbf{WGM-LO}  & 4 & 0.408 & \textbf{0.678} & \textbf{0.596} & 0.622 & \textbf{0.394} & 0.738 & 0.598 & 0.576\\

    \hline
  \end{tabular}
  
  \label{tab:result-QA-task}
\end{table*}

\begin{table}
  \centering
  \setlength{\tabcolsep}{0.4mm}
  \small 
  \caption{
  PPL performance of pretrained LLMs under 4-bit per-tensor quantization across three datasets.
  }
  \begin{tabular}{@{}lcc| cccc@{}} 
  
    \hline
    \textbf{Model} & \textbf{Method} & \textbf{Bits} & \textbf{Wikitext 2} $\downarrow$& \textbf{PTB} $\downarrow$& \textbf{C4} $\downarrow$& \textbf{Avg.} $\downarrow$\\
    \hline
    \multirow{7}{3em}{Llama 3.2 1B} & FP & 16 & 9.75 & 17.59 & 14.01 & 13.78\\
    \cline{2-7}
    & RTN  & 4 & 2057063.25 & 1895821.88 & 2303753.25 & 2085546.12\\
    & HQQ  & 4 & 3482935.25 & 2573738.25 & 3059556.75 & 3038743.42\\
    \rowcolor{gray!20} 
    \global\let\CT@@do@color\relax &
    \global\let\CT@@do@color\oriCT@@do@color
    \textbf{WGM}  & 4 & 28731.11 & 20510.02 & 12215.58 & 20485.57\\
    \rowcolor{gray!20} 
    \global\let\CT@@do@color\relax &
    \global\let\CT@@do@color\oriCT@@do@color
    \textbf{WGM-LO}  & 4 & 15098.34 & 12288.76 & 7531.40 & 11639.50\\

    \hline
    \multirow{7}{3em}{Llama 3.2 3B} & FP & 16 & 7.81 & 13.53 & 11.33 & 10.89\\
    \cline{2-7}
    & RTN  & 4 & 257957.31 & 96724.30 & 479669.94 & 278117.18\\
    & HQQ  & 4 & 1227376.50 & 1219055.50 & 1298070.25 & 1248167.42\\
    \rowcolor{gray!20} 
    \global\let\CT@@do@color\relax &
    \global\let\CT@@do@color\oriCT@@do@color
    \textbf{WGM}  & 4 & 10806.46 & 12205.56 & 6340.89 & 9784.30\\
    \rowcolor{gray!20} 
    \global\let\CT@@do@color\relax &
    \global\let\CT@@do@color\oriCT@@do@color
    \textbf{WGM-LO}  & 4 & 9696.50 & 10285.10 & 4389.21 & 8123.60\\

    \hline
    \multirow{7}{3em}{Falcon 3 1B} & FP & 16 & 9.15 & 19.67 & 17.65 & 15.49\\
    \cline{2-7}
    & RTN  & 4 & 2917153.50 & 4835918.00 & 6318141.00 & 4690404.17\\
    & HQQ  & 4 & 940907.81 & 2635117.00 & 1690569.75 & 1755531.52\\
    \rowcolor{gray!20} 
    \global\let\CT@@do@color\relax &
    \global\let\CT@@do@color\oriCT@@do@color
    \textbf{WGM}  & 4 & 77.70 & 185.41 & 158.94 & 140.68\\
    \rowcolor{gray!20} 
    \global\let\CT@@do@color\relax &
    \global\let\CT@@do@color\oriCT@@do@color
    \textbf{WGM-LO}  & 4 & 120.04 & 266.01 & 263.32 & 216.46\\

    \hline
    \multirow{7}{3em}{Falcon 3 3B} & FP & 16 & 8.02 & 16.49 & 15.99 & 13.50\\
    \cline{2-7}
    & RTN  & 4 & 11405618.00 & 17293052.00 & 19399576.00 & 16032748.67\\
    & HQQ  & 4 & 1336339.75 & 4439809.00 & 4443197.50 & 3406448.75\\
    \rowcolor{gray!20} 
    \global\let\CT@@do@color\relax &
    \global\let\CT@@do@color\oriCT@@do@color
    \textbf{WGM}  & 4 & 21.05 & 43.04 & 39.46 & 34.52\\
    \rowcolor{gray!20} 
    \global\let\CT@@do@color\relax &
    \global\let\CT@@do@color\oriCT@@do@color
    \textbf{WGM-LO}  & 4 & 73.75 & 93.26 & 136.10 & 101.04\\

    \hline
    \multirow{7}{3em}{Gemma 3 1b} & FP & 16 & 14.17& 104.12 & 20.00 & 46.10\\
    \cline{2-7}
    & RTN  & 4 & 21185708.00 & 31281078.00 & 12719446.00 & 21728744.00\\
    & HQQ  & 4 & 9682463.00 & 24361728.00 & 14554462.00 & 16199551.00\\
    \rowcolor{gray!20} 
    \global\let\CT@@do@color\relax &
    \global\let\CT@@do@color\oriCT@@do@color
    \textbf{WGM}  & 4 & 49.78 & 797.77 & 67.11 & 304.89\\
    \rowcolor{gray!20} 
    \global\let\CT@@do@color\relax &
    \global\let\CT@@do@color\oriCT@@do@color
    \textbf{WGM-LO}  & 4 & 55.29 & 1004.90 & 64.69 & 374.96\\

    \hline
    \multirow{7}{3em}{Gemma 3 4b} & FP & 16 & 10.77 & 239.02 & 16.78 & 88.86\\
    \cline{2-7}
    & RTN  & 4 & 43750133760.00 & 254194106368.00 & 173915586560.00 & 157286608896.00\\
    & HQQ  & 4 & 11806313472.00 & 1974162176.00 & 15918100480.00 & 9899525376.00\\
    \rowcolor{gray!20} 
    \global\let\CT@@do@color\relax &
    \global\let\CT@@do@color\oriCT@@do@color
    \textbf{WGM}  & 4 & 34.70 & 1890.78 & 55.13 & 660.21\\
    \rowcolor{gray!20} 
    \global\let\CT@@do@color\relax &
    \global\let\CT@@do@color\oriCT@@do@color
    \textbf{WGM-LO}  & 4 & 36.44 & 2768.71 & 55.18 & 953.44\\
    \hline
  \end{tabular}
  \label{tab:PPL-comparison}
\end{table}

\clearpage
\section{Double Quantization}
\label{sec:dq}

We further evaluate a double-quantization variant that quantizes the quantization parameters, scaler, in addition to the weights recursively once more using the same WGM algorithm. 
Tables~\ref{tab:DQ-PPL} and~\ref{tab:DQ-QA} compare QA and PPL between single and double quantization. The effect is stable across all tested models: double quantization consistently leads to a small degradation (lower QA and higher PPL) relative to single quantization. Since the observed changes are uniform and do not improve end-task quality, we focus on single quantization in the main paper and include double quantization here as a supplementary analysis.

\makeatletter
\global\let\oriCT@@do@color\CT@@do@color 

\begin{table*}[h] 
  \centering
  \setlength{\tabcolsep}{1mm}
  \footnotesize 
  \caption{QA performance of double quantization of pretrained LLMs using WGM in 4-bit block-wise quantization. Double quantization are on blocks of quantization parameters of size 2048, bit=6. so on average each parameter cost 6 + 32*16/2048=6.25 bits and the whole matrix cost 4 + 8*6.25/64=4.78125 bits.
  }
  \begin{tabular}{@{}lcc| cccccccccc@{}} 
    \hline
    \textbf{Model} & \textbf{Method} & \textbf{Bits} & \textbf{ARC-C} $\uparrow$& \textbf{ARC-E} $\uparrow$& \textbf{BoolQ} $\uparrow$& \textbf{Hellaswag} $\uparrow$& \textbf{OPQA} $\uparrow$& \textbf{PIQA} $\uparrow$& \textbf{Winogrande} $\uparrow$& \textbf{Avg.} $\uparrow$\\
    \hline
    \multirow{2}{3.6em}{Llama 3.2 1B pt} & WGM & 4 & 0.343 & 0.570 & 0.625 & 0.607 & 0.330 & 0.725 & 0.583 & 0.540\\
    \rowcolor{gray!20} 
    \global\let\CT@@do@color\relax &
    \global\let\CT@@do@color\oriCT@@do@color
    \textbf{WGM-dq}  & 4 & 0.335 & 0.567 & 0.597 & 0.600 & 0.332 & 0.719 & 0.588 & 0.534\\

    \hline
    \multirow{2}{3.6em}{Llama 3.2 1B it} & WGM &  4 & 0.363 & 0.607 & 0.668 & 0.587 & 0.336 & 0.720 & 0.577 & 0.551\\
    \rowcolor{gray!20} 
    \global\let\CT@@do@color\relax &
    \global\let\CT@@do@color\oriCT@@do@color
    \textbf{WGM-dq}  & 4 & 0.347 & 0.596 & 0.651 & 0.572 & 0.328 & 0.705 & 0.563 & 0.537\\

    \hline
    \multirow{2}{3.6em}{Llama 3.2 3B pt} & WGM & 4 & 0.434 & 0.683 & 0.726 & 0.714 & 0.406 & 0.761 & 0.694 & 0.631\\
    \rowcolor{gray!20} 
    \global\let\CT@@do@color\relax &
    \global\let\CT@@do@color\oriCT@@do@color
    \textbf{WGM-dq}  & 4 & 0.381 & 0.615 & 0.516 & 0.625 & 0.380 & 0.741 & 0.652 & 0.559\\

    \hline
    \multirow{2}{3.6em}{Llama 3.2 3B it} & WGM & 4 & 0.437 & 0.681 & 0.734 & 0.686 & 0.372 & 0.732 & 0.655 & 0.614\\
    \rowcolor{gray!20} 
    \global\let\CT@@do@color\relax &
    \global\let\CT@@do@color\oriCT@@do@color
    \textbf{WGM-dq}  & 4 & 0.361 & 0.593 & 0.559 & 0.583 & 0.336 & 0.695 & 0.596 & 0.532\\

    \hline
    \multirow{2}{3.6em}{Falcon3 1B pt} & WGM & 4 & 0.415 & 0.662 & 0.700 & 0.611 & 0.408 & 0.746 & 0.598 & 0.591\\
    \rowcolor{gray!20} 
    \global\let\CT@@do@color\relax &
    \global\let\CT@@do@color\oriCT@@do@color
    \textbf{WGM-dq}  & 4 & 0.411 & 0.666 & 0.709 & 0.608 & 0.410 & 0.745 & 0.597 & 0.592\\

    \hline
    \multirow{2}{3.6em}{Falcon3 1B it} & WGM & 4 & 0.459 & 0.683 & 0.728 & 0.627 & 0.402 & 0.751 & 0.606 & 0.608\\
    \rowcolor{gray!20} 
    \global\let\CT@@do@color\relax &
    \global\let\CT@@do@color\oriCT@@do@color
    \textbf{WGM-dq}  & 4 & 0.457 & 0.684 & 0.730 & 0.625 & 0.398 & 0.744 & 0.609 & 0.607\\

    \hline
    \multirow{2}{3.6em}{Falcon3 3B pt} & WGM & 4 & 0.455 & 0.714 & 0.703 & 0.635 & 0.384 & 0.751 & 0.638 & 0.611\\
    \rowcolor{gray!20} 
    \global\let\CT@@do@color\relax &
    \global\let\CT@@do@color\oriCT@@do@color
    \textbf{WGM-dq} & 4 & 0.452 & 0.716 & 0.701 & 0.630 & 0.388 & 0.751 & 0.646 & 0.612\\

    \hline
    \multirow{2}{3.6em}{Falcon3 3B it} & WGM & 4 & 0.480 & 0.735 & 0.720 & 0.638 & 0.416 & 0.749 & 0.595 & 0.619\\
    \rowcolor{gray!20} 
    \global\let\CT@@do@color\relax &
    \global\let\CT@@do@color\oriCT@@do@color
    \textbf{WGM-dq}  & 4 & 0.478 & 0.723 & 0.710 & 0.637 & 0.414 & 0.743 & 0.595 & 0.614\\

    \hline
    \multirow{2}{3.6em}{Gemma 3 1b pt} & WGM & 4 & 0.370 & 0.703 & 0.647 & 0.617 & 0.374 & 0.742 & 0.581 & 0.576\\
    \rowcolor{gray!20} 
    \global\let\CT@@do@color\relax &
    \global\let\CT@@do@color\oriCT@@do@color
    \textbf{WGM-dq}  & 4 & 0.369 & 0.697 & 0.621 & 0.614 & 0.366 & 0.747 & 0.579 & 0.571\\

    \hline
    \multirow{2}{3.6em}{Gemma 3 1b it} & WGM & 4 & 0.394 & 0.611 & 0.735 & 0.562 & 0.368 & 0.724 & 0.584 & 0.568\\
    \rowcolor{gray!20} 
    \global\let\CT@@do@color\relax &
    \global\let\CT@@do@color\oriCT@@do@color
    \textbf{WGM-dq}  & 4 & 0.384 & 0.629 & 0.735 & 0.558 & 0.362 & 0.719 & 0.568 & 0.565\\

    \hline
    \multirow{2}{3.6em}{Gemma 3 4b pt} & WGM & 4 & 0.529 & 0.817 & 0.786 & 0.754 & 0.434 & 0.799 & 0.698 & 0.688\\
    \rowcolor{gray!20} 
    \global\let\CT@@do@color\relax &
    \global\let\CT@@do@color\oriCT@@do@color
    \textbf{WGM-dq}  & 4 & 0.532 & 0.821 & 0.793 & 0.756 & 0.434 & 0.795 & 0.697 & 0.690\\

    \hline
    \multirow{2}{3.6em}{Gemma 3 4b it} & WGM & 4 & 0.561 & 0.754 & 0.831 & 0.738 & 0.468 & 0.772 & 0.679 & 0.686\\
    \rowcolor{gray!20} 
    \global\let\CT@@do@color\relax &
    \global\let\CT@@do@color\oriCT@@do@color
    \textbf{WGM-dq} & 4 & 0.552 & 0.743 & 0.829 & 0.737 & 0.462 & 0.768 & 0.676 & 0.681\\

    \hline
  \end{tabular}
  
  \label{tab:DQ-QA}
\end{table*}

\begin{table}
  \centering
  \setlength{\tabcolsep}{0.4mm}
  \small 
  \caption{
  PPL performance of double quantization of pretrained LLMs using WGM in 4-bit block-wise quantization. With the same setting.
  }
  \begin{tabular}{@{}lcc| cccc@{}} 
  
    \hline
    \textbf{Model} & \textbf{Method} & \textbf{Bits} & \textbf{Wikitext 2} $\downarrow$& \textbf{PTB} $\downarrow$& \textbf{C4} $\downarrow$& \textbf{Avg.} $\downarrow$\\
    \hline
    \multirow{2}{3.6em}{Llama 3.2 1B pt} & WGM & 4 & 10.89 & 19.47 & 15.98 & 15.45\\
    \cline{2-7}
    \rowcolor{gray!20} 
    \global\let\CT@@do@color\relax &
    \global\let\CT@@do@color\oriCT@@do@color
    \textbf{WGM dq}  & 4 & 11.17 & 19.92 & 16.39 & 15.83\\

    \hline
    \multirow{2}{3.5em}{Llama 3.2 1B it} & WGM & 4 & 15.25 & 29.39 & 23.90 & 22.85\\
    \cline{2-7}
    \rowcolor{gray!20} 
    \global\let\CT@@do@color\relax &
    \global\let\CT@@do@color\oriCT@@do@color
    \textbf{WGM dq}  & 4 & 16.71 & 32.06 & 25.90 & 24.89\\

    \hline
    \multirow{2}{3.6em}{Llama 3.2 3B pt} & WGM & 4 & 8.43 & 14.54 & 12.49 & 11.81\\
    \cline{2-7}
    \rowcolor{gray!20} 
    \global\let\CT@@do@color\relax &
    \global\let\CT@@do@color\oriCT@@do@color
    \textbf{WGM dq}  & 4 & 16.12 & 33.39 & 26.62 & 25.37\\

    \hline
    \multirow{2}{3.5em}{Llama 3.2 3B it} & WGM & 4 & 12.02 & 24.55 & 17.47 & 18.01\\
    \cline{2-7}
    \rowcolor{gray!20} 
    \global\let\CT@@do@color\relax &
    \global\let\CT@@do@color\oriCT@@do@color
    \textbf{WGM dq}  & 4 & 20.88 & 73.80 & 40.00 & 44.89\\

    \hline
    \multirow{2}{3em}{Falcon 3 1B pt} & WGM & 4 & 9.31 & 20.08 & 17.96 & 15.78\\
    \cline{2-7}
    \rowcolor{gray!20} 
    \global\let\CT@@do@color\relax &
    \global\let\CT@@do@color\oriCT@@do@color
    \textbf{WGM-dq}  & 4 & 9.37 & 20.18 & 18.05 & 15.87\\

    \hline
    \multirow{2}{3em}{Falcon 3 1B it} & WGM & 4 & 10.42 & 22.51 & 19.62 & 17.51\\
    \cline{2-7}
    \rowcolor{gray!20} 
    \global\let\CT@@do@color\relax &
    \global\let\CT@@do@color\oriCT@@do@color
    \textbf{WGM-dq}  & 4 & 10.51 & 22.59 & 19.69 & 17.60\\

    \hline
    \multirow{2}{3em}{Falcon 3 3B pt} & WGM  & 4 & 8.26 & 17.10 & 16.47 & 13.94\\
    \cline{2-7}
    \rowcolor{gray!20} 
    \global\let\CT@@do@color\relax &
    \global\let\CT@@do@color\oriCT@@do@color
    \textbf{WGM-dq}  & 4 & 8.36 & 17.32 & 16.60 & 14.09\\

    \hline
    \multirow{2}{3em}{Falcon 3 3B it} & WGM & 4 & 11.01 & 21.66 & 20.31 & 17.66\\
    \cline{2-7}
    \rowcolor{gray!20} 
    \global\let\CT@@do@color\relax &
    \global\let\CT@@do@color\oriCT@@do@color
    \textbf{WGM-dq}  & 4 & 11.47 & 22.33 & 20.85 & 18.21\\

    \hline
    \multirow{2}{3em}{Gemma 3 1b pt} & WGM & 4 & 246.69 & 1111.53 & 330.01 & 562.75\\
    \cline{2-7}
    \rowcolor{gray!20} 
    \global\let\CT@@do@color\relax &
    \global\let\CT@@do@color\oriCT@@do@color
    \textbf{WGM-dq}  & 4 & 253.96 & 1225.12 & 332.91 & 604.00\\

    \hline
    \multirow{2}{3em}{Gemma 3 1b it} & WGM & 4 & 538.23 & 1477.78 & 553.44 & 856.49\\
    \cline{2-7}
    \rowcolor{gray!20} 
    \global\let\CT@@do@color\relax &
    \global\let\CT@@do@color\oriCT@@do@color
    \textbf{WGM-dq}  & 4 & 534.46 & 1451.77 & 533.80 & 840.01\\

    \hline
    \multirow{2}{3em}{Gemma 3 4b pt} & WGM & 4 & 58.66 & 369.94 & 91.59 & 173.40\\
    \cline{2-7}
    \rowcolor{gray!20} 
    \global\let\CT@@do@color\relax &
    \global\let\CT@@do@color\oriCT@@do@color
    \textbf{WGM-dq}  & 4 & 63.87 & 387.97 & 99.62 & 183.82\\

    \hline
    \multirow{2}{3em}{Gemma 3 4b it} & WGM & 4 & 86.06 & 822.51 & 112.66 & 340.41\\
    \cline{2-7}
    \rowcolor{gray!20} 
    \global\let\CT@@do@color\relax &
    \global\let\CT@@do@color\oriCT@@do@color
    \textbf{WGM-dq}  & 4 & 85.01 & 849.22 & 112.73 & 348.97\\
    \hline
  \end{tabular}
  \label{tab:DQ-PPL}
\end{table}

\section{GPTQ performance degrade on Llama 3.2 1B pretrained}

So we evaluated using GPTQ for Llama-3.2 1B pretrained check point found on hugging face, namely shuyuej/Llama-3.2-1B-GPTQ (GPTQ A) and saul95/Llama-3.2-1B-GPTQ (GPTQ B). Then we also used known correct implementation, GPTQModel\footnote{https://github.com/ModelCloud/GPTQModel}, to quantize on Llama 3.2 1B (GPTQ C). Across these variants, GPTQ exhibited substantially degraded performance under our evaluation protocol, suggesting that the observed behavior is unlikely to be an artifact of a single repository, checkpoint, or invocation.

One possible explanation is sensitivity to the calibration procedure. GPTQ relies on small calibration sets, and misalignment between calibration data and the downstream evaluation distribution can lead to overfitting or instability in certain models or settings. In contrast, calibration-free PTQ approaches do not condition on such data and often display more uniform behavior across architectures and datasets, as observed in prior reports \citep{malinovskii2025higgs,Guo2024GPTQTQL,lin2023awq}. We emphasize that this remains a hypothesis; a systematic study varying calibration sample size, selection strategy, and layer-wise objectives would be required to establish causality.

\begin{table*}[t] 
  \centering
  \setlength{\tabcolsep}{1mm}
  \footnotesize 
  \caption{QA of GPTQ A,B and C}
  \begin{tabular}{@{}lcc| cccccccccc@{}} 
    \hline
    \textbf{Model} & \textbf{Method} & \textbf{Bits} & \textbf{ARC-C} $\uparrow$& \textbf{ARC-E} $\uparrow$& \textbf{BoolQ} $\uparrow$& \textbf{Hellaswag} $\uparrow$& \textbf{OPQA} $\uparrow$& \textbf{PIQA} $\uparrow$& \textbf{Winogrande} $\uparrow$& \textbf{Avg.} $\uparrow$\\
    \hline
    \multirow{4}{3em}{Llama 3.2 1B} & FP & 16 & 0.364 & 0.605 & 0.639 & 0.636 & 0.374 & 0.744 & 0.612 & 0.568\\
    \cline{2-11}
    & GPTQ A & 4 & 0.317 & 0.553 & 0.502 & 0.357 & 0.354 & 0.694 & 0.583 & 0.480\\
    & GPTQ B & 4 & 0.334 & 0.561 & 0.578 & 0.455 & 0.372 & 0.723 & 0.600 & 0.518\\
    & GPTQ C & 4 & 0.253 & 0.479 & 0.509 & 0.358 & 0.298 & 0.658 & 0.552 & 0.444\\
    \hline
  \end{tabular}
  
  \label{tab:result-QA-task}
\end{table*}

\makeatletter
\global\let\oriCT@@do@color\CT@@do@color

\begin{table}
  \centering
  \setlength{\tabcolsep}{0.4mm}
  \small 
  \caption{PPL of GPTQ A,B and C}
  \begin{tabular}{@{}lcc| cccc@{}} 
  
    \hline
    \textbf{Model} & \textbf{Method} & \textbf{Bits} & \textbf{Wikitext 2} $\downarrow$& \textbf{PTB} $\downarrow$& \textbf{C4} $\downarrow$& \textbf{Avg.} $\downarrow$\\
    \hline
    \multirow{4}{3em}{Llama 3.2 1B} & FP & 16 & 9.75 & 17.59 & 14.01 & 13.78\\
    \cline{2-7}
    & GPTQ A & 4 & 195.34 & 213.91 & 302.27 & 237.18\\
    & GPTQ B & 4 & 40.87 & 64.06 & 65.72 & 56.88\\
    & GPTQ C & 4 & 113.16 & 131.91 & 176.33 & 140.47\\

    \hline
  \end{tabular}
  \label{tab:PPL-comparison}
\end{table}

\end{document}